# Outlier Detection by Logic Programming

Fabrizio Angiulli[1], Gianluigi Greco[2], and Luigi Palopoli[3]

[1] ICAR-CNR, Via Pietro Bucci 41C, 87030 Rende (CS), Italy
angiulli@icar.cnr.it
[2] Dipartimento di Matematica - Università della Calabria, Via Pietro Bucci 30B, 87030 Rende (CS), Italy
ggreco@mat.unical.it
[3] DEIS - Università della Calabria, Via P. Bucci 41C, 87030 Rende (CS), Italy
palopoli@deis.unical.it

**Abstract.** The development of effective knowledge discovery techniques has become a very active research area in recent years due to the important impact it has had in several relevant application domains. One interesting task therein is that of singling out anomalous individuals from a given population, e.g., to detect rare events in time-series analysis settings, or to identify objects whose behavior is deviant w.r.t. a codified standard set of rules. Such exceptional individuals are usually referred to as *outliers* in the literature.
In this paper, the concept of outlier is formally stated in the context of knowledge-based systems, by generalizing that originally proposed in [7] in the context of default theories. The chosen formal framework here is that of logic programming, wherein potential applications of techniques for outlier detection are thoroughly discussed. The proposed formalization is a novel one and helps to shed light on the nature of outliers occurring in logic bases. Also the exploitation of minimality criteria in outlier detection is illustrated.
The computational complexity of outlier detection problems arising in this novel setting is also thoroughly investigated and accounted for in the paper. Finally, rewriting algorithms are proposed that transform any outlier detection problem into an equivalent inference problem under stable model semantics, thereby making outlier computation effective and realizable on top of any stable model solver.

# 1 Introduction

## 1.1 Statement of the Problem

The development of effective knowledge discovery techniques has become a very active research area in recent yeas due to the important impact it has had in several relevant application areas. Knowledge discovery comprises quite diverse tasks and associated methods. One interesting task therein is that of singling out anomalous individuals from a given population, e.g., to detect rare events in time-series analysis settings, or to identify objects whose behavior is deviant w.r.t. a codified standard set of "social" rules. Such exceptional individuals are usually referred to as *outliers* in the literature.

*Outlier detection* has important applications in bioinformatics [1], fraud detection [33, 34], and intrusion detection [32, 48], just to cite a few. As a consequence, several approaches have been already developed to realize outlier detection, mainly by means of data mining techniques including clustering-based and proximity-based methods as well as domain density analysis (see, e.g., [11, 3, 16, 55, 8, 12]). Usually, these approaches model the "normal" behavior of individuals by performing some statistical kind of computation on the given data set (various methods basically differ on the basis of the way such computation is carried out) and, then, single out those individuals whose behavior or characteristics "significantly" deviate from the normal ones.

As a first, quite simple, example of outlier detection, consider the following short story: *Nino is a young soccer player from Southern Italy. He has black hair and brown eyes. Where Nino lives, all people have black hair and brown eyes. Nino has only one big passion, that is, playing soccer and he does not know much about the world outside his home town. One day the boss of a team from Finland happens to see Nino playing and likes him so much that he decides to propose a contract for Nino to play with his team in Finland. A good wage is proposed, so Nino accepts, being also amazed by the strange appearance of this guy coming from Finland. He has blonde hair and blue eyes, indeed: 'He must really be a strange man', Nino must have thought. The day then arrives for Nino to move to Finland. His new boss welcomes him at the airport in Helsinki and, together, they go to the football ground. In a large hall, the boss introduces Nino to his colleagues and to other people working for the team. To his great surprise Nino notices that among so many people in the hall there is just one person with black hair and brown eyes: that is, Nino himself !*

The above example clarifies some simple and yet important concepts. At the beginning, *Nino* considered the team boss from Finland an outlier, on the basis of the statistical evidence that all the people he had ever seen had a different complexion. But at the end of the story, on the basis of the new observations he has acquired in the hall, the team boss no longer appears to *Nino* as a strange individual. Thus, the story firstly illustrates that outliers may be singled out precisely because there is a second set of data determining their abnormality, which below will be called *outlier witness*. Hence, there is no outlier without a witness. Moreover, the fact that at the end *Nino* himself appears to be an outlier with respect to his new colleagues sheds some light on another aspect of outlier detection. Indeed, the story also indicates that an individual might well be an outlier in one context, i.e., with respect to a given set of observations, while not being so in a different one. It follows that while looking for outliers in general no assumption can be made on the existence of properties characterizing outliers "per se". Rather, outliers are to be detected only on the basis of observations to hand, by eventually singling out some properties standing out for their abnormality.

In the short story, abnormality is just the result of a kind of implicitly computed statistic, which is precisely what is commonly assumed by most of the approaches in the literature making use of "quantitative" aspects of the observations only. However, while looking at a set of observations to discover outliers, it often happens that there are some "qualitative" descriptions of the domain of interest, encoding, e.g., what an expected normal behavior should be. As an example of such a description, which will be called *background* *knowledge* in the following, there might be a rule stating that people from Southern Italy have black hair and brown eyes, thereby precisely encoding *Nino*'s observations about people from his place. This description might be, for instance, derived by an expert and might be formalized by means of a suitable language for knowledge representation, as shown in the following example, where logic programs under the stable model semantics are considered.

*Example 1.* Suppose that during a visit to Australia you notice a *mammal*, say *Donald*, that you classify as a *platypus* because of its graceful, yet comical appearance. However, it seems to you that *Donald* is giving birth to a young, but this is very strange given that you know that a platypus is a peculiar mammal that lays eggs. Formalizing this scenario as an outlier detection problem is simple. Indeed, observations can be encoded as the facts $\{\mathtt{Mammal}(\mathtt{Donald}),\ \mathtt{GiveBirth}(\mathtt{Donald}),\ \mathtt{Platypus}(\mathtt{Donald})\}$ and the additional knowledge by means of the following logical rule:

$$\mathtt{Platypus}(\mathtt{X}) \leftarrow \mathtt{Mammal}(\mathtt{X}),\ \mathtt{not}\ \mathtt{GiveBirth}(\mathtt{X}).$$

It is worthwhile noting that if you had not observed that *Donald* was a platypus, you would not have inferred such a conclusion, given your background knowledge and the fact $\mathtt{GiveBirth}(\mathtt{Donald})$. However, if for some reason you have doubted the fact that *Donald* was giving birth to its young, then it would come as no surprise that *Donald* was a platypus. Therefore, the fact $\mathtt{GiveBirth}(\mathtt{Donald})$ is precisely recognized to represent an outlier, whose anomaly is indeed witnessed by the fact $\mathtt{Platypus}(\mathtt{Donald})$. $\triangleleft$

Though still very simple, the example above conveys some of the relevant features occurring in the exploitation of a background knowledge for detecting outliers in a given set of observations. First, the example evidences once again that abnormality of outliers is not defined in "absolute" terms, i.e., there is no explicit encoding for the exceptions in the theory. In particular, outliers show up as kinds of anomalous individuals that do not necessarily lead to a (logical) conflict, so that their isolation cannot be carried out in general by exploiting inconsistencies between observations and the knowledge at hand. For instance, with the encoding of the background knowledge in Example 1, there is no problem in assuming that all the three observations hold: the logical rule is not inconsistent with the observations, and a model in fact exists. In this respect, there is no explicit need to revise the knowledge and/or the observations.

Indeed, in outlier detection, the fact that an individual is an outlier has to be witnessed by a suitable associated set of observations, the outlier witness, which shows why the individual "deviates" from normality. While in traditional (data-mining) approaches an individual to deviate is formalized on the basis of some sort of statistics on the data, making use of qualitative descriptions about the domain of interest allows us to exploit a rich, logic-based framework.

Specifically, the choice here is to assume that outlier witnesses are sets of facts that are "normally" (that is in the absence of the outliers) explained by the domain knowledge which, in turns, entails precisely the opposite of the witnesses whenever outliers are not singled out. In this way, witnesses are meant to precisely characterize the abnormality of outliers with respect to both the theory and the other data at hand. For instance, in Example 1, $\mathtt{Platypus}(\mathtt{Donald})$ is a witness for the outlier $\mathtt{GiveBirth}(\mathtt{Donald})$, since $\mathtt{Platypus}(\mathtt{Donald})$ cannot be predicted by the theory given that one trusts in the fact $\mathtt{GiveBirth}(\mathtt{Donald})$, but is immediately entailed as soon as $\mathtt{GiveBirth}(\mathtt{Donald})$ is thrown out.

It is worthwhile noting that from this perspective, outlier detection may be abstractly seen as a form of diagnostic reasoning where diagnosis of anomalies has to be identified on the basis of some kind of disagreement with the background knowledge and must be supported by some further evidence in the data (such evidences being encoded in witness sets) — connections with this form of reasoning will be further clarified in the following with some examples (cf. Example 3) and discussions.

Clearly enough, detecting outliers by exploiting a logical characterization of the domain of interest is generally more complex than it appears from Example 1. Therefore, in the presence of complex and richer background knowledge, some automatic mechanisms for outlier identification via logic theories is definitively needed. Formalizing these kinds of mechanism, discussing their complexity and providing computation algorithms is, in a nutshell, the contribution of this paper. In particular, the formalization here is carried out by exploiting logic programs under the stable model semantics. This means that the background knowledge, that is, what is known in general about the world (also called in the following *rule component*), and the observations, that is, what is currently perceived about (possibly, a portion of) the world (also called in the following *observation component*), are respectively encoded in the form of a logic program and a set of facts under the stable models semantics.

In order to make the framework clearer, before detailing the major contributions of this paper, more examples of outlier detection are provided and relationships with some related reasoning mechanisms are discussed next.

### 1.2 Examples of Outlier Detection

Considering outlier detection problems in the presence of background knowledge may have several useful applications. For instance, the concept of the outlier in database applications might be quite natural in encoding such important tasks as maintaining database integrity through updates as discussed below.

While designing relational database applications, relations are often equipped with some kind of integrity constraints in order to enhance their expressiveness. In normal operative conditions databases are assumed to be *consistent*, i.e., to satisfy all the integrity constraints. However, it may happen that this is not the case, especially when data is the result of the integration of several autonomous sources such as in datawarehouse and data integration contexts. In fact, the problem of identifying and handling inconsistencies in databases has recently received a lot of attention (cf. [65, 14, 38, 62, 20]). Actually, most of the proposed approaches define constraints using suitable fragments of first order logics, and mainly focus on very simple constraints on data, such as functional dependencies and inclusion dependencies (see, e.g., [2]). In those cases, constraint violations may be identified by means of simple queries over the database at hand. However, more sophisticated forms of constraints modelling specific knowledge of the application domain are left generally unexpressed. These kinds of constraint may encode, e.g., organizational rules or routinely applied praxis, which should be formalized in order to construct a more precise modelling of the domain. In these cases, the problem is that there is no obvious way for checking database integrity.

Outlier detection can in fact be used to examine database integrity by allowing for more sophisticated, application-oriented forms of constraints. If an abnormal property is discovered in a database, i.e., a violation of some constraint, the data source which reported this observation would have to be double-checked. An example follows.

*Example 2.* Consider a bank $B$. The bank approves loan requests put forward by customers on the basis of certain policies. As an example, assume that loan approval policies prescribe that loans for amounts greater than 50K Euro have to come along with an endorsement provided as a guarantee by a third party. Information about approved loan requests is stored in a number of relational tables, that are:

– REQLOAN(LOAN ID, CUSTOMER, AMOUNT), that records loan requests;
– ENDORSEMENT(LOAN ID, PARTY), that records the guaranteeing parties for loan requests for more than 50K Euro;
– APPROVED(LOAN ID), that records approved loan requests.

REQLOAN:

| | | |
|---|---|---|
| $l_1$ | $c_1$ | 57.000 |
| $l_2$ | $c_3$ | 27.000 |
| $l_3$ | $c_2$ | 88.000 |
| $l_4$ | $c_3$ | 80.000 |

ENDORSEMENT:

| | |
|---|---|
| $l_1$ | $p_1$ |
| $l_3$ | $p_2$ |
| $l_4$ | $p_4$ |

APPROVED:

| |
|---|
| $l_1$ |
| $l_2$ |
| $l_4$ |

UNRELIABLE:

| |
|---|
| $p_1$ |
| $p_2$ |

**Fig. 1.** Example instance of the bank database.

Moreover, the bank stores information about unreliable customers in the table UNRELIABLE(CUSTOMER) collecting data from an external data source, such as, for example, consulting agencies providing information about the financial records of individuals and companies. In Figure 1, an instance of the bank database is reported.

With this knowledge to hand, the bank policy concerning loan approvals can be easily encoded using a logical rule like the following:

$$\mathtt{Approved(L)} \leftarrow \mathtt{ReqLoan(L,C,A)},\ \mathtt{A} > 50.000,\ \mathtt{Endorsement(L,P)},\ \mathtt{not}\ \mathtt{Unreliable(P)}.$$

According to this knowledge, and by using the straightforward and well-known correspondence between relational tuples and logical facts (e.g., the tuple $\langle l_1 \rangle$ in relation APPROVED corresponds to the fact $\mathtt{Approved}(l_1)$), it might be noticed that there is something strange with the loan $l_1$ in Figure 1. In fact, since the loan has been approved, the party $p_1$ is supposed to be reliable. However, this is not the case, as emerges by looking at the database provided by the consulting agency (see table UNRELIABLE).

Notice that if the fact $\mathtt{Approved}(l_1)$ were dropped, the exact opposite would have been concluded, namely that the loan $l_1$ is not in the relation APPROVED, i.e., according to the normal behavior of the bank, the loan $l_1$ is not to be approved. Furthermore, if both the facts $\mathtt{Unreliable}(p_1)$ and $\mathtt{Approved}(l_1)$ were dropped, again it would be concluded that $l_1$ might be approved. This implies that the loan request $l_1$ not being approved is a consequence of the fact that $p_1$ is not reliable, and hence $\mathtt{Unreliable}(p_1)$ is an outlier witnessed by $\mathtt{Approved}(l_1)$. This entails that, in the above scenario, the bank has not trusted the consulting agency since the bank has actually decided that $p_1$ is to be considered indeed reliable. $\lhd$

Further potential applications of the logic-based framework for outlier detection comes in the context of enhancing the reasoning capabilities of autonomous agents. Indeed, recall that computational logics have been successfully exploited recently in the context of agent systems applications, since it is quite an effective and powerful way for both modelling and prototypically implementing several forms of reasoning schemes in such systems [45, 85, 83]. For instance, abductive logic programming approaches have been fruitfully used to allow an agent to make hypotheses about the outer world and causes of observable events (e.g, [21]), modal logic operators have been exploited to describe and realize agent behavior and to put it into relationship with other agents in an agent society (e.g., [74]), and inductive techniques have been proposed to enable an agent to "learn" from experience (e.g, [15]).

While exploiting different syntaxes and specific characteristics of the domains of interest, most of these proposals share the same basic structural vision of the agent, which is often assumed to have its own, *trustable background* knowledge about the world that is encoded in the form of a suitable theory. Then, after some *observations* have been obtained describing the actual status of the outer environment, the agent might try to achieve its application goals by performing some suitable reasoning tasks on the basis of both its background knowledge and the observations. In usual agent operative conditions, one may assume that the background knowledge and the observations logically harmonize. However, situations may occur for which this is not the case. In such circumstances, it is desirable that the agent that has noticed some mismatch is $r_1$ : down(X) ← computer(X), not predecessorUp(X).
$r_2$ : predecessorUp(X) ← wired(Y, X), up(Y).
$r_3$ : up(X) ← computer(X), not down(X).
$r_4$ : computer(s). computer(a). ⋯ computer(t).
$r_5$ : wired(s, a). ⋯ wired(g, t).
$r_6$ : up(s).

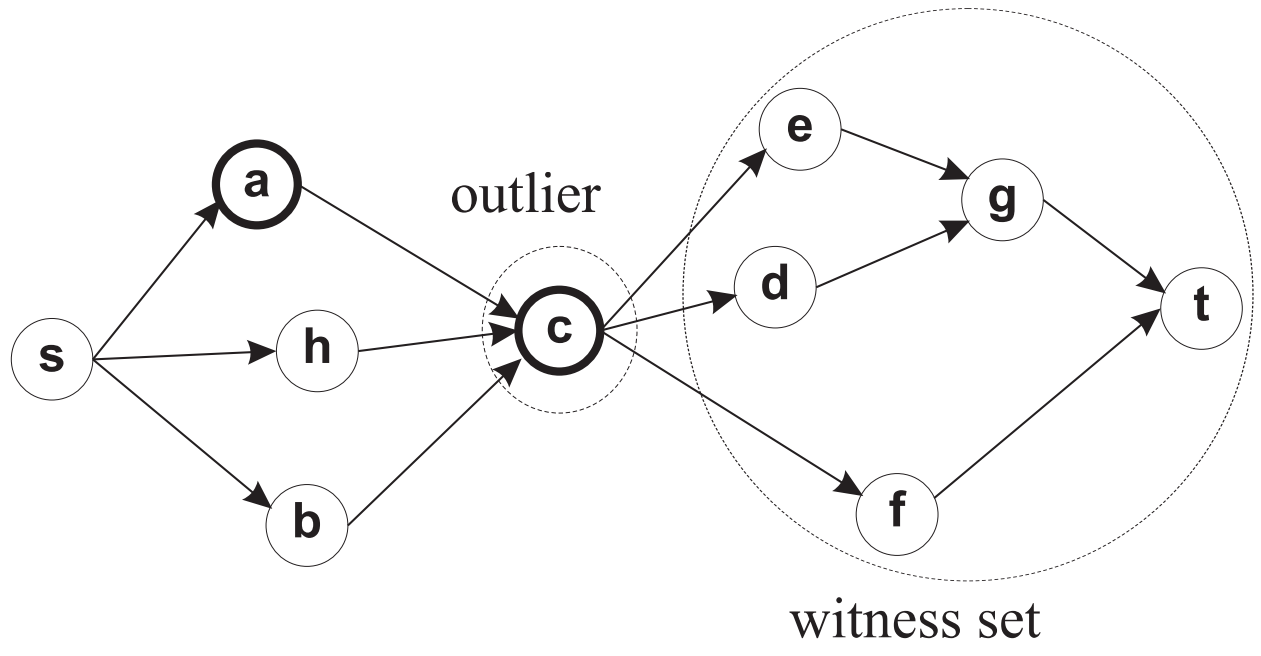

**Fig. 2.** Computer network example.

also able to revise its background knowledge by incorporating the new evidence of the world. In particular, in some cases the background knowledge and observations mismatch might be caused because there is something "wrong" or "anomalous" with the observations themselves. This might happen, for instance, as a consequence of noise occurring while sensing the outer environment or even because of the existence of malicious agents supplying wrong information, and outlier detection may be fruitfully exploited to identify such situations.

Clearly, the isolation of such anomalous observations offers a potentially useful source of additional knowledge that might be exploited to better support the interaction among agents or simply to optimize the way an agent pursues its goals. Indeed, outliers might be of a great interest, for they put into evidence the presence of some situations possibly requiring an alert or a quick reaction by the agent. A detailed example follows.

*Example 3.* Consider an agent $A$ that is in charge of monitoring the connectivity status of the computer network $\mathcal{N}$ shown on the right of Figure 2. The agent's background knowledge is modelled by a logic program $P_N^{\rm rls}$, which is used by $A$ for assessing whether the computer `s` is connected to `t`. Program $P_N^{\rm rls}$ (*rule component*) consists of the rules shown on the left of Figure 2. In $P_N^{\rm rls}$, each computer, say `X`, is represented by an atom `computer(X)`, and connections among computers are denoted by means of the binary relation `wired`. Moreover, for each computer `X`, the fact `down(X)` (resp. `up(X)`) encodes that `X` is offline (resp. online).

The topology of the network is encoded into the facts in $r_4$ and $r_5$. The meaning of the other rules in $P_N^{\rm rls}$ is as follows. Rule $r_1$ defines the unary predicate `down` in a way that `down(X)` is true if there exists no predecessor of the computer `X` (in the network) which is up; having a predecessor that is up is encoded in the unary predicate `predecessorUp` defined in rule $r_2$ that tells that a computer `Y` is a predecessor of `X` in the case `wired(Y, X)` holds, and therefore `PredecessorUp(X)` is true if there is a computer `Y` such that `wired(Y, X)` is true and such that `Y` is up. Rule $r_3$ says that a computer is up if it is not down. Finally, the fact in $r_6$ states that the source `s` is known to be `up`.

In order to monitor the net, $A$ observes the actual status of *each* computer `X` in the net. Hence, the agent $A$ has such an evidence modelled in the *observation component* by means of a set of facts, say $P_N^{\rm obs}$, over the predicates `down` and `up`.

It is important to note that program $P_N^{\rm rls}$ encodes the normal behavior of the network and does not explicitly account for possible anomalies. Its intended meaning is that, in usual operative conditions, a computer `X` is down if and only if there is no path in $\mathcal{N}$ connecting `s` to `X` only going through computers that are up. Thus, it is sufficient that any such path exists to have `X` be supposedly observed up. Armed with this knowledge, the agent is interested in singling out the observations that are anomalous according to the "normal" behavior of the system, modelled by $P_N^{\rm rls}$.

Assume now that, for instance, $P_N^{\rm obs}$ comprises the facts $\{\texttt{down(a)}, \texttt{up(b)}, \texttt{down(c)}, \texttt{up(d)}, \texttt{up(e)}, \texttt{up(f)}, \texttt{up(g)}, \texttt{up(h)}, \texttt{up(t)}\}$ — in the figure, the computers observed to be down are marked in bold. It is important to note that the program on the left of Figure 2 is *stratified* and, therefore, a stable model [42] always exists no matter what observations are added to it. Thus, no conflict (in the sense of classical inconsistency) may arise for the program $P_N^{\rm rls}$ and the observations in $P_N^{\rm obs}$. However, two strange things indeed come into play with the observations.

First, $\texttt{down(a)}$ comes true in $P_N^{\rm obs}$, which is not predicted by $P_N^{\rm rls}$. This observation may be clearly viewed as "unexpected" according to the background knowledge, even though it formally does not lead to an inconsistency. However, we do not have any argument for classifying it as a wrong observation. Indeed, this kind of disagreement may actually have happened for several reasons, for instance, because the background knowledge just encodes a partial description of the world and the observations involve some aspects which are not dealt with therein. E.g., the background knowledge may not encode the fact that a computer may be `down` also if some internal failure occurred, which does not depend on connectivity in the network.

In addition, there is a second strange aspect in the observations, which is the fact of `c` also being down. In this second case, however, the abnormality of this circumstance is also witnessed by some other observations. Specifically, if $A$ did not observe that computers `d`, `e`, `f`, `g`, and `t` are up, it would have inferred exactly the opposite conclusions (that are that `d`, `e`, `f`, `g`, and `t` are down) by exploiting its background knowledge, since the failure of `c` suffices for breaking the `s-t` connectivity. Thus, in this second case, there is some further evidence that the observation $\texttt{down(c)}$ is wrong. It follows that without any additional knowledge about the system, by reasoning this way and given the support of the observations on computers `d`, `e`, `f`, `g`, and `t`, a quite reliable diagnosis can be made.

In the framework that follows, the computer `c` being down is precisely recognized to represent an outlier, while the set $\mathcal{W}$ is an outlier witness, i.e., a set of facts which can be indeed explained by the rule component if and only if the outliers are not trusted in the observations. ◁

### 1.3 Comparison with Other Reasoning Tasks

In the light of the informal discussion and of the examples presented so far, one may wonder what kinds of connection outlier detection has with other well-known reasoning tasks. In particular, it should be evident that outlier detection shares some features with some well-known and studied problems in AI literature, such as *belief revision* and *diagnosis*. Outlier detection can be indeed abstractly seen as a form of revision since its ultimate goal is to logically harmonize the background theory with the observations to hand. Moreover, it can be also perceived as a form of diagnosis, because the main interest is in individuals that do not behave as predicted from the background knowledge.

Next the relationships with both belief revision and diagnosis are discussed. Specifically, given that diagnosis cames in different forms in the literature, two of the main approaches pursued in the AI community (cf. [72]) will be focused on, that are *consistency-based* and *abductive* diagnosis.

**(Consistency-Based) Diagnosis.** Diagnosis can be defined as the problem of finding what is wrong with some possibly malfunctioning systems based on knowledge about the design of that system and observations about its actual behavior (cf. [72, 71]).

In the consistency-based approach to diagnosis, there is no knowledge as to how malfunctioning occurs and manifests itself, and only the "normal" behavior is axiomatized [43, 26, 76]. Therefore, diagnosis consists in isolating components that are not consistent with all other components acting normally. Formally, there is a set $H$ of *hypothesis* (which in most of the formalizations comes in the form of abnormality-defining predicates), a background theory $T$, and a set $O$ of observations; then, the problem is to single out a set $\Delta \subseteq H$ so that $T \cup O \cup \Delta$ is consistent in the classical sense. For instance, in [76], a hypothesis $\neg ab(C)$ is introduced for each component $C$ that can possibly be faulty, and what follows from the assumptions of normality is cast into rules.

It appears that consistency-based diagnosis has a relationship with outlier detection as long as in both forms of reasoning one is interested in singling out "abnormality" in the knowledge at hand. However, the framework of outlier detection does not actually fit the diagnosis framework. First, there is an important difference in the encoding, in that, in outlier detection the background theory is not required to explicitly account for abnormal predicates and, more importantly, outliers are not required to be conflicting (in a logical sense) with the background theory, as evidenced in the previous examples.

Actually, one may be interested in assessing some more structural differences between these forms of reasoning, by assuming that these syntactic issues can be faced by means of some suitable rewriting. For instance, one can assume that the background knowledge at hand may be revised in order to explicitly model the occurrence of anomalous observations, e.g., by means of integrity constraints. Even though there are no immediate translation mechanisms between the two frameworks in general — the translation in fact depends on the semantics of the application — this may be particularly simple in some situations, such as in our network example, where the encoding of the knowledge (that currently prevents $\mathtt{up}(\mathtt{X})$ and $\mathtt{down}(\mathtt{X})$ from being both true) might be changed by adding a constraint $\leftarrow \mathtt{up}(\mathtt{X}),\ \mathtt{down}(\mathtt{X})$.

Then, it can be assumed that all the observations constitute the hypothesis for the problem, so that by consistency-based diagnosis bunches of observations conflicting with the theory can be singled out. However, at this point, the differences between the two frameworks emerge more clearly; indeed, the problem of outlier detection is not yet reduced to finding such observations that somehow "contradict" the background knowledge, because there are observations that, though different from what prescribed by the background theory, are not outliers. This has been made evident in Example 3 above. Indeed, after a proper encoding stating that whenever reachable, a computer must be up, one could derive that the observations $\mathtt{down}(\mathtt{a})$ and $\mathtt{down}(\mathtt{c})$ are both faulty. However, it has already been noticed that $\mathtt{down}(\mathtt{a})$ cannot indeed be considered an outlier because there are no other observations supporting this claim. Conversely, $\mathtt{down}(\mathtt{c})$ is an outlier, given that there are five observations (namely, $\mathtt{up}(\mathtt{d})$, $\mathtt{up}(\mathtt{e})$, $\mathtt{up}(\mathtt{f})$, $\mathtt{up}(\mathtt{g})$, and $\mathtt{up}(\mathtt{t})$) that support the fact that $\mathtt{c}$ is up.

From this perspective, outliers might abstractly be seen as faulty observations in some kinds of diagnosis whose reliability is further evidenced by the witness set, which would also have been considered faulty in the case where outliers are not thrown out. This extra criterion is responsible for greater reliability of such diagnosis, since the anomalies are not just inferred from the background knowledge (which can be in fact incomplete and lead to misclassifications, as it would be for $\mathtt{down}(\mathtt{a})$), but are further evidenced by the data in itself.

**Abduction and Diagnosis.** In the abductive approach to diagnosis [58, 59, 18] there is a description of the system to be diagnosed, observations (symptoms), and possible reasons (faults) for the observations. The aim is to single out faults that may be the actual cause for the symptoms to hand. Formally, again the sets $H$, $T$ and $O$ are given, and a subset $\Delta \subseteq H$ is sought that combined with $T$ entails $O$. Intuitively, observations in $O$ are assumed to be trustable, while the hypothesis may be revised.

Abduction was studied by Pierce [69]. Since then, it has been widely recognized as an important principle for common-sense reasoning, a powerful mechanisms for hypothetical reasoning in the presence of incomplete information, and a solid framework for modelling practical applications. Abduction has been also investigated in logic programming (see, e.g., [27] and the references therein). In this context, the most influential definition is due to Kakas and Mancarella [52], but several other approaches have been also pro-posed both from proof and model-theoretic perspectives (e.g, [64, 22, 53, 28, 31]). Moreover, extensions to settings where preferences are exploited while finding explanation (*abduction with penalization*) have been studied for both classical logics [30] and logic programming [70].

While trying to encode outlier detection in terms of abduction, the main problem is that in the former setting there is no explicit distinction between trustable observations and revisable hypothesis. Specifically, all the observations are revisable, i.e., they may be faulty, and it is not known in advance which of them should be granted, i.e., the set $O$ is generally unknown. Therefore, the translations appear to be even less immediate than for consistency-based approaches.

One possibility to build the translation might be to exploit the fact that in outlier detection it is known that singling out an outlier has the effect of justifying the witness set, which can be eventually seen as the set of trustable observations. Therefore, if this witness set, say $\mathcal{W}$, was known in advance, the problem might be encoded in a standard abduction framework by assuming all the observations but $\mathcal{W}$ to be revisable; then, explanations for $\mathcal{W}$ are in fact outliers. However, since $\mathcal{W}$ is in general not known, some efforts must be dedicated to discover it.

Again, the activity of identifying the witness sets constitutes the main source of computational complexity in outlier detection problems as well as their main distinguishing characteristics with respect to abductive problems.

**Belief Revision.** In such scenarios where an agent background theory mismatches with a set of observations, it is important to revise the background knowledge, say $T$, by incorporating that new evidence $O$ gained about the world. This process is generally known as *belief change* in the literature, and represents an active area of research in both Philosophy and Artificial Intelligence.

One of the best known theories of rational belief change is the *AGM theory of belief revision* originated by Alchourrón, Gäerdenfors, and Makinson [4] and further developed by Gäerdenfors [39] and Alchourrón and Makinson [5, 6]. Whenever $O$ is consistent with $T$, the *revision* of $T$ according to $O$, denoted by $T * O$, is the set of all the logical consequences of $T \cup \{O\}$, denoted by $T + O$. Otherwise, i.e., if $O$ is not consistent with $T$, $T * O$ is defined as $(T - \neg O) + O$, where $T - \neg O$ is the *contraction* of $T$ according to $\neg O$, that is the set of all maximal subsets of $T$ not entailing $\neg O$. This approach is the core of AGM theory, which eventually exploits a set of postulates whose aim is to characterize the intuition of *minimal change*. A number of authors have favored the postulation approach, and subsequent works concerned extensions and refinements of the basic paradigm, namely the distinction between *belief revision* and *belief update*, work on *iterated* belief revision, and the use of *epistemic states* [54, 36, 37, 25].

A criticism raised about the basic paradigm concerns the *success postulate*, i.e., the assertion that the agent believes the most recent thing he learns. Specifically, [13] noticed that this postulate is undesirable in the case when an agent observation may be itself imprecise or noisy (which is precisely the case we are interested in studying in the outlier detection framework), and proposed a model of belief revision under no obligation to incorporate observed propositions into the current belief set. Contextually, different models of *non-prioritized* belief change have been proposed in which no absolute priority is assigned to the new information due to its novelty [47]. These approaches can be grouped into three main categories: decision+revision, integrated choice, and expansion+consolidation. *Decision+revision* approaches first decide whether to accept or reject input $O$, and if the input is accepted, some of the beliefs in $T$ are given up in order to incorporate $O$ while retaining consistency. For example, *screened revision* [66] introduces a set $A$ of *core beliefs* that are kept immune from revision, and the belief set is revised only if $O$ is consistent with the set $A \cap T$. *Integrated choice* approaches perform the two above-mentioned steps simultaneously. This can be achieved by means of such choice mechanisms as *epistemic entrenchment* [39] or *spheres-based revision* [44]. Finally, *expansion+consolidation* approaches add $O$ to the belief set $T$, and then make the belief state consistent by deleting either (part of) $O$ or some original beliefs [46].

Clearly, the problem of outlier detection is related with such approaches to belief revision, given that its ultimate aim is to remove some kind of disagreement between the rule component and the observations. Specifically, since in the outlier detection setting observations are to be doubted, while taking the background knowledge for granted, two policies might be adopted in order to obtain outlier detection by revision: either a model of non-prioritized belief change might be directly exploited, or the roles of background knowledge and observations in the revision can be supposedly inverted, so that observations constitute the "initial" knowledge base and the background knowledge is used to revise it; then, the success postulate guarantees that the background theory remains unchanged and only observations are possibly given up.

However, both the above solutions strongly rely on the fact that a revision has to be made as soon as merging observations with the theory to hand leads to inconsistencies, which is in fact the starting point of any kind of belief revision. But this is not the case for outlier detection problems, where the notion of inconsistency (in the classical sense) plays no role and where the mismatch is given by a rather subtle form of disagreement between some observations and the other data at hand. Therefore, as in the case of consistency-based diagnosis, a preliminary step for carrying out outlier detection via revision is to encode the rule component in a way that explicitly accounts for the isolation of conflicts; but, this is not going to be obvious in all the circumstances.

However, the most important difficulty in the encoding lies again in the additional requirement for an outlier to be witnessed by some other set of observations. In this respect, outlier detection might be accommodated in a framework for computing some kinds of *preferred* revision, in order to avoid the revision of observations that are not outliers, as it would occur in Example 3 with the observation $\mathtt{down}(\mathtt{a})$. To this aim, if a witness set $\mathcal{W}$ had been known in advance, one would have been able to define an encoding leading to an inconsistency whenever $\mathcal{W}$ would not be entailed by the theory, so that the role of the revision would have been precisely to single out outliers having $\mathcal{W}$ as a witness set. Therefore, again, isolating the witness sets appears to be a distinguishing characteristic of outlier detection problems.

### 1.4 Contribution and Plan of the Paper

It is worth pointing out that *outlier detection* problems come in several different guises within settings that have been mainly investigated in the area of Knowledge Discovery in Databases and, recently, they have also emerged as a knowledge representation and reasoning problems, in the context of default logic [7] — refer to Section 7 for a thorough analysis of related literature.

In this paper, the definition provided in [7] is basically followed for identifying anomalies in observations, but the concept of outliers is formally stated in the context of logic programming under the stable model semantics for several reasons. Firstly, logic programs have been, in fact, proved to be a powerful tool for modelling reasoning capabilities in multi-agent systems [45, 85, 83]. Secondly, outlier detection problems formalized using the logic programming paradigm have a natural translation into standard logic inference problems, thus making the framework presented here quite easily implemented on top of any efficient inference engine (such as GnT [49], DLV [61], *Smodels* [67], and ASSAT [63]). Thirdly, the new formalization of outlier detection presented here is better suited as the basis for generalizing the outlier detection problems formalized in [7]. In more detail, the contributions of this paper are summarized below.

- ▷ The notion of outlier in the context of logic programming-based Knowledge systems is formally defined. In particular, the definition introduced in [7] is generalized by allowing an outlier to consist of a set of observations (modelled as ground facts) rather than of a single observation. This generalization is significant because there are actual situations where only non-singleton outliers can be detected.

| brave / cautious | EXISTENCE | WITNESS−CHECKING | OUTLIER−CHECKING | OW−CHECKING | COMPUTATION |
|---|---|---|---|---|---|
| General Logic Program | $\Sigma_2^P$-complete | $\Sigma_2^P$ / $D^P$-complete | $\Sigma_2^P$-complete | $D^P$-complete | $F\Sigma_2^P$-complete |
| Stratified Logic Program | NP-complete | NP-complete | NP-complete | P-complete | FNP-complete |

**Fig. 3.** Basic Results for outlier detection problems.

| brave ≡ cautious | EXISTENCE[k] | WITNESS−CHECKING[k] | OW−CHECKING[min] | COMPUTATION[min] |
|---|---|---|---|---|
| General Logic Program | $\Sigma_2^P$-complete | $D^P$-complete | $\Pi_2^P$-complete | $F\Delta_3^P$[O(log $n$)] |
| Stratified Logic Program | NP-complete | P-complete | co-NP-complete | FNP//OptP[O(log $n$)]-complete |

**Fig. 4.** Complexity of Minimum-size Outlier Detection Problems.

▷ Outlier detection problems are investigated in the context of skeptical semantics (the only one considered in [7]) as well as in the context of brave semantics. It should be noted that this does not simply add some formal details to the framework, since from the semantical viewpoint, referring to brave or skeptical reasoning in detection problems significantly changes the role that is played by outliers.

▷ The computational complexity of some natural decision outlier detection problems is thoroughly investigated for the case of *propositional* logic programs. The results of this study (both for brave and cautious reasoning) are summarized in Figure 3. It can be noted that the complexity figures range from P to $\Sigma_2^P$ depending on the specific detection problem considered. It is also worth pointing out that, differently from what happens with most logic-based reasoning frameworks, in most of cases considered brave and cautious semantics induce the *same* complexity. Furthermore, the complexity of *computing* outliers is analyzed.

▷ The *data complexity* of some basic outlier detection problems is investigated. This analysis is particularly useful in the context of outlier detection in database applications, where one is usually interested in understanding how the complexity of a problem varies as a function of the database size, and the rule component is assumed to held fixed, or anyway less frequently varied, as it usually encodes a set of constraints on the database schema. Hence, in the data complexity scenario, other than continuing investigating propositional programs, also *non-ground* logic programs are considered.

▷ Several cost-based generalizations of outlier detection problems are formalized, accounting for a number of interesting situations in which the computation of just *any* outlier is not what is really sought. Moreover, how this generalization influences the complexity of outlier detection is also studied — see Figure 4.

▷ The basic outlier detection framework assumes that observations (and, hence, outliers) come into play as sets of facts encoding some aspects of the current status of the world. However, there are situations where it would be desirable to have observations encoded as a logical theory (this might be required, for instance, for agents to be able to reconstruct an internal and possibly complex description of the outer environment by learning logical rules and then reasoning on the basis of them, e.g., about the behaviors of other agents). To this aim the concept of outlier is extended to be denoted, in general, by a set of logical rules and facts. The computational complexity of the problems arising in this extended setting is also accounted for in the paper.

▷ In order to ease fast prototyping of outlier-based reasoning frameworks, sound and complete algorithms for transforming any outlier problem into an equivalent inference problem under stable model semantics are presented. The transformations can thus be used for effectively implementing outlier detection on top of any available stable model solver (e.g., [49, 61, 67, 63]).

The rest of the paper is organized as follows. Section 2 contains some preliminaries on logic programs and on the main complexity classes dealt with in the paper. The basic definition of outliers under both brave and cautious semantics is introduced in Section 3, where the complexity of some outlier detection problems is also investigated. Section 4 proposes a generalization of the basic framework in which minimum size-outliers are sought, and studies how this additional requirement influences the basic difficulty of the detection problems. A different kind of extension, that is, the possibility of having observations encoded as logic theories is discussed in Section 5. Then, Section 6 illustrates a sound and complete rewriting for implementing outlier detection on top of stable model engines. Finally, Section 7 discusses some related work and, in Section 8, conclusions are drawn.

## 2 Preliminaries

### 2.1 Logic Programs

A *term* is a constant or a variable. An *atom* is of the form $\mathtt{p(t_1, ..., t_k)}$ where $\mathtt{p}$ is a $k$-ary predicate symbol and $\mathtt{t_1, ..., t_k}$ are terms; in the case $k = 0$, the atom is called *propositional letter* and parenthesis are omitted. A *literal* is an atom $\mathtt{a}$ or its negation $\mathtt{not\ a}$.

A *rule* $r$ is a syntactic clause of the form: $\mathtt{a} \leftarrow \mathtt{b_1, \cdots, b_k, not\ c_1, \cdots, not\ c_n.}$, where $k, n \geq 0$, and $\mathtt{a, b_1, \cdots, b_k, c_1, \cdots, c_n}$ are atoms. The atom $\mathtt{a}$, also denoted by $\mathbf{h}(r)$, is the *head* of $r$, while the conjunction $\mathtt{b_1, \ldots, b_k, not\ c_1, \cdots, not\ c_n}$, also denoted by $\mathbf{b}(r)$, is the *body* of $r$. A rule with $n = 0$ is called *positive*. A rule with an empty body (i.e. $n = k = 0$) is called a *fact* ($\leftarrow$ is omitted).

A *logic program* (short: LP) $P$ is a finite set of rules. $P$ is *positive* if all the rules are positive. $P$ is *stratified*, if there is an assignment $s(\cdot)$ of integers to the predicate symbols in $P$ such that for each rule $r$ in $P$ the following holds: if $p$ is the atom in the head of $r$ and $q$ (resp. $not\ q$) occurs in $r$, then $s(p) \geq s(q)$ (resp. $s(p) > s(q)$). Moreover, $P$ is *propositional* if all the atoms in it are propositional letters.

The *Herbrand Universe* $U_P$ of a program $P$ is the set of all constants appearing in $P$, and its *Herbrand Base* $B_P$ is the set of all ground atoms constructed from the predicates appearing in $P$ and the constants from $U_P$. A *ground* term (resp. an atom, a literal, a rule or a program) is a term (resp. an atom, a literal, a rule or a program) where no variables occur. A rule $r'$ is a *ground instance* of a rule $r$, if $r'$ can be obtained from $r$ by consistently replacing variables occurring in $r$ with constants in $U_P$. By $ground(P)$ the set of all ground instances of the rules in $P$ is denoted.

In the following, background knowledge bases encoded by means of ground programs, or, equivalently, by means of propositional logic programs[4] are (mainly) considered. While this is a rather natural setting most often adopted in the literature for introducing and discussing the complexity of various basic reasoning tasks, it is relevant to note that the proposed outlier detection framework is general enough to cope with rule components encoded also by means of non-ground programs (the reader may check that no modification at all is required in the basic definitions). This is, for instance, the case of our running examples, where the use of variables has been pursued to keep the encoding compact and to help the reader's intuition in understanding the main features of outlier detection problems. Clearly, in these cases, as far as complexity studies are concerned, given a logic program $P$ with variables, the input to our reasoning tasks is to be understood as its ground version $ground(P)$. Notably, the semantics of a program $P$ is in fact precisely defined in terms of its ground version, as discussed below.

[4] Indeed, any ground program $P$ may be equivalently seen as a propositional one, by replacing each atom of the form $\mathtt{p(t_1, ..., t_k)}$ where each $\mathtt{t_i}$ ($1 \leq i \leq k$) is a constant, with the propositional letter $\mathtt{p^{t_1, \ldots, t_k}}$.

An interpretation of $P$ is any subset of $B_P$. The truth value of a ground atom $\mathtt{L}$ w.r.t. an interpretation $I$, denoted $value_I(\mathtt{L})$, is 1 ($true$) if $\mathtt{L} \in I$ and 0 ($false$) otherwise. The value of a ground negated literal $\mathtt{not\ L}$ is $1 - value_I(\mathtt{L})$. The truth value of a conjunction of ground literals $C = \mathtt{L_1}, \ldots, \mathtt{L_n}$ is the minimum over the values of the $\mathtt{L_i}$, i.e. $value_I(C) = min(\{value_I(\mathtt{L_i}) \mid 1 \leq i \leq n\})$. If $n = 0$, then $value_I(C) = true$. A ground rule $r$ is *satisfied* by $I$ if $value_I(\mathbf{h}(r)) \geq value_I(\mathbf{b}(r))$. Thus, a rule $r$ with empty body is satisfied by $I$ if $value_I(\mathbf{h}(r)) = true$. An interpretation $M$ for $P$ is a model of $P$ if $M$ satisfies all rules in $ground(P)$.

The *minimal model semantics* assigns to a positive program $P$ its unique *minimal model* $\mathcal{MM}(P)$. A model $M$ for $P$ is minimal if no proper subset of $M$ is a model for $P$. For a general program $P$, the *stable model semantics* [41] assigns to $P$ the set $\mathcal{SM}(P)$ of its *stable models* defined as follows. Let $P$ be a logic program and let $I$ be an interpretation for $P$. Then, the reduct of $P$ w.r.t $I$, denoted by $P^I$, is the ground positive program derived from $ground(P)$ by (1) removing all rules that contain a negative literal $\mathtt{not\ a}$ in the body and $\mathtt{a} \in I$, and (2) removing all negative literals from the remaining rules. An interpretation $M$ is a stable model for $P$ if and only if $M = \mathcal{MM}(P^M)$. It is well known that stable models are minimal models and that stratified logic programs have a unique stable model (see, e.g., [24]).

Let $\mathcal{W}$ be a set of facts. Then, program $P$ *bravely entails* $\mathcal{W}$ (resp. $\neg\mathcal{W}$), denoted by $P \models_b \mathcal{W}$ (resp. $P \models_b \neg\mathcal{W}$), if *there exists* $M \in \mathcal{SM}(P)$ such that each fact in $\mathcal{W}$ is evaluated true (resp. false) in $M$. Conversely, $P$ *cautiously entails* $\mathcal{W}$ (resp. $\neg\mathcal{W}$), denoted by $P \models_c \mathcal{W}$ (resp. $P \models_c \neg\mathcal{W}$), if *for each* model $M \in \mathcal{SM}(P)$, each fact in $\mathcal{W}$ is true (resp. false) in $M$. Clearly, for a positive or stratified program $P$, $P \models_c \mathcal{W}$ iff $P \models_b \mathcal{W}$.

### 2.2 Computational Complexity

Some basic definitions about complexity theory are recalled next. The reader is referred to [68, 50] for more on this.

*Decision* problems are maps from strings (encoding the input instance over a suitable alphabet) to the set $\{$"$yes$", "$no$"$\}$. A (possibly nondeterministic) Turing machine $M$ answers a decision problem if on a given input $x$, (*i*) a branch of $M$ halts in an accepting state iff $x$ is a "yes" instance, and (*ii*) all the branches of $M$ halt in some rejecting state iff $x$ is a "no"instance.

The class P is the set of decision problems that can be answered by a deterministic Turing machine in polynomial time. The classes $\Sigma_k^P$ and $\Pi_k^P$, forming the *polynomial hierarchy*, are defined as follows: $\Sigma_0^P = \Pi_0^P = \mathrm{P}$ and for all $k \geq 1$, $\Sigma_k^P = \mathrm{NP}^{\Sigma_{k-1}^P}$, $\Delta_k^P = \mathrm{P}^{\Sigma_{k-1}^P}$, and $\Pi_k^P = \text{co-}\Sigma_k^P$ where co-$\Sigma_k^P$ denotes the class of problems whose complementary version is solvable in $\Sigma_k^P$, and where $\Sigma_k^P$ (resp. $\Delta_k^P$) models computability by a nondeterministic (resp. deterministic) polynomial-time Turing machine which may use an oracle that is, loosely speaking, a subprogram, that can be run with no computational cost, for solving a problem in $\Sigma_{k-1}^P$. The class $\Sigma_1^P$ of decision problems that can be solved by a nondeterministic Turing machine in polynomial time is also denoted by NP, while the class $\Pi_1^P$ of decision problems whose complementary problem is in NP, is denoted by co-NP. The class $D_k^P$, $k \geq 1$, is the class of problems defined as a conjunction of two independent problems, one from $\Sigma_k^P$ and one from $\Pi_k^P$, respectively. Note that, for all $k \geq 1$, $\Sigma_k^P \subseteq D_k^P \subseteq \Sigma_{k+1}^P$.

*Functions* (also *computation problems*) are (partial) maps from strings to strings, which can be computed by suitable Turing machines, called *transducers*, that have an output tape. In particular, a transducer $T$ computes a string $y$ on input $x$, if some branch of the computation of $T$ on $x$ halts in an accepting state and, in that state, $y$ is on the output tape of $T$. Thus, a function $f$ is computed by $T$, if (*i*) $T$ computes $y$ on input $x$ iff $f(x) = y$, and (*ii*) all the branches of $T$ halt in some rejecting state iff $f(x)$ is undefined.

In this paper, some classes of computation problems will be referred to which are illustrated next (see, also, [57, 79]). The class FP is the set of all the polynomial time computable functions, which are functions computed by polynomial-time bounded deterministic transducers. More generally, for each class of decision problems, say $\mathcal{C}$, F$\mathcal{C}$ denotes its functional version; for instance, FNP denotes the class of functions computed by nondeterministic transducers in polynomial time, F$\Sigma_2^P$ denotes the class of functions computed in polynomial time by nondeterministic transducers which use an NP oracle, and F$\Delta_2^P$ denotes the functions computed, in polynomial time, by a deterministic transducer which uses an NP oracle. In the following some further classes will also be referred to, which will be defined when needed.

In conclusion, the notion of reduction for decision and computation problems should be recalled. A decision problem $A_1$ is *polynomially reducible* to a decision problem $A_2$ if there is a polynomial time computable function $h$ such that for every $x$, $h(x)$ is defined and $A_1$ output "yes" on input $x$ iff $A_2$ outputs "yes" on input $h(x)$. A decision problem $A$ is *complete* for the class $\mathcal{C}$ of the polynomial hierarchy iff $A$ belongs to $\mathcal{C}$ and every problem in $\mathcal{C}$ is polynomially reducible to $A$. Moreover, a function $f_1$ is *reducible* to a function $f_2$ if there is a pair of polynomial-time computable functions $h_1, h_2$ such that, for every $x$, $h_1(x)$ is defined, and $f_1(x) = h_2(x, w)$ where $w = f_2(h_1(x))$. A function $f$ is hard for a class of functions F$\mathcal{C}$, if every $f' \in \mathcal{F}$ is polynomially reducible to $f$, and is complete for F$\mathcal{C}$, if it is hard for F$\mathcal{C}$ and belongs to F$\mathcal{C}$.

## 3 Defining Outliers

In this section, the notions and the basic definitions involved in our framework are introduced and the main problems studied in the paper are described and formalized.

### 3.1 Formal Framework

Let $P^{\text{rls}}$ be a logic program encoding general knowledge about the world, called *rule component*, and let $P^{\text{obs}}$ be a set of facts encoding some *observed* aspects of the current status of the world, called *observation component*. Then, the structure $\mathcal{P} = \langle P^{\text{rls}}, P^{\text{obs}} \rangle$, relating the general knowledge encoded in $P^{\text{rls}}$ with the evidence about the world encoded in $P^{\text{obs}}$, is a *rule-observation pair*, and it constitutes the input for outlier detection problems.

Indeed, given $\mathcal{P}$, it is interesting to identify (if there is one) a set $\mathcal{O}$ of *observations* (facts in $P^{\text{obs}}$) that are "anomalous" according to the general theory $P^{\text{rls}}$ and the other facts in $P^{\text{obs}} \setminus \mathcal{O}$. Quite roughly speaking, the idea underlying the identification of $\mathcal{O}$ is to discover a *witness set* $\mathcal{W} \subseteq P^{\text{obs}}$, that is, a set of facts which would be explained in the theory if and only if all the facts in $\mathcal{O}$ were not observed. This intuition is formalized in the following definition.

**Definition 1 (Outlier).** Let $\mathcal{P} = \langle P^{\text{rls}}, P^{\text{obs}} \rangle$ be a rule-observation pair and let $\mathcal{O} \subseteq P^{\text{obs}}$ be a set facts. Then, $\mathcal{O}$ is an *outlier* in $\mathcal{P}$ if there is a non-empty set $\mathcal{W} \subseteq P^{\text{obs}}$ with $\mathcal{W} \cap \mathcal{O} = \emptyset$, called *outlier witness* for $\mathcal{O}$ in $\mathcal{P}$, such that:

1. $P(\mathcal{P})_{\mathcal{W}} \models \neg\mathcal{W}$, and
2. $P(\mathcal{P})_{\mathcal{W},\mathcal{O}} \not\models \neg\mathcal{W}$

where $P(\mathcal{P}) = P^{\text{rls}} \cup P^{\text{obs}}$, $P(\mathcal{P})_{\mathcal{W}} = P(\mathcal{P}) \setminus \mathcal{W}$, $P(\mathcal{P})_{\mathcal{W},\mathcal{O}} = P(\mathcal{P})_{\mathcal{W}} \setminus \mathcal{O}$, and $\models$ denotes entailment under either *cautious semantics* ($\models_c$) or *brave semantics* ($\models_b$). □

As an example application of the definition above, let us consider again the network of Example 3.

*Example 4.* Let us consider the rule-observation pair $\mathcal{P}_N = \langle P_N^{\rm rls}, P_N^{\rm obs}\rangle$, where the program $P_N^{\rm rls}$ consists of the rules shown on the left of Figure 2, and $P_N^{\rm obs}$ comprises the observed facts $\{\mathtt{down(a)}, \mathtt{up(b)}, \mathtt{down(c)}, \mathtt{up(d)}, \mathtt{up(e)}, \mathtt{up(f)}, \mathtt{up(g)}, \mathtt{up(h)}, \mathtt{up(t)}\}$.

Let $\mathcal{W}$ be the set $\{\mathtt{up(d)}, \mathtt{up(e)}, \mathtt{up(g)}, \mathtt{up(f)}, \mathtt{up(t)}\}$ and $\mathcal{O} = \{\mathtt{down(c)}\}$. Then, it is easy to see that $P(\mathcal{P}_N)_\mathcal{W} \models_b \neg\mathcal{W}$ and $P(\mathcal{P}_N)_{\mathcal{W},\mathcal{O}} \not\models_b \neg\mathcal{W}$. Therefore, $\{\mathtt{down(c)}\}$ is an outlier in $\mathcal{P}_N$, and $\mathcal{W}$ is an outlier witness for $\mathcal{O}$ in $\mathcal{P}_N$ (under the *brave* semantics). Actually, since the program is stratified and there is exactly one stable model, it is the case that $\{\mathtt{down(c)}\}$ is an outlier and $\mathcal{W}$ its witness also under *cautious* semantics. $\triangleleft$

Let us now take a closer look at Definition 1. First, it is worthwhile noting that the definition is a generalization of the one proposed in [7], since an outlier is not constrained to be a literal, but it might consists of several individual facts. Accordingly, it was explicitly required that outliers must not overlap with their witness sets, in order to avoid situations where a set of facts supports by itself its anomaly. As an example of such a situation, consider the rule-observation pair $\mathcal{P}_0 = \langle P_0^{\rm rls}, P_0^{\rm obs}\rangle$, where $P_0^{\rm rls} = \{\mathtt{a} \leftarrow \mathtt{b}.\ \mathtt{b} \leftarrow \mathtt{not}\ \mathtt{c}.\}$ and $P_0^{\rm obs} = \{\mathtt{a},\ \mathtt{b},\ \mathtt{c}\}$. Consider also the sets $\mathcal{O}_0 = \{\mathtt{b},\ \mathtt{c}\}$ and $\mathcal{W}_0 = \{\mathtt{a},\ \mathtt{b}\}$. Clearly, $\mathcal{O}_0$ and $\mathcal{W}_0$ satisfy both conditions (1) and (2) in Definition 1, but they are not considered to be an outlier and its associated witness, respectively, because they are not disjoint. The problem here is that $\mathtt{b}$ appears to be an outlier only if this is witnessed by $\mathtt{b}$ itself — it is easy to check that $\mathcal{W}_0' = \{\mathtt{a}\}$ is not a witness for $\mathcal{O}_0$. Note that, situations such as the one described above cannot occur when outliers are singleton sets (cf. [7]) for which disjointness between an outlier and its associated witness is trivially guaranteed in order to satisfy condition (2) in Definition 1.

The second important feature accounted for in Definition 1 is the possibility of dealing with the two different semantics that are commonly adopted in the logic programming framework, which are, brave and the cautious semantics. Indeed, the semantics is assumed to be part of the input in the problem of outlier detection, and it is therefore fixed after a suitable entailment operator $\models$ in the set $\{\models_b, \models_c\}$ is selected. After this choice was made by the designer of the rule component, the process of singling out outliers will be carried out consistently. Obviously, if the program has a unique stable model (for instance, in the case it is positive or stratified as in Example 4), then brave and cautious semantics coincide. For this reason, in the rest of the paper, for stratified or positive programs there is no distinction among the semantics - for instance, it will be said simply that $P$ entails a set $\mathcal{W}$.

However, next is shown that in some scenarios one notion of entailment appears to be more appropriate with respect to the other, as discussed in the following two paragraphs.

**Cautious Semantics.** Let us consider again Example 2, and let us denote by $P_{DB}^{\rm obs}$ the set of facts shown in Figure 1, and by $P_{DB}^{\rm rls}$ the rule component including just the following rule:

$$\mathtt{Approved(L)} \leftarrow \mathtt{ReqLoan(L,C,A)},\ \mathtt{A} > 50.000,\ \mathtt{Endorsement(L,P)},\ \mathtt{not}\ \mathtt{Unreliable(P)}.$$

Let $\mathcal{P}_{DB}$ be the rule-observation pair $= \langle P_{DB}^{\rm rls}, P_{DB}^{\rm obs}\rangle$, and observe that the set $\{\mathtt{Unreliable}(p_1)\}$ is an outlier in $\mathcal{P}_{DB}$ whose witness is $\{\mathtt{Approved}(l_1)\}$.

Assume now that the database is updated with some new data that the Bank has collected by integrating several distributed local databases into a unique datawharehouse, and let $\bar{P}_{DB}^{\rm obs}$ be such a modified database. Data stored in different sources are not required to satisfy integrity constraints issued on the Bank schema. Therefore, after the integration is carried out, it might happen that some integrity constraints are violated. Specifically, assume for instance that the first two attributes of REQLOAN are in fact a *key* for the relation, and that the tuple REQLOAN($l_1$,$c_1$,10.000) is added to $\bar{P}_{DB}^{\rm obs}$ in the integration process, so that a conflict with REQLOAN($l_1$,$c_1$,57.000) occurs.

The standard approach in the literature for facing the presence of inconsistencies with respect to integrity constraints is to carry out some "repair" of the data [9], i.e., to identify a suitable (minimal) set of deletion/addition of facts in the database that restore the system to a consistent state. For instance, in our example, there are two possible ways for repairing the database, that are either deleting the tuple ReqLoan($l_1$,$c_1$,10.000) or deleting the tuple ReqLoan($l_1$,$c_1$,57.000) — let $R_1$ and $R_2$ be the two repairs that are computed according to such modifications. Then, whenever a query is issued over the repaired database, only *consistent answers* are retrieved, i.e., answers that are evaluated true with respect to all the possible repairs. In fact, several data integration systems supporting consistent query answering have already been proposed in the literature [38, 62, 20], which exploit a suitable encoding in terms of logic programs that ensures a one-to-one correspondence between stable models and repairs for the system.

As an example, the repair approach can be encoded by means of the program $\bar{P}^{\rm rls}_{DB}$ defined as follows:

$$\mathtt{ReqLoan'(L,C,A)} \leftarrow \mathtt{ReqLoan(L,C,A)}, \mathtt{ReqLoan(L,C,A1)}, \mathtt{A} \neq \mathtt{A1}, \mathtt{not}\ \mathtt{ReqLoan'(L,C,A1)}.$$

$$\mathtt{Approved(L)} \leftarrow \mathtt{ReqLoan'(L,C,A)},\ \mathtt{A} > \mathtt{50.000},\ \mathtt{Endorsement(L,P)},\ \mathtt{not}\ \mathtt{Unreliable(P)}.$$

where the first rule takes care of the key on ReqLoan (in particular, it ensures that the primed relation do not violate the key on ReqLoan), and the second rule is the rewriting of the original rule-component.

Then, stable models for the program $\bar{P}^{\rm rls}_{DB} \cup \bar{P}^{\rm obs}_{DB}$ are in one-to-one correspondence with repairs of the data integration system, and therefore consistent query answering coincides with cautious reasoning over the encoding, which is more appropriate than brave reasoning for data integration tasks. Accordingly, letting $\bar{\mathcal{P}}_{DB} = \langle \bar{P}^{\rm rls}_{DB}, \bar{P}^{\rm obs}_{DB} \rangle$, we have that $P(\bar{\mathcal{P}}_{DB})_{\{\mathtt{Approved}(l_1)\}} \models_c \neg\mathtt{Approved}(l_1)$, since in both the repairs $R_1$ and $R_2$, $\mathtt{Approved}(l_1)$ cannot be entailed because $p_1$ is unreliable. Moreover, $P(\bar{\mathcal{P}}_{DB})_{\{\mathtt{Approved}(l_1)\},\{\mathtt{Unreliable}(p_1)\}}$ does not cautiously entail $\neg\mathtt{Approved}(l_1)$. Indeed, at one hand, according to repair $R_2$, i.e., by deleting ReqLoan($l_1$,$c_1$,57.000), there is no need at all for deriving $\mathtt{Approved}(l_1)$ since the loan does no longer require any approval. However, at the other hand, according to repair $R_1$, i.e., by deleting ReqLoan($l_1$,$c_1$,10.000), $\mathtt{Approved}(l_1)$ can be entailed given that $\mathtt{Unreliable}(p_1)$ is being doubted about. Thus, we cannot be completely sure that $\neg\mathtt{Approved}(l_1)$ is entailed by the program $P(\bar{\mathcal{P}}_{DB})_{\{\mathtt{Approved}(l_1)\},\{\mathtt{Unreliable}(p_1)\}}$ and, therefore, $\{\mathtt{Unreliable}(p_1)\}$ and $\{\mathtt{Approved}(l_1)\}$ represent an outlier and its associated witness under the cautious semantics.

**Brave Semantics.** Let us focus again on Example 3, and consider a slight modification of the encoding in $P^{\rm rls}_N$, where the following rules have been added:

$$\begin{array}{ll}
r_7: & \mathtt{wired(c',X)} \leftarrow \mathtt{wired(c,X)}.\\
r_8: & \mathtt{wired(X,c')} \leftarrow \mathtt{wired(X,c)}.\\
r_9: & \mathtt{computer(c')}.\\
r_{10}: & \mathtt{down(c')} \leftarrow \mathtt{internalFailure(c')}.\\
r_{11}: & \mathtt{internalFailure(c')} \leftarrow \mathtt{computer(c')}, \mathtt{not}\ \mathtt{properlyWorking(c')}.\\
r_{12}: & \mathtt{properlyWorking(c')} \leftarrow \mathtt{computer(c')}, \mathtt{not}\ \mathtt{internalFailure(c')}.
\end{array}$$

encoding the fact that another computer $\mathtt{c'}$ with the same connections as $\mathtt{c}$ is added to the network and that $\mathtt{c'}$ is an old piece of hardware that is known to be subject to possibly internal faults (rules $r_{10}$...$r_{12}$).

Let $\bar{P}^{\rm rls}_N$ be such a modified rule-component, and let $\bar{\mathcal{P}}_N = \langle \bar{P}^{\rm rls}_N, P^{\rm obs}_N \rangle$, i.e., assume that the same set of observations of Example 3 has been obtained by agent $A$, which therefore has no knowledge about the current status of $\mathtt{c'}$.

Clearly, $P(\bar{\mathcal{P}}_N)$ has two possible models, corresponding to the situation in which $\mathtt{c}'$ is either up or down. In order to single out any possible anomalous situations brave semantics should be adopted, in this case. This is also often assumed in most diagnostic approaches.

Let $\mathcal{W}$ be the set $\{\mathtt{up}(\mathtt{d}), \mathtt{up}(\mathtt{e}), \mathtt{up}(\mathtt{g}), \mathtt{up}(\mathtt{f}), \mathtt{up}(\mathtt{t})\}$ and $\mathcal{O} = \{\mathtt{down}(\mathtt{c})\}$. Then, it is easy to see that $P(\bar{\mathcal{P}}_N)_{\mathcal{W}} \models_b \neg\mathcal{W}$; indeed, in the case $\mathtt{c}'$ is down, none of $\mathtt{d}$, $\mathtt{e}$, $\mathtt{f}$, $\mathtt{g}$, and $\mathtt{t}$ can be up. Thus, condition (1) in Definition 1 is satisfied. As for condition (2), it is easy to see that $P(\bar{\mathcal{P}}_N)_{\mathcal{W},\mathcal{O}} \not\models_b \neg\mathcal{W}$. Indeed, if $\mathtt{down}(\mathtt{c})$ is not trusted, then all the computers in $\mathtt{d}$, $\mathtt{e}$, $\mathtt{f}$, $\mathtt{g}$, and $\mathtt{t}$ will be entailed to be up, no matter what the status of $\mathtt{c}'$ is. Thus, under the brave semantics, $\mathtt{down}(\mathtt{c})$ remains an outlier in this modified scenario. It is worthwhile noting that according to the cautious semantics, instead, $\neg\mathcal{W}$ is not entailed by $P(\bar{\mathcal{P}}_N)_{\mathcal{W}}$, given that there is a model (where $\mathtt{c}'$ is up) in which nothing anomalous can be singled out in these observations in $\mathcal{W}$.

Therefore, brave semantics appears to be quite useful in this kind of diagnostic scenario for it allows the identification of all the situations which are possibly anomalous rather than focusing only on absolutely reliable ones.

So, there are scenarios where one of the either forms of reasoning is suitably applied. It is therefore natural to ask whether there are some particular relationships between the two notions. This section is concluded by noting that this is not the case, as can be formally verified by exploiting the asymmetry of Definition 1. Indeed, under cautious reasoning the definition is strict in requiring that for each stable model, the witness set $\mathcal{W}$ is not entailed by the theory obtained by removing $\mathcal{W}$ itself (point 1), but then it just requires the existence of a model explaining some facts in $\mathcal{W}$ after the removal of the outlier $\mathcal{O}$ (point 2). Conversely, under brave semantics the definition is loose in the first point, because it requires that $\mathcal{W}$ is true in at least one stable model, but, for point (2), it requires that each model of $P(\mathcal{P})_{\mathcal{W},\mathcal{O}}$ entails some facts in $\mathcal{W}$.

It turns out that the two semantics are both of interest, since, intuitively, referring to brave or skeptical reasoning in outlier detection problems significantly changes the role that, from a computational viewpoint, is played by literals encoding outliers — this role will be further clarified in the following sections, while discussing the complexity of outlier detection problems. Moreover, by looking at the example above, it even seems that the brave semantics better captures the intuition behind outlier detection in some kinds of diagnostic application; in this respect, extending the definition in [7] to encompass both the semantics (rather than referring to the cautious one alone) allowed us to look at outlier detection problems from a different perspective.

### 3.2 Basic Results

Now that the notion of outlier has been formalized, next the study of the most basic problems arising in this setting is addressed. The first basic problem considered here is the problem EXISTENCE defined as follows. Given in input a rule-observation pair $\mathcal{P} = \langle P^{\mathrm{rls}}, P^{\mathrm{obs}} \rangle$, EXISTENCE is the problem of deciding the existence of an outlier in $\mathcal{P}$. Clearly, the complexity of EXISTENCE strictly depends on what kind of logic program $P^{\mathrm{rls}}$ is. A very simple case is where $P^{\mathrm{rls}}$ is a positive logic program.

**Theorem 1.** *Let $\mathcal{P} = \langle P^{\mathrm{rls}}, P^{\mathrm{obs}} \rangle$ be a rule-observation pair such that $P^{\mathrm{rls}}$ is positive. Then, there are no outliers in $\mathcal{P}$.*

*Proof.* By contradiction, assume that there is an outlier $\mathcal{O} \subseteq P^{\mathrm{obs}}$ with witness $\mathcal{W} \subseteq P^{\mathrm{obs}}$ in $\mathcal{P}$. Let $P'$ denote the logic program $P(\mathcal{P})_{\mathcal{W},\mathcal{O}}$. Notice that $P'$ has a unique model, say $M$. From condition (2) in Definition 1 it is known that $P' \not\models \neg\mathcal{W}$. Hence, it can be inferred that there is $w \in \mathcal{W}$ such that $w \in M$, i.e.

that $P' \models w$. Thus, it also holds, for the monotonicity property, that $P' \cup \mathcal{O} \models w$, i.e. that $P(\mathcal{P})_{\mathcal{W}} \not\models \neg\mathcal{W}$ and thus $\mathcal{W}$ is not a witness, since it would violate condition (1) in Definition 1. □

Let us now consider a more involved scenario, in which $P^{\text{rls}}$ is stratified. Even though in logic programming adding stratified negation does not increase the complexity of identifying the unique minimal model with respect to the negation-free case, it is next shown that negation (even in the stratified form) does indeed matter in the context of outlier detection. Indeed, the `EXISTENCE` problem becomes more difficult in this case, and even unlikely to be solvable in polynomial time.

Before proving the results, some basic definitions are introduced that will be used in the proofs. Let $L$ be a set of literals. Then, it is denoted by $L^+$ the set $\{p \mid p \text{ is a positive literal in } L\}$, and by $L^-$ the set $\{p \mid \neg p \text{ is a negative literal in } L\}$. Let $L$ be a consistent set of literals. By $\sigma_L$ the truth assignment on the set of letters occurring in $L$ is denoted such that, for each literal $p \in L^+$, $\sigma_L(p) = \mathbf{true}$, and for each literal $\neg p \in L^-$, $\sigma_L(p) = \mathbf{false}$. Let $\sigma$ be a truth assignment of the set $\{x_1, \ldots, x_n\}$ of boolean variables. Then, it is denoted by $\text{Lit}(\sigma)$ the set of literals $\{\ell_1, \ldots, \ell_n\}$, such that $\ell_i$ is $x_i$ if $\sigma(x_i) = \mathbf{true}$ and is $\neg x_i$ if $\sigma(x_i) = \mathbf{false}$, for $i = 1, \ldots, n$. Let $\ell$ be a literal. Then, it is denoted by $\rho(\ell)$ the expression $\ell$, if $\ell$ is positive, and the expression $not\ \ell'$, if $\ell$ is negative and $\ell = \neg\ell'$.

**Theorem 2.** *Let $\mathcal{P} = \langle P^{\text{rls}}, P^{\text{obs}} \rangle$ be a rule-observation pair such that $P^{\text{rls}}$ is stratified. Then,* `EXISTENCE` *is* NP*-complete.*

*Proof.* (Membership) Given a rule-observation pair $\mathcal{P} = \langle P^{\text{rls}}, P^{\text{obs}} \rangle$, it must be shown that there are two sets $\mathcal{W}, \mathcal{O} \subseteq P^{\text{obs}}$ such that $P(\mathcal{P})_{\mathcal{W}} \models \neg\mathcal{W}$ (query $q'$) and $P(\mathcal{P})_{\mathcal{W},\mathcal{O}} \not\models \neg\mathcal{W}$ (query $q''$). $P(\mathcal{P})$ is stratified and, hence, has a unique stable model. Moreover, both query $q'$ and $q''$ are P-complete (see, e.g., [24]). Thus, a polynomial-time nondeterministic Turing machine can be built solving `EXISTENCE` as follows: the machine guesses both the sets $\mathcal{W}$ and $\mathcal{O}$ and then solves queries $q'$ and $q''$ in polynomial time.

(Hardness) Recall that deciding whether a Boolean formula in conjunctive normal form $\Phi = c_1 \wedge \ldots \wedge c_m$ over the variables $x_1, \ldots, x_n$ is satisfiable, i.e., deciding whether there are truth assignments to the variables making each clause $c_j$ true, is an NP-hard problem, even if each clause contains at most three distinct (positive or negated) variables, and each variable occurs in at most three clauses [40]. Without loss of generality, assume $\Phi$ contains at least one clause and one variable.

A rule-observation pair $\mathcal{P}(\Phi) = \langle P^{\text{rls}}(\Phi), P^{\text{obs}}(\Phi) \rangle$ is defined such that: (i) $P^{\text{obs}}(\Phi)$ contains exactly the fact $x_i$ for each variable $x_i$ in $\Phi$, and the facts $sat$ and $disabled$; (ii) $P^{\text{rls}}(\Phi)$ is

$$\left.\begin{array}{l} c_j \leftarrow \rho(t_{j,1}),\ not\ disabled. \\ c_j \leftarrow \rho(t_{j,2}),\ not\ disabled. \\ c_j \leftarrow \rho(t_{j,3}),\ not\ disabled. \end{array}\right\} \quad \forall 1 \leq j \leq m, \text{ s.t. } c_j = t_{j,1} \vee t_{j,2} \vee t_{j,3}$$

$$sat \leftarrow c_1,\ \ldots,\ c_m.$$

Clearly, $\mathcal{P}(\Phi)$ is stratified and can be built in polynomial time. Now it is shown that $\Phi$ is satisfiable $\Leftrightarrow$ there is an outlier in $\mathcal{P}(\Phi)$.

($\Rightarrow$) Suppose that $\Phi$ is satisfiable, and take one of its satisfying truth assignments, say $\sigma^e$, for the variables $x_1, \ldots, x_n$. Let $\mathcal{X}$ be the set $\text{Lit}(\sigma^e)^-$, and $\mathcal{X}'$ be a generic subset of $\mathcal{X}$. It is shown that $\mathcal{O} = \{disabled\} \cup (\mathcal{X} \setminus \mathcal{X}')$ is an outlier with witness $\mathcal{W} = \{sat\} \cup \mathcal{X}'$ in $\mathcal{P}(\Phi)$.
Indeed, the unique stable model of the program $P(\mathcal{P}(\Phi))_{\mathcal{W}}$ is such that each atom $c_j$ (associated to a clause) is false since $disabled$ is true, since it being not removed from $P^{\text{obs}}(\Phi)$. Hence, $sat$ cannot be entailed in $P(\mathcal{P}(\Phi))_{\mathcal{W}}$; moreover, any atom in $\mathcal{X}'$ cannot be entailed in $P(\mathcal{P}(\Phi))_{\mathcal{W}}$ also (because there is no rule suitable for the entailment) and, therefore, condition (1) in Definition 1 is satisfied. Consider, now, the set $\mathcal{O} = \{disabled\} \cup (\mathcal{X} \setminus \mathcal{X}')$. It is easy to see that an atom associated with a variable, say $x_i$, is false in $P(\mathcal{P}(\Phi))_{\mathcal{W},\mathcal{O}}$ if and only if $x_i \in \mathcal{X}$. Thus, $P(\mathcal{P}(\Phi))_{\mathcal{W},\mathcal{O}}$ has the effect of evaluating the truth value of the assignment $\sigma^e$. Given that $\sigma^e$ is a satisfying assignment, the unique stable model of $P(\mathcal{P}(\Phi))_{\mathcal{W},\mathcal{O}}$ contains $sat$, thereby satisfying condition (2) in Definition 1. Hence, $\mathcal{O}$ is an outlier in $\mathcal{P}(\Phi)$, and $\mathcal{W}$ is a witness for it.

($\Leftarrow$) Suppose that there is an outlier $\mathcal{O}$ in $\mathcal{P}(\Phi)$, and let $\mathcal{W}$ be its associated witness set. Notice that in order to satisfy condition (2) in Definition 1, $\mathcal{W}$ must contain at least a fact that can be eventually entailed by $P(\Phi)_{\mathcal{W},\mathcal{O}}$. Clearly, the only fact satisfying this requirement among those in $P^{\rm obs}(\Phi)$ is $sat$. Therefore, it must be the case that $P(\Phi)_{\mathcal{W},\mathcal{O}} \models sat$ and, consequently, $disabled$ is in $\mathcal{W} \cup \mathcal{O}$, and hence it is false in $P(\Phi)_{\mathcal{W},\mathcal{O}}$. Then, given that whenever $disabled$ is false the rule component evaluates the truth value of formula $\Phi$, it is the case that there is a satisfying assignment $\sigma^e$ such that $(\mathcal{W} \cup \mathcal{O}) \cap \{x_1, \ldots, x_n\} = {\rm Lit}(\sigma^e)^-$.

Before closing the proof, note that in the '$\Rightarrow$'-part above, the task of guessing a truth assignment is "shared" by the two sets $\mathcal{W}$ and $\mathcal{O}$. Moreover, the existence of an outlier implies ('$\Leftarrow$'-part) that $\Phi$ is satisfiable and, eventually (from the '$\Rightarrow$'-part), that there is an outlier $\mathcal{O}$ having the form $\{disabled\} \cup (\mathcal{X} \setminus \mathcal{X}')$. Thus, the following stronger claim also holds: $\Phi$ is satisfiable $\Leftrightarrow$ there is an outlier in $\mathcal{P}(\Phi)$ $\Leftrightarrow$ $\mathcal{O} = \{disabled\} \cup (\mathcal{X} \setminus \mathcal{X}')$ is an outlier with witness $\mathcal{W} = \{sat\} \cup \mathcal{X}'$ in $\mathcal{P}(\Phi)$, where $\mathcal{X} = {\rm Lit}(\sigma^e)^-$ with $\sigma^e$ being a truth assignment for the variables $x_1, \ldots, x_n$ that makes $\Phi$ true, and $\mathcal{X}'$ is a generic subset of $\mathcal{X}$. □

The complexity of the `EXISTENCE` problem in the most general setting is studied next. The following theorem shows that, under the *brave* semantics, the problem for general programs lies one level up in the polynomial hierarchy w.r.t. the complexity associated with stratified programs.

**Theorem 3.** `EXISTENCE` *under the brave semantics is* $\Sigma_2^P$*-complete.*

*Proof.* (Membership) Given a rule-observation pair $\mathcal{P} = \langle P^{\rm rls}, P^{\rm obs} \rangle$, it must be shown that there are two sets $\mathcal{W}, \mathcal{O} \subseteq P^{\rm obs}$ such that $P(\mathcal{P})_{\mathcal{W}} \models_b \neg\mathcal{W}$ (query $q'$) and $P(\mathcal{P})_{\mathcal{W},\mathcal{O}} \not\models_b \neg\mathcal{W}$ (query $q''$). Query $q'$ is NP-complete, while query $q''$ is co-NP-complete (see, e.g., [24]). Thus, a polynomial-time nondeterministic Turing machine can be built with a NP oracle, solving query `EXISTENCE` as follows: the machine guesses both the sets $\mathcal{W}$ and $\mathcal{O}$ and then solves queries $q'$ and $q''$ by two calls to the oracle.

(Hardness) Let $\Phi = \exists \mathbf{X} \forall \mathbf{Y} f$ be a quantified Boolean formula in disjunctive normal form, i.e., $f$ is a Boolean formula of the form $d_1 \vee \ldots \vee d_m$, over the variables $\mathbf{X} = x_1, \ldots x_n$, and $\mathbf{Y} = y_1, \ldots y_q$. Deciding the validity of these formulas is a well-known $\Sigma_2^P$-complete problem. Without loss of generality, assume that each disjunct $d_j$ contains three literals at most. With $\Phi$ the rule-observation pair $\mathcal{P}(\Phi) = \langle P^{\rm rls}(\Phi), P^{\rm obs}(\Phi) \rangle$ is associated such that: (i) $P^{\rm obs}(\Phi)$ contains exactly a fact $x_i$ for each variable $x_i$ in $\Phi$, and the facts $sat$ and $disabled$; (ii) $P^{\rm rls}(\Phi)$ is

$$
\begin{array}{ll}
y_i \leftarrow not\ b_i. & 1 \le i \le q \\
b_i \leftarrow not\ y_i. & 1 \le i \le q \\
sat \leftarrow \rho(t_{j,1}), \rho(t_{j,2}), \rho(t_{j,3}),\ not\ disabled. & 1 \le j \le m \text{ s.t. } d_j = t_{j,1} \wedge t_{j,2} \wedge t_{j,3}
\end{array}
$$

Clearly, $\mathcal{P}(\Phi)$ can be built in polynomial time. Now it is shown that $\Phi$ is valid $\Leftrightarrow$ there is an outlier in $\mathcal{P}(\Phi)$.

(⇒) Suppose that $\Phi$ is valid, and let $\sigma^X$ be a truth value assignment for the existentially quantified variables $\mathbf{X}$ that makes $\Phi$ valid. Next the sets $\mathcal{W}$ and $\mathcal{O}$ are built, by exploiting the ideas of the reduction in Theorem 2: Let $\mathcal{X}$ be the set $\mathrm{Lit}(\sigma^X)^-$, and $\mathcal{X}'$ be a generic subset of $\mathcal{X}$. It is shown that $\mathcal{O} = \{disabled\} \cup (\mathcal{X} \setminus \mathcal{X}')$ is an outlier with witness $\mathcal{W} = \{sat\} \cup \mathcal{X}'$ in $\mathcal{P}(\Phi)$.
Note that, since $disabled \notin \mathcal{W}$, the program $P(\mathcal{P}(\Phi))_{\mathcal{W}}$ cannot entail $sat$ (under any semantics); moreover, any atom in $\mathcal{X}'$ cannot be entailed in $P(\mathcal{P}(\Phi))_{\mathcal{W}}$ as well (because there is no rule suitable for the entailment) and, therefore, condition (1) in Definition 1 is satisfied. As for condition (2) in Definition 1, note that the stable models of the program $P(\mathcal{P}(\Phi))_{\mathcal{W},\mathcal{O}}$ are in one-to-one correspondence with the truth assignments $\sigma^Y$ of the universally quantified variables. In particular, given that $\sigma^X$ makes the formula valid, each stable model has the form $\{sat\} \cup \mathrm{Lit}(\sigma^X)^+ \cup \mathrm{Lit}(\sigma^Y)^+ \cup \{b_i \mid y_i \in \mathrm{Lit}(\sigma^Y)^-\}$. Hence, it holds that $P(\mathcal{P}(\Phi))_{\mathcal{W},\mathcal{O}} \not\models_b \neg sat$.

(⇐) Suppose that there is an outlier $\mathcal{O}$ with witness $\mathcal{W}$ in $\mathcal{P}(\Phi)$. As $sat$ is the unique fact in $P^{\mathrm{obs}}(\Phi)$ that can be derived by $P(\mathcal{P}(\Phi))_{\mathcal{W},\mathcal{O}}$, then in order to satisfy condition (2) of Definition 1, it is the case that $\mathcal{W}$ contains $sat$. Furthermore, in order to satisfy condition (1) of Definition 1, $sat$ must be not entailed in $P(\mathcal{P}(\Phi))_{\mathcal{W}}$. To this aim, either $disabled$ is in $\mathcal{W}$ or, letting $\mathcal{X} = \mathcal{W} \cap \{x_1, \ldots, x_n\}$, for any truth-value assignment $\sigma_{(\{x_1,\ldots,x_n\}\setminus\mathcal{X})\cup\neg\mathcal{X}}$, there is no assignment for the universally quantified variables satisfying the formula. Finally, in order to have $P(\mathcal{P}(\Phi))_{\mathcal{W},\mathcal{O}} \not\models_b \neg sat$, the set $\mathcal{O}$ is such that, letting $\mathcal{X}' = (\mathcal{O} \cup \mathcal{W}) \cap \{x_1, \ldots, x_n\}$, then $\sigma_{(\{x_1,\ldots,x_n\}\setminus\mathcal{X}')\cup\neg\mathcal{X}'}$ is a truth value assignment for the existentially quantified variables $\mathbf{X}$ that makes $\Phi$ valid. □

Note that even though outlier detection on general logic programs turned out to be "intrinsically" more complex than detection on stratified logic programs, the sources of the difficulties remain unchanged under brave semantics. In fact, the reduction in the above theorem points out that the task of guessing a satisfying truth assignment is still "shared" by the two sets $\mathcal{W}$ and $\mathcal{O}$, in the same way as the reduction in Theorem 2.

Next *cautious* semantics is considered. Whereas, for most reasoning tasks, switching from brave to cautious reasoning usually implies the complexity to "switch" accordingly from a certain class $C$ to the complementary class co-$C$, this is not the case for our `EXISTENCE` problem.

**Theorem 4.** `EXISTENCE` *under the cautious semantics is* $\Sigma_2^P$*-complete.*

*Proof.* (Membership) Given a rule-observation pair $\mathcal{P} = \langle P^{\mathrm{rls}}, P^{\mathrm{obs}} \rangle$, it must be shown that there are two sets $\mathcal{W}, \mathcal{O} \subseteq P^{\mathrm{obs}}$ such that $P(\mathcal{P})_{\mathcal{W}} \models_c \neg\mathcal{W}$ (query $q'$) and $P(\mathcal{P})_{\mathcal{W},\mathcal{O}} \not\models_c \neg\mathcal{W}$ (query $q''$). Query $q'$ is co-NP-complete, while query $q''$ is NP-complete (see, e.g., [24]). Thus, a polynomial-time nondeterministic Turing machine with a NP oracle can be built, solving query `EXISTENCE` as follows: the machine guesses both the sets $\mathcal{W}$ and $\mathcal{O}$ and then solves queries $q'$ and $q''$ by two calls to the oracle.

(Hardness) Let $\Phi = \exists \mathbf{X} \forall \mathbf{Y} f$ be a quantified Boolean formula in disjunctive normal form, i.e., $f$ is a Boolean formula of the form $D_1 \vee \ldots \vee D_m$, over the variables $\mathbf{X} = x_1, \ldots x_n$, and $\mathbf{Y} = y_1, \ldots y_q$. We associate with $\Phi$ the rule-observation pair $\mathcal{P}(\Phi) = \langle P^{\mathrm{rls}}(\Phi), P^{\mathrm{obs}}(\Phi) \rangle$ such that: (i) $P^{\mathrm{obs}}(\Phi)$ contains exactly a fact $x_i$ for each variable $x_i$ in $\Phi$, and the facts $unsat$ and $disabled$; (ii) $P^{\mathrm{rls}}(\Phi)$ is

| | |
|---|---|
| $y_i \leftarrow not\ b_i.$ | $1 \leq i \leq q$ |
| $b_i \leftarrow not\ y_i.$ | $1 \leq i \leq q$ |
| $sat \leftarrow \rho(t_{j,1}), \rho(t_{j,2}), \rho(t_{j,3}).$ | $1 \leq j \leq m$ s.t. $D_j = t_{j,1} \wedge t_{j,2} \wedge t_{j,3}$ |
| $unsat \leftarrow not\ sat.$ | |
| $unsat \leftarrow not\ disabled.$ | |

Clearly, $\mathcal{P}(\Phi)$ can be built in polynomial time. Now it is shown that $\Phi$ is valid ⇔ there is an outlier in $\mathcal{P}(\Phi)$.

($\Rightarrow$) Suppose that $\Phi$ is valid, and let $\sigma^X$ be a truth value assignment for the existentially quantified variables $\mathbf{X}$ that makes $\Phi$ valid. Consider the set $\mathcal{W}$ composed by the fact $unsat$ plus all the facts associated to the variables that are false in $\sigma^X$, that is the set $\{unsat\} \cup \mathrm{Lit}(\sigma^X)^-$, and consider the set $\mathcal{O}$ composed only by the fact $disabled$. Note that, the stable models of the program $P(\mathcal{P}(\Phi))_{\mathcal{W}}$ are in one-to-one correspondence with the truth assignments $\sigma_Y$ of the universally quantified variables. In particular, the unique stable model $\mathcal{M}_Y$ is given by the set $\mathrm{Lit}(\sigma^X)^+ \cup \mathrm{Lit}(\sigma^Y)^+ \cup \{b_i \mid y_i \in \mathrm{Lit}(\sigma^Y)^-\} \cup \{sat, disabled\}$. Indeed, since the formula is satisfied by $\sigma^X$, for each $\mathcal{M}_Y$, $sat \in \mathcal{M}_Y$ and $unsat \notin \mathcal{M}_Y$. Hence, $P(\mathcal{P}(\Phi))_{\mathcal{W}} \models_c \neg\mathcal{W}$. Conversely, the program $P(\mathcal{P}(\Phi))_{\mathcal{W},\mathcal{O}}$ in which $disabled$ is false, trivially derives $unsat$. It can be concluded that $\mathcal{O}$ is an outlier in $\mathcal{P}(\Phi)$, and $\mathcal{W}$ is a witness for it.

($\Leftarrow$) Suppose that there is an outlier $\mathcal{O}$ with witness $\mathcal{W}$ in $\mathcal{P}(\Phi)$. As $unsat$ is the unique fact in $P^{\mathrm{obs}}(\Phi)$ that can be derived by $P(\mathcal{P}(\Phi))_{\mathcal{W},\mathcal{O}}$, then in order to satisfy condition (2) of Definition 1, it is the case that $\mathcal{W}$ contains $unsat$ and that $P(\mathcal{P}(\Phi))_{\mathcal{W}} \models_c \neg unsat$. Furthermore, in order to satisfy condition (1) of Definition 1, $disabled$ does not belong to $\mathcal{W}$. Thus, $\{unsat\} \subseteq \mathcal{W} \subseteq \{unsat, x_1, \ldots, x_n\}$. Let $\mathcal{X}$ be the subset $\mathcal{W} \setminus \{unsat\}$ and let $\sigma^X$ be the truth value assignment $\sigma_{(\{x_1,\ldots,x_n\}\setminus\mathcal{X})\cup\neg\mathcal{X}}$ to the set of variables $\mathbf{X}$. Clearly, $P(\mathcal{P}(\Phi))_{\mathcal{W}} \models_c (\{x_1,\ldots,x_n\} \setminus \mathcal{X}) \cup \neg\mathcal{X}$. As $P(\mathcal{P}(\Phi))_{\mathcal{W}} \models_c \neg unsat$, then it is the case that for each subset $Y$ of $\mathbf{Y}$, the stable model $\mathcal{M}_Y$ of $P(\mathcal{P}(\Phi))_{\mathcal{W}}$ associated with $Y$, that is the model $\mathcal{M}_Y$ containing $Y$ and no other fact from $\mathbf{Y}$, is such that $sat \in \mathcal{M}_Y$. That is, for each truth value assignment $\sigma^Y$ to the variables in the set $Y$, there is at least a disjunct such that $\sigma^X \circ \sigma^Y$ makes the formula $f$ true. As a consequence, $\Phi$ is valid. To conclude the proof, note that $\mathcal{O} = \{disabled\}$ is always an outlier having such a witness. □

It should be pointed out in conclusion that even though the complexity of the `EXISTENCE` problem turned out to be the same for both brave and cautious semantics, the nature of the problems are still quite different. In fact, the proof of the above theorem shows that under cautious semantics the witness is alone responsible for guessing the whole satisfying assignment. Conversely, it has already been noticed that under brave semantics outlier detection requires the efforts of determining both the witness and the outliers that, in fact, both contribute to the task of guessing the satisfying assignment - see proof of Theorem 3. Intuitively, the reason for this behavior under cautious semantics lies in the fact that the condition (2) of Definition 1, i.e., $P(\mathcal{P}(\Phi))_{\mathcal{W},\mathcal{O}} \not\models_c \mathcal{W}$, just amounts at identifying a model of $P(\mathcal{P}(\Phi))_{\mathcal{W},\mathcal{O}}$ that does not entail an element of $\mathcal{W}$. Hence, outliers act just as "switches" under cautious semantics, for they solely prevent the entailment of any element of $\mathcal{W}$.

The consequences of this difference will become more evident in the next section, while discussing the complexity of other outlier detection problems. Intuitively, expect that each time the witness set is given or has its size fixed, then the outlier detection problem will have the same complexity as the general case under brave semantics, but will become easier under cautious semantics.

### 3.3 Computational Complexity of Outlier Checking Problems

In this section the complexity of some further problems related to outlier identification is studied. Specifically, given a rule-observation pair $\mathcal{P} = \langle P^{\mathrm{rls}}, P^{\mathrm{obs}} \rangle$, the following problems will be considered:

- `OUTLIER−CHECKING`: given $\mathcal{O} \subseteq P^{\mathrm{obs}}$, is $\mathcal{O}$ an outlier for some witness set $\mathcal{W}$?
- `WITNESS−CHECKING` problem: given $\mathcal{W} \subseteq P^{\mathrm{obs}}$, is $\mathcal{W}$ a witness for some outlier $\mathcal{O}$ in $\mathcal{P}$?
- `OW−CHECKING`: given $\mathcal{O}, \mathcal{W} \subseteq P^{\mathrm{obs}}$, is $\mathcal{O}$ an outlier in $\mathcal{P}$ with witness $\mathcal{W}$?

The following theorem states the complexity of the first of the problems listed above.

**Theorem 5.** *Let $\mathcal{P} = \langle P^{\mathrm{rls}}, P^{\mathrm{obs}} \rangle$ be a rule-observation pair. Then,* `OUTLIER−CHECKING` *is*

1. NP-*complete, for stratified LPs, and*
2. $\Sigma_2^P$-*complete (under both brave and cautious semantics) for general LPs.*

*Proof.*

1. As for the membership, given $\mathcal{O} \subseteq P^{\mathrm{obs}}$, let us guess a set $\mathcal{W}$ and verify that it is an outlier witness in $\mathcal{P}$ for $\mathcal{O}$. To this aim it must be verified that conditions $P(\mathcal{P})_{\mathcal{W}} \models \neg\mathcal{W}$ and $P(\mathcal{P})_{\mathcal{W},\mathcal{O}} \not\models \neg\mathcal{W}$ hold. These tasks are feasible in polynomial time since $P(\mathcal{P})$ is stratified and hence it has a unique stable model that can be computed in polynomial time. As for the hardness, the same line of reasoning as the proof of Theorem 2 is exploited, in which given a formula $\Phi$ rule-observation pair $\mathcal{P}(\Phi) = \langle P^{\mathrm{rls}}(\Phi), P^{\mathrm{obs}}(\Phi) \rangle$ has been built. It follows immediately from the proof of Theorem 2 (by letting $\mathcal{X}' = \mathcal{X}$) that $\mathcal{O} = \{disabled\}$ is an outlier $\Leftrightarrow$ the formula $\Phi$ is satisfiable.
2. As for the membership, given $\mathcal{O} \subseteq P^{\mathrm{obs}}$, let us guess a set $\mathcal{W}$ and verify that it is an outlier witness in $\mathcal{P}$ for $\mathcal{O}$. These latter tasks amount to solving an NP and a co-NP problem, as seen in Theorems 3 and 4. As for the hardness, the same reduction as Theorems 3 and 4 is exploited, in which given a formula $\Phi$ a rule-observation pair $\mathcal{P}(\Phi) = \langle P^{\mathrm{rls}}(\Phi), P^{\mathrm{obs}}(\Phi) \rangle$ has been built. It follows immediately from the proofs of Theorems 3 and 4 that $\mathcal{O} = \{disabled\}$ is an outlier $\Leftrightarrow$ the formula $\Phi$ is satisfiable. Specifically, to see why this is the case for the proof of Theorem 3, it is sufficient to let $\mathcal{X}' = \mathcal{X}$ in the '$\Rightarrow$'-part. □

Next the complexity of the `WITNESS−CHECKING` problem is studied. Interestingly, since `WITNESS−CHECKING` assumes the witness to be provided in the input, its complexity is affected by the chosen semantics, as briefly outlined in the previous section.

**Theorem 6.** *Let $\mathcal{P} = \langle P^{\mathrm{rls}}, P^{\mathrm{obs}} \rangle$ be a rule-observation pair. Then,* `WITNESS−CHECKING` *is*

1. NP-*complete, for stratified $P^{\mathrm{rls}}$,*
2. $\Sigma_2^P$-*complete under brave semantics for general $P^{\mathrm{rls}}$, and*
3. $\mathrm{D}^P$-*complete under cautious semantics for general $P^{\mathrm{rls}}$.*

*Proof.*

1. As for the membership, given $\mathcal{W} \subseteq P^{\mathrm{obs}}$, let us guess a set $\mathcal{O}$ and check that it is an outlier in $\mathcal{P}$ with witness $\mathcal{W}$. To this aim it must be verified that conditions $P(\mathcal{P})_{\mathcal{W}} \models \neg\mathcal{W}$ and $P(\mathcal{P})_{\mathcal{W},\mathcal{O}} \not\models \neg\mathcal{W}$ hold. Since $P(\mathcal{P})$ is stratified this check is feasible in polynomial time. The same reduction as Theorem 2 can be exploited, in which given a formula $\Phi$ a rule-observation pair $\mathcal{P}(\Phi) = \langle P^{\mathrm{rls}}(\Phi), P^{\mathrm{obs}}(\Phi) \rangle$ has been built. It follows immediately from the proof of Theorem 2 (by letting $\mathcal{X}' = \emptyset$) that $\mathcal{W} = \{sat\}$ is a witness for an outlier in $\mathcal{P}(\Phi)$ if and only if the formula $\Phi$ is satisfiable.
2. As for the membership, given $\mathcal{W} \subseteq P^{\mathrm{obs}}$, let us guess a set $\mathcal{O}$ and verify that it is an outlier in $\mathcal{P}$ with witness $\mathcal{W}$. These latter tasks amount to solving an NP and a co-NP problem, as seen in Theorem 3. As for the hardness,the same reduction of Theorem 3 can be exploited, in which given a formula $\Phi$ a rule-observation pair $\mathcal{P}(\Phi) = \langle P^{\mathrm{rls}}(\Phi), P^{\mathrm{obs}}(\Phi) \rangle$ has been built. It follows immediately from the proof of Theorem 3 (by letting $\mathcal{X}' = \emptyset$) that $\mathcal{W} = \{sat\}$ a witness for an outlier in $\mathcal{P}(\Phi) \Leftrightarrow$ the formula $\Phi$ is satisfiable.
3. Both conditions $P(\mathcal{P})_{\mathcal{W}} \models_c \neg\mathcal{W}$ and $P(\mathcal{P})_{\mathcal{W},\mathcal{O}} \not\models_c \neg\mathcal{W}$ have to hold for a set $\mathcal{O}$. The former condition can be checked in co-NP, whereas the latter amounts to guessing both an outlier $\mathcal{O}$ and a model for $P(\mathcal{P})_{\mathcal{W},\mathcal{O}}$, which is feasible in NP.

As for the hardness, let $\phi' = c'_1 \wedge \ldots \wedge c'_r$ be a boolean formula on the set of variables $x_1, \ldots, x_n$, $c'_k = t'_{k,1} \vee t'_{k,2} \vee t'_{k,3}$ $(1 \leq k \leq r)$, and let $\phi'' = c''_1 \wedge \ldots \wedge c''_s$ be a boolean formula on the set of variables $y_1, \ldots, y_m$, $c''_h = t''_{h,1} \vee t''_{h,2} \vee t''_{h,3}$ $(1 \leq h \leq s)$. Without loss of generality, assume that the sets $x_1, \ldots, x_n$ and $y_1, \ldots, y_m$ have no variables in common. A rule-observation pair $\mathcal{P}(\phi', \phi'') = \langle P^{\rm rls}(\phi', \phi''), P^{\rm obs}(\phi', \phi'')\rangle$ is defined such that ($i$) $P(\phi', \phi'')^{\rm obs}$ is the set $\{sat, o\}$, and ($ii$) $P(\phi', \phi'')^{\rm rls}$ is

$$
\begin{array}{ll}
x_i \leftarrow not\ a_i, o. & 1 \leq i \leq n \\
a_i \leftarrow not\ x_i, o. & 1 \leq i \leq n \\
c'_k \leftarrow \rho(t'_{k,1}), o. & 1 \leq k \leq r \\
c'_k \leftarrow \rho(t'_{k,2}), o. & 1 \leq k \leq r \\
c'_k \leftarrow \rho(t'_{k,3}), o. & 1 \leq k \leq r \\
y_j \leftarrow not\ b_j, not\ o. & 1 \leq j \leq m \\
b_j \leftarrow not\ y_j, not\ o. & 1 \leq j \leq m \\
c''_h \leftarrow \rho(t''_{h,1}), not\ o. & 1 \leq h \leq s \\
c''_h \leftarrow \rho(t''_{h,2}), not\ o. & 1 \leq h \leq s \\
c''_h \leftarrow \rho(t''_{h,3}), not\ o. & 1 \leq h \leq s \\
sat \leftarrow c'_1, \ldots, c'_r. & \\
sat \leftarrow c''_1, \ldots, c''_s. &
\end{array}
$$

Now it is shown that $\phi'$ is unsatisfiable and $\phi''$ is satisfiable $\Leftrightarrow$ there exists an outlier in $P(\phi', \phi'')$ with witness $\mathcal{W} = \{sat\}$.

($\Rightarrow$) Assume that $\phi'$ is unsatisfiable and $\phi''$ is satisfiable. It is shown that $\{o\}$ is an outlier with witness $\{sat\}$. Indeed, given that $o$ is true in $\mathcal{P}(\phi', \phi'')_{\{sat\}}$ and given the encoding of this program, there is no model $M' \in \mathcal{SM}(\mathcal{P}(\phi', \phi'')_{\{sat\}})$ such that $sat \in M'$, because $\phi'$ is not satisfiable. Thus, $\mathcal{P}(\phi', \phi'')_{\{sat\}} \models_c \neg\ sat$ holds and condition (1) in Definition 1 is satisfied. Similarly, given that $o$ is false in $\mathcal{P}(\phi', \phi'')_{\{sat\},\{o\}}$ and given the encoding of the program, there is a model $M'' \in \mathcal{SM}(\mathcal{P}(\phi', \phi'')_{\{sat\},\{o\}})$ such that $sat \in M''$, because $\phi''$ is satisfiable. Thus, $\mathcal{P}(\phi', \phi'')_{\{sat\},\{o\}} \not\models_c \neg\ sat$ holds, and condition (2) in Definition 1 is satisfied as well.

($\Rightarrow$) Assume that there is an outlier $\mathcal{O}$ in $P(\phi', \phi'')$ with witness $\mathcal{W} = \{sat\}$. It is shown that $\phi'$ is unsatisfiable and $\phi''$ is satisfiable. Indeed, notice beforehand that the only set candidate to be an outlier is $\{o\}$, because $sat$ and $o$ are the only observations at hand. Then, because of Definition 1: (1) there is no model $M' \in \mathcal{SM}(\mathcal{P}(\phi', \phi'')_{\{sat\}})$ such that $sat \in M'$, and (2) there is a model $M'' \in \mathcal{SM}(\mathcal{P}(\phi', \phi'')_{\{sat\},\{o\}})$ such that $sat \in M''$. Given the encoding of the programs above and the fact that $o$ is true in $\mathcal{P}(\phi', \phi'')_{\{sat\}}$ while it is false in $\mathcal{P}(\phi', \phi'')_{\{sat\},\{o\}}$, we conclude from (1) that $\phi'$ is not satisfiable, and from (2) that $\phi''$ is satisfiable. □

Finally, the next theorem provides the complexity for the `OW−CHECKING` problem, in which it is simply checked whether two given sets $\mathcal{O}$ and $\mathcal{W}$ are an outlier and a witness for it, respectively. Notice that problem `OW−CHECKING` is relevant as it may constitute the basic operator to be implemented in a system of outlier detection. Interestingly, in this case, the complexity does not depend on the semantics.

**Theorem 7.** *Let $\mathcal{P} = \langle P^{\rm rls}, P^{\rm obs}\rangle$ be a rule-observation pair. Then,* `OW−CHECKING` *is*

1. P*-complete, for stratified $P^{\rm rls}$, and*
2. D$^P$*-complete (under both brave and cautious semantics) for general $P^{\rm rls}$.*

*Proof.*

1. As for the membership, given $\mathcal{W}, \mathcal{O} \subseteq P^{\text{obs}}$, let us check in polynomial time that $P(\mathcal{P})_{\mathcal{W}} \models \neg \mathcal{W}$ and that $P(\mathcal{P})_{\mathcal{W},\mathcal{O}} \not\models \neg \mathcal{W}$. As for the hardness, given a stratified logic program $P$ and an atom $a$, the P-complete problem of deciding whether $P \models \neg\, a$ is reduced to `OW−CHECKING`. Consider the rule-observation pair $\mathcal{P}$ with $P^{\text{rls}} = P \cup \{a \leftarrow \mathit{not}\ b\}$ and $P^{\text{obs}} = \{a, b\}$, where $b$ is a new propositional letter not occurring in $P$. It is shown that $P \models \neg a \Leftrightarrow \{a\}$ is an outlier witness for $\{b\}$ in $\mathcal{P}$.
   - ($\Rightarrow$) Assume that $P \models \neg a$. Then, $\{a\}$ is an outlier witness for $\{b\}$ in $\mathcal{P}$. Indeed, consider the program $P(\mathcal{P})_{\{a\}}$, and notice that $a$ cannot be entailed because of the assumption and of the fact that the body of the rule $a \leftarrow \mathit{not}\ b$ is false because $b$ is true in $P(\mathcal{P})_{\{a\}}$. Thus, condition (1) in Definition 1 holds. As for condition (2), consider the program $P(\mathcal{P})_{\{a\},\{b\}}$ and notice that it entails $a$, precisely because of the rule $a \leftarrow \mathit{not}\ b$.
   - ($\Leftarrow$) Assume that $\{a\}$ is an outlier witness for $\{b\}$ in $\mathcal{P}$. To conclude that $P \models \neg a$, it is sufficient to consider condition (1) in Definition 1. Indeed, it must be the case that $P(\mathcal{P})_{\{a\}}$ does not entail $a$. Given that the rule $a \leftarrow \mathit{not}\ b$ cannot be used for entailing $a$ (because $b$ is true in $P(\mathcal{P})_{\{a\}}$), it can be concluded that $a$ cannot be entailed by $P$.
2. As for the membership, given $\mathcal{W}, \mathcal{O} \subseteq P^{\text{obs}}$, conditions $P(\mathcal{P})_{\mathcal{W}} \models \neg \mathcal{W}$ and $P(\mathcal{P})_{\mathcal{W},\mathcal{O}} \not\models \neg \mathcal{W}$ can be checked respectively in co-NP and NP under cautious semantics, and respectively in NP and co-NP under brave semantics, hence in both cases the conjunction of a NP and a co-NP problem has to be answered, that is a $D^P$ problem. As for the hardness the proof is analogous to Point 3 of Theorem 6, by letting $\mathcal{W} = \{sat\}$ and $\mathcal{O} = \{o\}$. □

Up to this point, attention has been focused on outlier decision problems. Turning to outlier computation problems, the following result can be established by noticing that in the proofs of the `EXISTENCE` problem (Theorems 2, 3 and 4), solving a satisfiability problem is reduced to computing an outlier.

**Corollary 1.** *Let $\mathcal{P} = \langle P^{\text{rls}}, P^{\text{obs}} \rangle$ be a rule-observation pair. Then, the* `COMPUTATION` *problem, i.e., computing an arbitrary outlier in $\mathcal{P}$, is (i)* FNP*-complete, for stratified rule components, and (ii)* $\mathrm{F}\Sigma_2^P$*-complete, for general rule components.*

### 3.4 Data Complexity of Outlier Problems

In all the complexity results derived so far, a setting has been considered in which both the rule component and the observation component are part of the outlier detection problem input. In order to have a more complete picture, in this section the *data* complexity of these problems is investigated. That is, a fixed rule component is considered and a set of observations are, instead, provided as the input. Such a kind of analysis may be useful in the context of database applications, where one is usually interested in understanding how the complexity of a problem varies as a function of the database size (cf. [84]). In this respect, the analysis becomes more interesting if also non-ground programs are considered. Therefore, in the following, both the ground and non-ground settings are considered.

**Ground programs.** Some further computational complexity notions are now recalled. Let $C$ be a boolean circuit. The *size* of $C$ is the total number of its gates. The *depth* of $C$ is the number of gates in the longest path from any input to any output in $C$. A family $\{C_i\}$ of boolean circuits, where $C_i$ accepts strings of size $i$, is *uniform* if there exists a Turing machine $T$ that, on input $i$, produces the circuit $C_i$. $\{C_i\}$ is *logspace uniform* if $T$ carries out its task using $O(\log i)$ space. Then, $\mathrm{AC}^0$ is the class of decision problems solved by logspace uniform families of circuits of polynomial size and constant depth, with AND, OR, and NOT gates of unbounded fan-in.

It is not difficult to see that, handling ground programs, all the basic outlier detection problems have their data complexity lying in $\mathrm{AC}^0$. Indeed, let $A$ denote the set of propositional letters occurring in the rule component. Under the data complexity measure, the rule component is fixed, while the input of the problem consists of the observation component. Nonetheless, it can be shown that the complexity of outlier detection problems on ground programs is independent of the observation component out of letters in $A$. Indeed, being the rule component fixed, if $\mathcal{O}$ is an outlier with associated witness $\mathcal{W}$, then $\mathcal{O} \cap A$ is an outlier as well with associated witness $\mathcal{W} \cap A$. Thus, given a pair $\mathcal{P} = \langle P^{\mathrm{rls}}, P^{\mathrm{obs}} \rangle$, it holds that there is an outlier in $\mathcal{P}$ iff there is an outlier in $\mathcal{P}_e = \langle P^{\mathrm{rls}}, P^{\mathrm{obs}} \cap A \rangle$. Once the pair $\mathcal{P}_e$ is available, conditions 1 and 2 of Definition 1 can be tested using a constant amount of time as the number of outlier/witness pairs in $\mathcal{P}_e$ is upper bounded by $2^{2n}$ and the number of models of $\mathcal{P}_e$ is upper bounded by $2^n$, where $n$ denotes the number of letters in the set $A$, which is a constant. Both the latter and the former tasks can be solved in $\mathrm{AC}^0$. Thus, the following theorem holds.

**Theorem 8.** *Let $\mathcal{P} = \langle P^{\mathrm{rls}}, P^{\mathrm{obs}} \rangle$ be a rule-observation pair such that $P^{\mathrm{rls}}$ is a fixed propositional logic theory. Then,* `EXISTENCE`*,* `OUTLIER−CHECKING`*,* `WITNESS−CHECKING`*, and* `OW−CHECKING` *are in* $\mathrm{AC}^0$.

As a consequence, outlier detection problems under the data complexity measure can be solved in polynomial time and are highly-parallelizable.

**Non-ground programs.** The investigation of the data complexity for this setting can be carried out, by preliminary putting into evidence the features it shares with the ground case when both rule and observation components are part of the input. The basic idea is that the reductions used in proving complexity results for outlier detection problems (carried out via encodings into the rule component which is, therefore, required to be part of the input) can be "simulated" by some *fixed* non-ground encoding. In particular, the instantiation of such non-ground encodings can be made in such a way to simulate the encodings exploited in the case of ground programs. Let us consider, for instance, the data complexity for problem `EXISTENCE` under the cautious semantics.

**Theorem 9.** *Let $\mathcal{P} = \langle P^{\mathrm{rls}}, P^{\mathrm{obs}} \rangle$ be a rule-observation pair such that $P^{\mathrm{rls}}$ is a fixed general logic program. Then* `EXISTENCE` *under the cautious semantics is $\Sigma_2^P$-complete.*

*Proof.* (Membership) Given a fixed general logic program $P^{\mathrm{rls}}$ and a set of ground facts $P^{\mathrm{obs}}$, it must be shown that there are two disjoint sets $\mathcal{W}, \mathcal{O} \subseteq P^{\mathrm{obs}}$ such that $P(\mathcal{P})_{\mathcal{W}} \models \neg\mathcal{W}$ (query $q'$) and $P(\mathcal{P})_{\mathcal{W},\mathcal{O}} \not\models \neg\mathcal{W}$ (query $q''$). Recall that the complexity of the entailment problem for general propositional logic programs is co-NP-complete. Thus, a polynomial-time nondeterministic Turing machine with an NP oracle can be built solving `EXISTENCE` as follows: the machine guesses both the sets $\mathcal{W}$ and $\mathcal{O}$, computes the propositional logic programs $ground(P(\mathcal{P})_{\mathcal{W}})$ and $ground(P(\mathcal{P})_{\mathcal{W},\mathcal{O}})$ – this task can be done in polynomial time since the size of these programs is polynomially related to the size of $P^{\mathrm{obs}}$, and then solves queries $q'$ and $q''$ by two calls to the oracle.

(Hardness) Let $\Phi = \exists\mathbf{X}\forall\mathbf{Y} f$ be a quantified Boolean formula in disjunctive normal form, i.e., $f$ is a Boolean formula of the form $d_1 \vee \ldots \vee d_m$, over the variables $\mathbf{X} = x_1, \ldots x_n$, and $\mathbf{Y} = y_1, \ldots y_q$. With $\Phi$ the following set of facts $P^{\mathrm{obs}}(\Phi)$ is associated:

$$
\begin{array}{ll}
o_1 : unsat. & \\
o_2 : disabled. & \\
o_{3,k} : variable\exists(x_k). & 1 \leq k \leq n \\
o_{4,i} : variable\forall(y_i, y_{(i+1) \bmod (q+1)}). & 1 \leq i \leq q \\
o_{5,j} : disjunct(d_j, d_{(j+1) \bmod (m+1)}, \wp(t_{j,1}), \ell(t_{j,1}), \wp(t_{j,2}), \ell(t_{j,2}), \wp(t_{j,3}), \ell(t_{j,3})). & 1 \leq j \leq m
\end{array}
$$

where $d_j = t_{j,1} \wedge t_{j,2} \wedge t_{j,3}$, $1 \leq j \leq m$, $\ell(t)$ denotes the atom occurring in the literal $t$, and $\wp(t)$ is the constant $pos$, if $t$ is a positive literal, and the constant $neg$, if $t$ is a negative literal. Intuitively, the atoms $o_{3,k}$, $o_{4,i}$, and $o_{5,j}$ together provide an encoding of the formula $\Phi$. Such an encoding will be exploited by the subsequent rule part (see below) in order to evaluate the truth value of the formula $\Phi$. In particular, each atom $o_{3,k}$ is associated to a distinct existentially quantified variables, each atom $o_{4,i}$ is associated to a distinct universally quantified variable, while each atom $o_{5,j}$ is associated to a distinct disjunct $d_j$ occurring in to the formula $\Phi$. As for the atoms $unsat$ and $disabled$, they have the same role played in Theorem 4.

The rule part of the pair is composed by the following fixed general logic program $P^{\mathrm{rls}}$:

$$\begin{cases} r_0 : disjunctTrue \leftarrow disjunct(_,_,pos,X_1,pos,X_2,pos,X_3), \\ \qquad\qquad variable\exists(X_1), variable\exists(X_2), variable\exists(X_3). \\ r_1 : disjunctTrue \leftarrow disjunct(_,_,neg,X_1,pos,X_2,pos,X_3), \\ \qquad\qquad not\ variable\exists(X_1), variable\exists(X_2), variable\exists(X_3). \\ \vdots \\ r_7 : disjunctTrue \leftarrow disjunct(_,_,neg,X_1,neg,X_2,neg,X_3), \\ \qquad\qquad not\ variable\exists(X_1), not\ variable\exists(X_2), not\ variable\exists(X_3). \\ r_8 : disjunctTrue \leftarrow disjunct(_,_,pos,Y_1,pos,X_2,pos,X_3), \\ \qquad\qquad variable\forall True(Y_1), variable\exists(X_2), variable\exists(X_3). \\ \vdots \\ r_{63} : disjunctTrue \leftarrow disjunct(_,_,neg,Y_1,neg,Y_2,neg,Y_3), \\ \qquad\qquad not\ variable\forall True(Y_1), not\ variable\forall True(Y_2), not\ variable\forall True(Y_3). \end{cases}$$

$$\begin{cases} r_{64} : variable\forall True(Y) \leftarrow variable\forall(Y,_), not\ variable\forall False(Y). \\ r_{65} : variable\forall False(Y) \leftarrow variable\forall(Y,_), not\ variable\forall True(Y). \end{cases}$$

$$\begin{cases} r_{66} : disjunct(d_0,d_1,pos,x_0,pos,x_0,pos,x_0). \\ r_{67} : variable\exists(x_0). \\ r_{68} : variable\forall(y_0,y_1). \\ r_{69} : unsound \leftarrow disjunct(C1,C2,_,_,_,_,_,_), not\ disjunctIN(C2). \\ r_{70} : disjunctIN(C2) \leftarrow disjunct(C2,_,_,_,_,_,_,_). \\ r_{71} : unsound \leftarrow variable\forall(Y1,Y2), not\ variable\forall IN(Y2). \\ r_{72} : variable\forall IN(Y2) \leftarrow variable\forall IN(Y2,_). \\ r_{73} : sound \leftarrow not\ unsound. \end{cases}$$

$$\begin{cases} r_{74} : sat \leftarrow sound, disjunctTrue. \\ r_{75} : unsat \leftarrow not\ sat. \\ r_{76} : unsat \leftarrow not\ disabled. \end{cases}$$

Next, some comments on the rules composing the program $P^{\mathrm{rls}}$ are provided.

Rules $r_0 \ldots r_{63}$ are introduced in order to compute the truth value of the disjuncts composing the formula $\Phi$. Indeed, in the body of each of these rules, there is exactly one atom $o_{5,j}$, encoding a generic disjunct $d_j$ in $\Phi$, and three atoms evaluating the truth value of the three literals occurring in to the disjunct. Notice that, 64 rules are needed in order to represent all the possible schemes of disjuncts occurring into a $\Phi$ formula, that is, all the possible conjunctions of three boolean variables, either negated or positive, of two distinct types, i.e., either existentially or universally quantified.

Rules $r_{64}$ and $r_{65}$ serve the purpose to guess a possible truth value assignment to the universally quantified variables in $\Phi$. Intuitively, being the ground atom $variable\forall True(y_i)$ ($variable\forall False(y_i)$ resp.) true in a model of the overall program, means the corresponding universally quantified variable $y_i$ is intended to be true (false resp.) in a suitable truth value assignment to the variables of $\Phi$ which is encoded by the model.

Rules $r_{69} \ldots r_{73}$ prevent that the atoms $o_{4,i}$ and $o_{5,j}$ of the observation part, associated respectively to the universally quantified variables $y_i$ and to the disjuncts $d_j$ of $\Phi$, belong to some outlier or witness of the overall pair. Indeed, if some of these atoms is removed from the observation component, then the rule component will not evaluate $\Phi$ correctly. To this aim, the rule $r_{71}$ ($r_{69}$ resp.) will imply the atom $unsound$ whenever an atom $o_{4,i}$ ($o_{5,j}$ resp.) is present in the model while the subsequent atom $o_{4,i+1}$ ($o_{5,j+1}$ resp.) is not. This check can be accomplished since each atom $o_{4,i}$ ($o_{5,j}$ resp.) carry both the constant $y_i$ ($d_j$ resp.) associated to the variable $y_i$ (the disjunct $d_j$ resp.) and the constant $y_{i+1}$ ($d_{j+1}$ resp.) associated to the subsequent variable $y_{i+1}$ (disjunct $d_{j+1}$ resp.). Thus, loosely speaking, universally quantified variables and disjuncts of the formula are chained together so that either none of them or all together can be removed without entailing $unsound$. In order to prevent the latter possibility, a dummy disjunct (see rule $r_{66}$), which evaluates always true since it contains the dummy existentially quantified variable $x_0$ which is always assumed to be true in its turn (see rule $r_{67}$), and a dummy universally quantified variable $y_0$ (see rule $r_{68}$) are introduced in the rule part, referring respectively to the first disjunct $d_1$ and to the first universally quantified variable $y_1$ of $\Phi$. Since both $r_{66}$ and $r_{68}$ cannot be removed for sure from the overall pair, then they guarantee that $unsound$ is true in a model of the overall program if and only if at least an observation $o_{4,i}$ or $o_{5,j}$ is removed from the observations.

Finally, rules $r_{74} \ldots r_{76}$ have the same purpose of the corresponding rules in the reduction described in Theorem 4, the only difference being now that in order to entail $sat$ the rule $r_{74}$ requires the formula is syntactically sound other than evaluating true.

It is now clear that the rest of the theorem will follow a line of reasoning analogous to that of Theorem 4. For completeness, we continue completing the formal proof.

Now it is shown that $\Phi$ is valid $\Leftrightarrow$ there is an outlier in $\mathcal{P}(\Phi) = \langle P^{\text{rls}}, P^{\text{obs}} \rangle$.

($\Rightarrow$) Suppose that $\Phi$ is valid, and let $\sigma^X$ be a truth value assignment for the existentially quantified variables $\mathbf{X}$ that satisfies $f$. Consider the set $\mathcal{W}$ composed by the fact $unsat$ plus all the facts $variable\exists(x_i)$ associated to the variables that are false in $\sigma^X$, that is the set $\{unsat\} \cup \{variable\exists(x) \mid x \in \text{Lit}(\sigma^X)^-\}$, and consider the set $\mathcal{O}$ composed only by the fact $disabled$. Note that the stable models $\mathcal{M}_Y$ of the program $P(\mathcal{P}(\Phi))_{\mathcal{W}}$ are in one-to-one correspondence with the truth assignments $\sigma_Y$ of the universally quantified variables (consider rules $r_{64}$ and $r_{65}$). Now, since the formula is satisfied by $\sigma^X$, for each $\mathcal{M}_Y$, $sat \in \mathcal{M}_Y$ and $unsat \notin \mathcal{M}_Y$. Hence, $P(\mathcal{P}(\Phi))_{\mathcal{W}} \models_c \neg\mathcal{W}$. Conversely, the program $P(\mathcal{P}(\Phi))_{\mathcal{W},\mathcal{O}}$ in which $disabled$ is false, trivially derives $unsat$. It can be concluded that $\mathcal{O}$ is an outlier in $\mathcal{P}(\Phi)$, and $\mathcal{W}$ is a witness for it.

($\Leftarrow$) Suppose that there is an outlier $\mathcal{O}$ with witness $\mathcal{W}$ in $\mathcal{P}(\Phi)$. As $unsat$ is the unique fact in $P^{\text{obs}}(\Phi)$ that can be derived by $P(\mathcal{P}(\Phi))_{\mathcal{W},\mathcal{O}}$, then in order to satisfy condition (2) of Definition 1, it is the case that $\mathcal{W}$ contains $unsat$. Furthermore, in order to satisfy condition (1) of Definition 1, $disabled$ does not belong to $\mathcal{W}$. Now, it is shown that $\{unsat\} \subseteq \mathcal{W} \subseteq \{unsat, variable\exists(x_1), \ldots, variable\exists(x_n)\}$. Since $unsat$ must be false in $P(\Phi)_{\mathcal{W}}$, because of condition (1) in Definition 1, it can be observed that $P(\Phi)_{\mathcal{W}}$ entails $sat$ and, thus, by rules $r_{73}$ and $r_{74}$, it does not entail $unsound$. Assume now, for the sake of contradiction, that a fact $o_{4,i}$ ($o_{5,j}$ resp.) belongs to $\mathcal{W}$. Then, program $P(\Phi)_{\mathcal{W}}$ entails $unsound$, which is impossible — see the general comments about the reduction reported above.
Let $\mathcal{X}$ be the subset $\{variable\exists(x) \mid x \in (\mathcal{W} \setminus \{unsat\})\}$ and let $\sigma^X$ be the truth value assignment $\sigma_{(\{x_1,\ldots,x_n\}\setminus\mathcal{X})\cup\neg\mathcal{X}}$ to the set of variables $\mathbf{X}$. Clearly, $P(\mathcal{P}(\Phi))_{\mathcal{W}} \models_c (\{variable\exists(x_1), \ldots, variable\exists(x_n)\} \setminus \mathcal{X}) \cup \neg\mathcal{X}$. Furthermore, as $P(\mathcal{P}(\Phi))_{\mathcal{W}} \models_c \neg unsat$, then it is the case that for each subset $Y$ of $\mathbf{Y}$, the stable model $\mathcal{M}_Y$ of $P(\mathcal{P}(\Phi))_{\mathcal{W}}$ associated with $Y$, that is the model $\mathcal{M}_Y$ containing $\{variable\forall True(y) \mid y \in Y\}$ and no other fact of the same predicate, is such that $sat \in \mathcal{M}_Y$. That is, for each truth value assignment $\sigma^Y$ to the variables in the set $Y$, there is at least a disjunct such that $\sigma^X \circ \sigma^Y$ makes the formula $f$ true. As a consequence, $\Phi$ is valid. To conclude the proof, note that $\mathcal{O} = \{disabled\}$ is always an outlier having such a witness. □

Theorem 9 depicts a strategy to reformulate the reduction illustrated in previous Theorem 4 in terms of a reduction exploiting a non-ground rule-observation pair whose rule component is kept fixed. Notably, this strategy can be also adopted for all the other reductions exploited in the proofs presented so far. For instance, we leave to the careful reader to check that, by adapting the same line of reasoning to Theorems 2 and 3, the following results concerning outlier detection problems under the data complexity measure can be eventually obtained.

**Theorem 10.** *Let $\mathcal{P} = \langle P^{\rm rls}, P^{\rm obs}\rangle$ be a rule-observation pair such that $P^{\rm rls}$ is a fixed stratified logic program. Then* `EXISTENCE` *is* NP*-complete.*

**Theorem 11.** *Let $\mathcal{P} = \langle P^{\rm rls}, P^{\rm obs}\rangle$ be a rule-observation pair such that $P^{\rm rls}$ is a fixed general logic program. Then* `EXISTENCE` *under the brave semantics is $\Sigma_2^P$-complete.*

## 4 Minimum-size Outlier Detection

### 4.1 Extending the Framework

There are several practical situations in which the computation of just *any* outlier is not what is really sought. Consider, for instance, the short story in the Introduction, and recall that *Nino* is considered to be an outlier by his Finnish colleagues because he is the *only* person with black hair and brown eyes among the other people working for the team. This assumption appears quite reasonable, given our intuition of outliers being individuals whose behavior is deviant w.r.t. the "normal" one. However, it is not the only possible one; indeed, by changing the roles played by *Nino* and the other team members, it might be concluded that all the team members but *Nino* are in fact outliers and that *Nino* having black hair and brown eyes is the associated witness. If asked to assess the correctness of the latter conclusion, most of us would certain disagree with it, because we implicitly associate the notion of normality with the characteristics embodied by the majority of the observations at hand, and for we are inclined to label an individual as anomalous precisely because it is rare in the observations.

Therefore, it is sometimes natural to constrain the basic notion of outliers, formally defined in Definition 1, in order to account for some criteria aiming at singling out outliers of *minimum size*. In this section, the outlier detection problem will be studied with the additional constraint of minimizing the outlier size. It is worthwhile noting that this setting differs with what is generally required by the principle of minimal diagnosis [76], where minimality according to set inclusion rather that to cardinality is often considered. In fact, the inclusion based minimality criterion does not generally prevent the possibility of having outliers involving the majority of the observations at hand as, e.g., in the above reported example. As a further (extreme) example, consider the rule-observation pair $\mathcal{P}_1 = \langle P_1^{\rm rls}, P_1^{\rm obs}\rangle$, where $P_1^{\rm rls} = \{\mathtt{a} \leftarrow \mathtt{not}\ \mathtt{o_1}, ..., \mathtt{not}\ \mathtt{o_n}.\ \mathtt{b} \leftarrow \mathtt{not}\ \mathtt{o}.\}$ and $P_1^{\rm obs} = \{\mathtt{o_1}, ..., \mathtt{o_n},\ \mathtt{o}\}$. Then, there are two minimal outliers in $\mathcal{P}_1$: $\{\mathtt{o_1}, ..., \mathtt{o_n}\}$ whose associated witness is $\{\mathtt{a}\}$, and $\{\mathtt{o}\}$ whose associated witness is $\{\mathtt{b}\}$. Clearly, singling out the former outlier may be undesirable in several situations, since it includes all but one observation and therefore fails in conveying information about abnormality with respect to the observed population; the latter outlier, instead, seems to better reflect the intuition beyond outlier detection problems.

A first natural problem that arises in this setting is to decide about the existence of outliers of bounded size. For instance, it may be interesting to decide whether there are outliers consisting at most the 5% of the observations or whether there are outliers consisting of one individual only. Actually, it is next shown that bounding the size of the outliers we are looking for does not affect the complexity of the `EXISTENCE` problem.

**Theorem 12.** *Given in input a rule-observation pair $\mathcal{P} = \langle P^{\rm rls}, P^{\rm obs} \rangle$, and a natural number $k$, the* $\texttt{EXISTENCE}[\texttt{k}]$ *problem of deciding the existence of outlier $\mathcal{O}$ of size at most $k$ ($|\mathcal{O}| \leq k$) in $\mathcal{P}$ is*

1. NP-*complete, for stratified $P^{\rm rls}$, and*
2. $\Sigma_2^P$*, for general $P^{\rm rls}$.*

*Proof.* For the membership it is sufficient to observe that the membership parts in Theorems 2, 3 and 4 can be modified by verifying that the guessed outlier $\mathcal{O}$ has size at most $k$. Such a test is feasible in polynomial time, and hence does not affect the complexity of the algorithms. As for the hardness it is sufficient to observe that in the proof of the theorems above, if $\{disabled\}$ is an outlier, then the formula is satisfied. □

Similarly, an analogous version of the problem $\texttt{WITNESS}-\texttt{CHECKING}$ can be formalized: given $\mathcal{W} \subseteq P^{\rm obs}$ and a fixed natural number $k$, is $\mathcal{W}$ a witness for any outlier $\mathcal{O}$ in $\mathcal{P}$, such that $|\mathcal{O}| \leq k$? This problem will be called $\texttt{WITNESS}-\texttt{CHECKING}[\texttt{k}]$.

Interestingly, this time bounding the size of outlier indeed influences the complexity associated with the problem. In fact, for general LPs it becomes D$^P$-complete even under brave semantics (rather than $\Sigma_2^P$-complete), and for stratified LPs it becomes even feasible in polynomial time. This is shown in the following theorem.

**Theorem 13.** *Let $\mathcal{P} = \langle P^{\rm rls}, P^{\rm obs} \rangle$ be a rule-observation pair. Then,* $\texttt{WITNESS}-\texttt{CHECKING}[\texttt{k}]$ *is*

1. P-*complete, for stratified $P^{\rm rls}$, and*
2. *$D^P$-complete (under both brave and cautious semantics) for general $P^{\rm rls}$.*

*Proof.* 1. As for the membership, given $\mathcal{W} \subseteq P^{\rm obs}$, it has to be verified that, there is $\mathcal{O} \subseteq (P^{\rm obs} \setminus \mathcal{W})$ such that $|\mathcal{O}| \leq k$, both $P(\mathcal{P})_{\mathcal{W}} \models \neg\mathcal{W}$ and $P(\mathcal{P})_{\mathcal{W},\mathcal{O}}$ hold. Since the number of such outliers is upper bounded by $|P^{\rm obs}|^k$, hence polynomially time bounded, and since $P(\mathcal{P})$ is stratified, the overall check is feasible in polynomial time. As for the hardness, a reduction is exploited to the P-complete problem PROPOSITIONAL STRATIFIED LOGIC PROGRAMMING, i.e. the problem: given a propositional stratified logic program $P$ and an atom $a$, decide whether $P \models \neg a$. Consider the rule-observation pair $\mathcal{P} = \langle P^{\rm rls}, P^{\rm obs} \rangle$, with $P^{\rm rls} = P \cup \{a \leftarrow \neg b\}$ and $P^{\rm obs} = \{a, b\}$, where $b$ is a propositional letter not occurring in $P$. Clearly, $P^{\rm rls}$ is stratified and there is an outlier $\mathcal{O}$ of size $|\mathcal{O}| \leq 1$ in $\mathcal{P}$ having witness $\mathcal{W} = \{a\}$ iff $P \models \neg a$.

2. Firstly, let us consider brave semantics. Let $\mathcal{W}$ be a subset of $P^{\rm obs}$, and let $\mathcal{O}_1, \ldots, \mathcal{O}_m$, $m = \binom{|P^{\rm obs} \setminus \mathcal{W}|}{1} + \binom{|P^{\rm obs} \setminus \mathcal{W}|}{2} + \ldots + \binom{|P^{\rm obs} \setminus \mathcal{W}|}{k}$, be all the subsets of $P^{\rm obs} \setminus \mathcal{W}$ having size equal or less than $k$. Notice that deciding checking whether $P(\mathcal{P})_{\mathcal{W}} \models_b \neg\mathcal{W}$, i.e., whether condition (1) in Definition1 is satisfied, is feasible in NP since it amounts to guessing a model for $P(\mathcal{P})_{\mathcal{W}}$ and at verifying that it entails $\neg\mathcal{W}$.
Let us consider, instead, condition (2) subject to the size constraint, i.e., it must be decided the existence of a set $\mathcal{O}$, with $|\mathcal{O}| \leq k$, such that $P(\mathcal{P})_{\mathcal{W},\mathcal{O}} \not\models_b \neg\mathcal{W}$ . Actually, the complementary condition can be faced, denoted by $\overline{\texttt{C}_2}$, of deciding whether for each set $\mathcal{O} \in \{\mathcal{O}_1, \ldots, \mathcal{O}_m\}$, there is a model $M$ of $P(\mathcal{P})_{\mathcal{W},\mathcal{O}}$ such that $M$ entails $\neg\mathcal{W}$.
To this aim, let $P_i$, $1 \leq i \leq m$, denote the new logic program obtained from $P(\mathcal{P})_{\mathcal{W},\mathcal{O}_i}$ by replacing each propositional letter $p$ occurring there with a local copy of $p$, say $p^i$, and let $\mathcal{W}_i$ denote the new set obtained from $\mathcal{W}$ by substituting to each propositional letter $w$ there occurring, the local copy $w^i$ of $w$ in $P_i$. Consider the logic program $P' = P_1 \cup \ldots \cup P_m$ and the set $\mathcal{W}' = \mathcal{W}_1 \cup \ldots \cup \mathcal{W}_m$. We note that both $P'$ and $\mathcal{W}'$ can be built in polynomial time. Now, it is easy to see that, by construction, $P' \models_b \neg\mathcal{W}'$ if and only if $\overline{\mathtt{C}_2}$ is satisfied. Thus, given $\mathcal{W} \subseteq P^{\rm obs}$, checking that $P(\mathcal{P})_{\mathcal{W}} \models_b \neg\mathcal{W}$ and that $P(\mathcal{P})_{\mathcal{W},\mathcal{O}} \not\models_b \neg\mathcal{W}$ amounts to solving an NP and a co-NP problem. As for the hardness, given two boolean formulas $\phi$ and $\phi'$, the problem of deciding whether $\phi$ is satisfiable and $\phi'$ is unsatisfiable can be reduced to $\mathtt{WITNESS-CHECKING}[\mathtt{k}]$ under brave semantics using a reduction similar to that shown in Point 3 of the proof of Theorem 6.
Finally, the proof for cautious semantics follows the one of Point 3 of Theorem 6. □

### 4.2 Complexity of Computation Problems

As already done in the context of Section 3.2, next let us concentrate on computation problems. Specifically, interest is in the $\mathtt{COMPUTATION}[\min]$ problem: computing an outlier whose size is the minimum over the sizes of all the outliers – by $min(\mathcal{P})$ the minimum size value is denoted. Notice that in the case of no outlier, $min(\mathcal{P})$ is undefined. To this aim, the computational complexity of a (still, decision) variant of the $\mathtt{OW-CHECKING}$ problem is studied, denoted by $\mathtt{OW-CHECKING}[\min]$, in which the attention is focused on checking minimum-size outliers only: given $\mathcal{O}, \mathcal{W} \subseteq P^{\rm obs}$, is $\mathcal{O}$ an outlier in $\mathcal{P}$, with witness $\mathcal{W}$, such that $|\mathcal{O}| = min(\mathcal{P})$?

**Theorem 14.** *Let $\mathcal{P} = \langle P^{\rm rls}, P^{\rm obs}\rangle$ be a rule-observation pair. Then, the problem* $\mathtt{OW-CHECKING}[\min]$ *is*

1. co-NP-*complete, for stratified $P^{\rm rls}$, and*
2. *$\Pi_2^P$-complete (under both brave and cautious semantics), for general $P^{\rm rls}$.*

*Proof.*

1. (Membership) Given $\mathcal{O}, \mathcal{W} \subseteq P^{\rm obs}$, let us consider the complementary problem $\overline{\mathtt{OW-CHECKING}[\min]}$ of deciding whether *it is not true* that $\mathcal{O}$ is an outlier in $\mathcal{P}$, with witness $\mathcal{W}$, such that $|\mathcal{O}| = min(\mathcal{P})$. This problem can be solved by building a polynomial-time nondeterministic Turing machine that $(i)$ verifies in polynomial time that $\mathcal{O}$ is an outlier with witness $\mathcal{W}$ in $\mathcal{P}$ (if it is not the case then the machines stops replying "yes"), $(ii)$ guesses the sets set $\mathcal{O}'$ and $\mathcal{W}'$, and verifies in polynomial time that $(iii)$ $\mathcal{O}'$ is an outlier with witness $\mathcal{W}'$ in $\mathcal{P}$, and $(iv)$ $|\mathcal{O}'| < |\mathcal{O}|$ – notice that both $(i)$ and $(iv)$ have been observed to be feasible in polynomial time in the membership part of Theorem 2. Then, $\overline{\mathtt{OW-CHECKING}[\min]}$ is feasible in NP and, hence, $\mathtt{OW-CHECKING}[\min]$ is in co-NP.
   (Hardness) Recall that deciding whether a Boolean formula $\Phi$ in conjunctive normal form is not satisfiable, is an co-NP-complete problem. Consider again the rule-observation pair $\mathcal{P}(\Phi) = \langle P^{\rm rls}(\Phi), P^{\rm obs}(\Phi)\rangle$ built in the proof of Theorem 2. Then, we build in polynomial time the rule-observation pair $\mathcal{P}^*(\Phi) = \langle P^{*\rm rls}(\Phi), P^{*\rm obs}(\Phi)\rangle$ such that: $(i)$ $P^{*\rm obs}(\Phi) = P^{\rm obs}(\Phi) \cup P'^{\rm obs}$, where $P'^{\rm obs} = \{w, o_1, o_2\}$, and $(ii)$ $P^{*\rm rls}(\Phi) = P^{\rm rls}(\Phi) \cup P'^{\rm rls}$ where $P'^{\rm rls}$ is:

$$w \leftarrow not\ o_1, not\ o_2.$$
$$o_1 \leftarrow o_2.$$
$$o_2 \leftarrow o_1.$$

   It is easy to see that the set $\mathcal{O} = \{o_1, o_2\}$ is an outlier in $\mathcal{P}^*(\Phi)$ with witness $\mathcal{W} = \{w\}$. Moreover, $min(\mathcal{P}^*(\Phi)) = |\mathcal{O}| \Leftrightarrow \Phi$ is not satisfiable. Indeed, in Theorem 2 it was shown that $\Phi$ is satisfiable $\Leftrightarrow$ $\mathcal{O}' = \{disabled\}$ is an outlier in $\mathcal{P}(\Phi)$. Hence, the result follows by observing that $\mathcal{O}'$ is such that $|\mathcal{O}'| < |\mathcal{O}|$.

2. (Membership) The membership in $\Pi_2^P$ derives from the fact that the complementary problem $\overline{\texttt{OW}-\texttt{CHECKING}[\mathrm{min}]}$ can be solved in $\Sigma_2^P$ under both brave and cautious semantics. Indeed, a polynomial-time nondeterministic Turing machine with an NP oracle can be built that $(i)$ verifies that $\mathcal{O}$ is an outlier with witness $\mathcal{W}$ in $\mathcal{P}$ making two calls to the oracle (if it is not the case then the machines stops replying "yes"), $(ii)$ guesses the sets set $\mathcal{O}'$ and $\mathcal{W}'$, and verifies $(iii)$ that $\mathcal{O}'$ is an outlier with witness $\mathcal{W}'$ in $\mathcal{P}$ making two other calls to the oracle, and $(iv)$ that $|\mathcal{O}'| < |\mathcal{O}|$ – notice that both $(i)$ and $(iv)$ have been observed to be $D^P$-complete in Theorem 7. Then, $\overline{\texttt{OW}-\texttt{CHECKING}[\mathrm{min}]}$ is feasible in $\Sigma_2^P$ and, hence, $\texttt{OW}-\texttt{CHECKING}[\mathrm{min}]$ is in $\Pi_2^P$.
(Hardness) Let $\Phi$ be a quantified Boolean formula in disjunctive normal form. Recall that deciding whether $\Phi$ is not valid is a $\Pi_2^P$-complete problem. The result follows immediately by exploiting the reduction described in the Hardness part in the Point 1 above, the only difference being that the rule-observation pairs $\mathcal{P}(\Phi)$ to be considered are respectively those described in Theorems 3 and 4. To this aim it is sufficient to observe that in the proofs for the two theorems above, it was shown that the formula $\Phi$ is satisfiable if and only if there is an outlier of the form $\{disabled\}$. □

By exploiting Theorem 14, it can be shown an $\mathrm{F}\Sigma_2^P$ ($\mathrm{F}\Sigma_3^P$ resp.) upper bound to the complexity of the problem of computing an outlier having minimum size in stratified (resp., general) pairs. This can be done by first guessing any outlier with an associated witness and then verifying that it is indeed minimal. But, actually better than this can be achieved by defining a more efficient computation method based on identifying the actual value of $min(\mathcal{P})$ and, then, guessing an outlier whose size equals $min(\mathcal{P})$.

The case of stratified rule observation pairs is firstly considered, for which the problem turns out to be $\mathrm{F}\Delta_2^P$[O(log $n$)]-complete, where $n$ denotes the size of the rule-observation pair. The result can be derived by establishing a one-to-one correspondence between outliers and cliques in a graph – computing the size of the maximum clique was, in fact, shown to be $\mathrm{F}\Delta_2^P$[O(log $n$)]-complete in [57]. Notice that, to this end, the proof of Theorem 2 does not help because it exploits a construction from the satisfiability problem and, more importantly, because the guess of the true variables in the assignment is shared by the outlier and its associated witness, so that for any satisfying assignment there is an exponential number of associated outliers. Therefore, the most relevant technical problem to be solved in order to prove the result is to establish the cited one-to-one correspondence between outliers and graph cliques, by exploiting a more intricate reduction[5] than the one used in Theorem 2.

**Theorem 15.** *Given a rule-observation pair $\mathcal{P} = \langle P^{\mathrm{rls}}, P^{\mathrm{obs}}\rangle$, computing the value $min(\mathcal{P})$ (if defined) is $\mathrm{F}\Delta_2^P$[O(log $n$)]-complete, for stratified $P^{\mathrm{rls}}$.*

*Proof.* (Membership) Given the pair $\mathcal{P}$, it is preliminary observed that the maximum value of $min(\mathcal{P})$ is $max = |P^{\mathrm{obs}}| = O(|\mathcal{P}|)$. Then, by a binary search on the range $[0, max]$, the value $min(\mathcal{P})$ can be computed: at each step of the search, a threshold is given in the range $[0, max]$, say $k$, and it has to be decided whether the problem $\texttt{EXISTENCE}[\texttt{k}]$ has some solution. After $\log max$ steps at most the procedure ends, and the value of $min(\mathcal{P})$ can be returned, if this value is not zero. Since $\texttt{EXISTENCE}[\texttt{k}]$ is feasible in NP, it follows that the procedure is feasible in $\mathrm{FP}^{\mathrm{NP}[O(\log n)]} = \mathrm{F}\Delta_2^P$[O(log $n$)], where $n$ is $|\mathcal{P}|$.

(Hardness) Given a graph $G = \langle A, E\rangle$, with $A = \{1, \ldots, n\}$ being a set of nodes and $E \subseteq A \times A$ a set of edges, a clique $C$ for $G$ is a set of nodes such that $\forall i, j \in C$, there is an edge $(i, j) \in E$. Recall that computing the size of the maximum clique in a graph is $\mathrm{F}\Delta_2^P$[O(log $n$)] [19]. A rule-observation pair

[5] This reduction might indeed also be used to prove Theorem 2. However, its intricacies would have made understanding the intrinsic complexity of the basic outlier detection problem much more difficult to grasp.

$\mathcal{P}(G) = \langle P^{\text{rls}}(G), P^{\text{obs}}(G)\rangle$ is built such that: ($i$) $P^{\text{obs}}(G)$ contains the facts $x_i^{in}$, $x_i^{out}$, and $y_i$, for each node $i$ in $A$, and the fact $yes$; ($ii$) $P^{\text{rls}}(G)$ is

$$
\begin{array}{lll}
r_1: & unsat \leftarrow x_i^{in}, x_j^{in}. & 1 \le i,j \le n \text{ s.t. } (i,j) \notin E \\
r_2: & unsat \leftarrow x_i^{in}, x_i^{out}. & 1 \le i \le n \\
r_3: & ok_i \leftarrow x_i^{in}. & 1 \le i \le n \\
r_4: & ok_i \leftarrow x_i^{out}. & 1 \le i \le n \\
r_5: & ok \leftarrow ok_1, \ldots, ok_n. & \\
r_6: & yes \leftarrow not\ unsat, ok. & \\
r_7: & x_i^{in} \leftarrow x_j^{in}, unsat, ok. & 1 \le i,j \le n \\
r_8: & x_i^{in} \leftarrow x_j^{out}, unsat, ok. & 1 \le i,j \le n \\
r_9: & x_i^{out} \leftarrow x_j^{in}, unsat, ok. & 1 \le i,j \le n \\
r_{10}: & x_i^{out} \leftarrow x_j^{out}, unsat, ok. & 1 \le i,j \le n \\
r_{11}: & y_i \leftarrow x_i^{in}, ok. & 1 \le i \le n \\
r_{12}: & x_i^{in} \leftarrow y_i, ok. & 1 \le i \le n
\end{array}
$$

Let $C$ be a clique in $G$. Let us denote by $\mathcal{O}^C$ the set $\{x_1^{\ell_1}, \ldots, x_n^{\ell_n}\} \cup \{y_i \mid \ell_i = in\}$, where each $\ell_i$ is $in$ (resp. $out$) if and only if the node $i$ is not (resp. is) in $C$. Next it is shown that the above construction is such that ($i$) for each clique $C$, the set $\mathcal{O}^C$ is an outlier with witness $\mathcal{W} = \{yes\}$ in $\mathcal{P}(G)$; and ($ii$) each outlier $\mathcal{O}$ in $\mathcal{P}(G)$ is of the form $\mathcal{O} = \mathcal{O}^C$, for some clique $C$.

($i$) Let $C$ be a clique in $G$. Consider the set $\mathcal{W} = \{yes\}$. Then, each fact of the form $x_i^{in}$, $x_i^{out}$, and $y_i$ is true in the program $P(\mathcal{P}(G))_{\mathcal{W}}$. Hence, due to the rules $unsat \leftarrow x_i^{in}, x_i^{out}$, $unsat$ is true in turn, and $P(\mathcal{P}(G))_{\mathcal{W}} \models \neg yes$, i.e., condition (1) in Definition 1 is satisfied. Let us now consider the program $P(\mathcal{P}(G))_{\mathcal{W},\mathcal{O}}$, for $\mathcal{O} = \mathcal{O}^C$. It must be shown that $P(\mathcal{P}(G))_{\mathcal{W},\mathcal{O}} \models yes$. To this aim, the unique stable model of $P(\mathcal{P}(G))_{\mathcal{W},\mathcal{O}}$ is proven to be $M = \{yes\} \cup \{ok, ok_1, \ldots, ok_n\} \cup \{x_1^{in}, \ldots, x_n^{in}\} \cup \{x_1^{out}, \ldots, x_n^{out}\} \cup \{y_1, \ldots, y_n\} \setminus \mathcal{O}^C$, i.e., $M$ is the unique minimal model of $P(\mathcal{P}(G))^M_{\mathcal{W},\mathcal{O}}$. Actually, the structure of $P(\mathcal{P}(G))^M_{\mathcal{W},\mathcal{O}}$ is the same as the one of $P(\mathcal{P}(G))_{\mathcal{W},\mathcal{O}}$ but for the rule $r_6 : yes \leftarrow not\ unsat, ok$ replaced by the rule $r_6' : yes \leftarrow ok.$, because of the fact that $unsat$ is false in $M$. Let us first note that $M$ is in fact a model of $P(\mathcal{P}(G))^M_{\mathcal{W},\mathcal{O}}$, by exploiting the following arguments:

- Rules $r_3$, $r_4$ and $r_5$ are satisfied by $M$, as $ok$ and all the facts of the form $ok_i$ are true in $M$.
- Rule $r_6'$ is satisfied by $M$, because $yes$ belongs to $M$.
- Rules $r_7$, $r_8$, $r_9$, and $r_{10}$ are satisfied by $M$, because $unsat$ does not belong to $M$.
- Rules $r_{11}$ and $r_{12}$ are satisfied by $M$. Indeed, by construction of the set $\mathcal{O}^C$, $y_i \notin \mathcal{O}^C$ iff $x_i^{out} \in \mathcal{O}^C$ and, therefore, $y_i \in M$ iff $x_i^{in} \in M$ holds as well.
- Rule $r_2$ is satisfied by $M$, because of the construction of $M$ and $\mathcal{O}^C$ which prevents from $x_i^{in}$ and $x_i^{out}$ being both true at the same time (a node is either in the clique or it is not).
- Rule $r_1$ is satisfied by $M$. Indeed, it is preliminary noticed that the body of the rule accounts for pairs of nodes that are not connected by means of an edge in $E$. Then, for each pair of predicates $x_i^{in}$ and $x_j^{in}$ being true in $M$, it is the case that they are connected by means of an edge in $E$; this is because $x_i^{out}$ and $x_j^{out}$ are in fact in $\mathcal{O}^C$ and, therefore, are nodes of the clique $C$. Thus, $unsat$ cannot be entailed by this rule.

To conclude the proof of ($i$), it is necessary now to show that $M$ is in fact the minimal model of the program $P(\mathcal{P}(G))^M_{\mathcal{W},\mathcal{O}}$. Let us preliminary notice that the atoms in $\{x_1^{in}, \ldots, x_n^{in}\} \cup \{x_1^{out}, \ldots, x_n^{out}\} \cup \{y_1, \ldots, y_n\} \setminus \mathcal{O}^C$ came as facts in $P(\mathcal{P}(G))^M_{\mathcal{W},\mathcal{O}}$ and, therefore, they must be in any model of $P(\mathcal{P}(G))^M_{\mathcal{W},\mathcal{O}}$. Then, the result follows by noticing that $P(\mathcal{P}(G))^M_{\mathcal{W},\mathcal{O}}$ entails $\{yes\} \cup \{ok, ok_1, \ldots, ok_n\}$. Indeed, $\mathcal{O}^C$ contains by construction one element in the set $\{x_i^{in}, x_i^{out}\}$, for each $x_i$. Therefore, it is possible to entail $ok_i$, for each $1 \leq i \leq n$, and $ok$ in turn. Consequently $\{yes\}$ must occur in any model of $P(\mathcal{P}(G))^M_{\mathcal{W},\mathcal{O}}$.

$(ii)$ Let $\mathcal{O}$ be an outlier for $\mathcal{P}(G)$. In order to prove the result, some properties of the encoding of $G$ are discussed.

**Property $P_1$:** Any witness $\mathcal{W}$ for $\mathcal{O}$ is such that $P(\mathcal{P}(G))_{\mathcal{W}} \models ok$ and also that $P(\mathcal{P}(G))_{\mathcal{W},\mathcal{O}} \models ok$.

*Proof.* For the sake of contradiction, assume that $ok$ is not entailed in $P(\mathcal{P}(G))_{\mathcal{W}}$. Then, due to rules $r_3$ and $r_4$, it must be the case that there is a node $i$ such that both $x_i^{in}$ and $x_i^{out}$ are in $\mathcal{W}$. But, if this is the case, whatever the elements in $\mathcal{O}$ are, there is no chance of entailing $x_i^{in}$ or $x_i^{out}$ in $P(\mathcal{P}(G))_{\mathcal{W},\mathcal{O}}$. Indeed, although stratified logic programs are not monotonic, the monotonicity property still holds for predicates in the first stratum and the result follows by noting that, except for the predicate $yes$, all the other predicates lie in the first stratum of the stratified logic program $P(\mathcal{P}(G))$. It can be concluded that $P(\mathcal{P}(G))_{\mathcal{W},\mathcal{O}} \models \neg ok$ and, hence, $P(\mathcal{P}(G))_{\mathcal{W},\mathcal{O}}$ cannot entail any fact by means of rules whose body contains $ok$. Given that these rules are those which may lead to entail facts in the observation components, this is a violation of condition (2) in Definition 1.

**Property $P_2$:** Any witness $\mathcal{W}$ for $\mathcal{O}$ is such that $P(\mathcal{P}(G))_{\mathcal{W}} \models unsat$.

*Proof.* After $P_1$ it can be assumed, without loss of generality, that $ok$ is true in $P(\mathcal{P}(G))_{\mathcal{W}}$. Beforehand notice that if $x_i^{in}$ is in $\mathcal{W}$ then $y_i$ is in $\mathcal{W}$ too, otherwise, $x_i^{in}$ would be entailed by rule $r_{12}$ in $P(\mathcal{P}(G))_{\mathcal{W}}$ thereby violating condition (1) in Definition 1. Symmetrically, if $y_i$ is in $\mathcal{W}$ then $x_i^{in}$ is in $\mathcal{W}$ too, because of rule $r_{11}$. This means that the body of the rules $r_{11}$ and $r_{12}$ are always false in the program $P(\mathcal{P}(G))_{\mathcal{W},\mathcal{O}}$. Now, assume for the sake of contradiction that $unsat$ is not entailed by $P(\mathcal{P}(G))_{\mathcal{W}}$. As $unsat$ belongs to the first stratum of $P(\mathcal{P}(G))$, then it follows that $unsat$ is also false in the program $P(\mathcal{P}(G))_{\mathcal{W},\mathcal{O}}$. Then, the only fact among those in $P^{\text{obs}}(G)$ that can be entailed in $P(\mathcal{P}(G))_{\mathcal{W},\mathcal{O}}$ is $yes$ that, consequently, must belong to $\mathcal{W}$. But this cannot be the case because if $unsat$ is false, also $P(\mathcal{P}(G))_{\mathcal{W}} \models yes$. Hence, $\mathcal{O}$ is not an outlier.

**Property $P_3$:** Any witness $\mathcal{W}$ for $\mathcal{O}$ is composed exactly by the fact $yes$.

*Proof.* Recall that $\mathcal{W}$ is such that $ok$ and $unsat$ are true in $P(\mathcal{P}(G))_{\mathcal{W}}$ (property $P_1$ and $P_2$ above). Then, assume that any fact of the form $x_i^{in}$, $x_i^{out}$, or $y_i$ is in $\mathcal{W}$. In order to satisfy condition (1) in Definition 1, it must be the case that all the facts having this form are in $\mathcal{W}$, since any fact $x_i^{in}$, $x_i^{out}$, or $y_i$ will suffice for entailing all the others. However, in this case $ok$ is false. Contradiction.

Armed with these properties, it can be concluded that the outlier $\mathcal{O}$ must be such that $yes$ is entailed in the program $P(\mathcal{P}(G))_{\mathcal{W},\mathcal{O}}$, i.e. that both $\neg unsat$ and $ok$ are entailed in the same program. To this aim, for each node $i$, either $x_i^{in}$ or $x_i^{out}$ is in $\mathcal{O}$ (but not together, because of rule $r_2$), and $x_i^{in}$ is in $\mathcal{O}$ if and only if $y_i$ is in $\mathcal{O}$ (because of rules $r_{11}$ and $r_{12}$). Furthermore, in order to have $unsat$ not entailed, it must be the case that the set $C = \{i \mid x_i^{in} \notin \mathcal{O}\}$ corresponds to a clique, because of the rule $r_1$ ($unsat$ is entailed as soon as two nodes $x_i$ and $x_j$ are marked $in$ while being not connected by means of an edge). The result follows by observing that $\mathcal{O}$ with the properties stated above coincides with $\mathcal{O}^C$.

In order to conclude the proof, simply observe that given a clique $C$ the corresponding outlier $\mathcal{O}^C$ is such that $|\mathcal{O}^C| = n + |\{y_i \mid \ell_i = in\}| = n + n - |C|$. Thus, the clique having maximum size is in one-to-one correspondence with the outlier having minimum size, and computing $min(\mathcal{P})$ amounts to compute the size of the maximum clique. □

One may ask what is the complexity to be paid to compute the value $min(\mathcal{P})$ for general programs. The following theorem partially answers the question, by providing an upper bound for it, while leaving it open an exact characterization of its intrinsic complexity.

**Theorem 16.** *Given a rule-observation pair $\mathcal{P} = \langle P^{\rm rls}, P^{\rm obs}\rangle$, computing the value $min(\mathcal{P})$ (if defined) is in $\mathrm{F}\Delta_3^P[\mathrm{O}(\log n)]$ (under both brave and cautious semantics), for general $P^{\rm rls}$.*

*Proof.* The proof is the same as for the membership part in the Theorem 15, except for the fact that $\mathtt{EXISTENCE}[\mathtt{k}]$ is feasible in $\Sigma_2^P$ for general logic programs. Then, the binary search can be implemented in $\mathrm{FP}^{\Sigma_2^P[O(\log n)]} = \mathrm{F}\Delta_3^P[\mathrm{O}(\log n)]$. □

Using the result demonstrated above, it is not difficult to see that given a rule-observation pair $\mathcal{P}$, an outlier of minimum size can be computed in polynomial time using an NP (resp. $\Sigma_2^P$) oracle for stratified (resp. general) logic programs.

**Theorem 17.** *Given a rule-observation pair $\mathcal{P} = \langle P^{\rm rls}, P^{\rm obs}\rangle$, computing an arbitrary outlier $\mathcal{O}$ such that $min(\mathcal{P}) = |\mathcal{O}|$ (if defined) is*

1. *in $\mathrm{F}\Delta_2^P[\mathrm{O}(\log n)]$, for stratified $P^{\rm rls}$, and*
2. *in $\mathrm{F}\Delta_3^P[\mathrm{O}(\log n)]$ (under both brave and cautious semantics), for general $P^{\rm rls}$.*

*Proof.* The problem can be solved by $(i)$ computing the value $min(\mathcal{P})$ — it has been seen that this value can be computed performing $O(\log |\mathcal{P}|)$ calls to an NP (resp. $\Sigma_2^P$) oracle, and then $(ii)$ guessing an outlier $\mathcal{O}$ having size $|\mathcal{O}| = min(\mathcal{P})$, together with its witness $\mathcal{W}$, and checking conditions (1) and (2) of Definition 1. Point $(ii)$ above is feasible with an extra NP (resp. $\Sigma_2^P$) oracle call for stratified (resp. general) logic programs. Hence, the following result follows. □

One may wonder whether the computation problem is, in fact, complete for the above complexity classes. Actually, for the case of stratified LPs, the simple membership result can be sharpened by assessing its precise complexity, which account for the possibility of having several outliers with size $min(\mathcal{P})$. To this end, it is necessary to recall some further complexity notions.

An NP *metric Turing machine* $MT$ is a polynomial-time bounded nondeterministic Turing machine that on every computation branch halts and outputs a binary number. The result computed by $MT$ is the maximum over all these numbers. The class $\mathrm{OptP}$ contains all integer functions that are computed by an NP metric Turing machine, whereas $\mathrm{OptP}[O(\log n)]$ is that subclass thereof containing all functions $f$ whose value $f(x)$ has $O(\log n)$ bits, where $n = |x|$. The class FNP//OptP[O(log *n*)] contains all (partial) multi-valued functions $g$ for which a polynomially-bounded nondeterministic Turing machine $T$ and a function $h \in \mathrm{OptP}[O(\log n)]$ exist such that, for every $x$, $T$ computes the value $g(x)$, provided that both $x$ and $h(x)$ are taken in input (see [19]). Notice that it is well-known that every multi-valued function $g \in$ FNP//OptP[O(log *n*)] has a refinement (single-valued) function $f \in \mathrm{F}\Delta_2^P[\mathrm{O}(\log n)]$, i.e., for every $x$ it holds that $g(x)$ is defined iff $f(x)$ is defined and $f(x) \in g(x)$. Moreover, a problem (with possibly multiple solutions for a given instance) is solvable in FNP//OptP[O(log *n*)] iff any of such solutions is computable in $\mathrm{F}\Delta_2^P[\mathrm{O}(\log n)]$.

In the light of this observation, the proof of Theorem 17 provides an algorithm for computing a refinement of $\mathtt{COMPUTATION}[\mathrm{min}]$; the complexity of $\mathtt{COMPUTATION}[\mathrm{min}]$ on its own is, instead, more naturally defined in terms of the class FNP//OptP[O(log *n*)], where $n = |\mathcal{P}|$ is the size of the rule-observation pair, as shown next.

**Theorem 18.** *Let $\mathcal{P} = \langle P^{\mathrm{rls}}, P^{\mathrm{obs}} \rangle$ be a stratified rule-observation pair. Then,* `COMPUTATION`$[\min]$ *is* FNP//OptP[O(log *n*)]*-complete.*

*Proof.* (Membership) The value $min(\mathcal{P})$ has $O(\log n)$ many bits at most, where $n$ is the size of the input, and it can be computed by an NP metric Turing machine. Then, let us guess two sets $\mathcal{O}$ and $\mathcal{W}$ and verify in polynomial time that (i) $\mathcal{O}$ is an outlier with witness $\mathcal{W}$, and (ii) the size of $\mathcal{O}$ is $min(\mathcal{P})$.

(Hardness) A reduction can be shown to the $X$-MAXIMAL MODEL problem: Given a formula $\phi = c_1 \wedge \ldots \wedge c_m$ in conjunctive normal form on the variables $Y = \{Y_1, ..., Y_h\}$ and a subset $X \subseteq Y$, compute a satisfying truth assignment $M$ for $\phi$ whose $X$-part is maximal, i.e., for every other satisfying assignment $M'$ there is a variable in $X$ which is true in $M$ and false in $M'$. This problem was proved to be FNP//OptP[O(log *n*)]-complete in [19] under the following notion of metric reduction: A problem $\mathit{\Pi}$ reduces to a problem $\mathit{\Pi}'$, if there are polynomial-time computable functions $f(x)$ and $g(x, y)$, such that: (i) for any instance $I$ of $\mathit{\Pi}$, $f(I)$ is an instance of $\mathit{\Pi}'$, and $f(I)$ has solution iff $I$ has a solution, and (ii) for any arbitrary solution $S$ of $f(I)$, $g(I, S)$ is a solution of $I$.

A rule-observation pair $\mathcal{P}(\phi) = \langle P^{\mathrm{rls}}(\phi), P^{\mathrm{obs}}(\phi) \rangle$ is built such that: (i) $P^{\mathrm{obs}}(\phi)$ contains the two facts $x_i^{in}$, $x_i^{out}$ for each variable $Y_i$ in $Y$, the fact $y_i$ for each variable $Y_i$ in $X$, and the fact $yes$; (ii) $P^{\mathrm{rls}}(\mathit{\Phi})$ consists of the rules $r_2, \ldots, r_{12}$ of the encoding in the proof of Theorem 15, plus the following rule:

$$r_1' : unsat \leftarrow \nu(t_{j,1}), \nu(t_{j,2}), \nu(t_{j,3}). \quad 1 \leq j \leq m \text{ s.t. } c_j = t_{j,1} \vee t_{j,2} \vee t_{j,3}$$

where $\nu$ is the following mapping:

$$\nu(t) = \begin{cases} x_i^{out} & \text{, if } t = Y_i \\ x_i^{in} & \text{, if } t = \neg Y_i \end{cases}$$

Let $\sigma$ be an assignment for $\phi$. Let us denote by $\mathcal{O}^\sigma$ the set $\{x_1^{\ell_1}, \ldots, x_n^{\ell_n}\} \cup \{y_i \mid \ell_i = in \wedge Y_i \in X\}$, where each $\ell_i$ is $in$ (resp. $out$) if and only if the variable $Y_i$ is not (resp. is) true in $\sigma$.

Let us comment the only difference with the encoding used in Theorem 15, i.e., the substitution of rule $r_1$ with rule $r_1'$. Actually, rule $r_1$ was used to evaluate whether the set of facts of the form $x_i^{in}$ form a clique: if the clique is not correctly formed, then $unsat$ is entailed by the program. The behavior of rule $r_1'$ is symmetric, because the rule checks whether the assignment at hand satisfies the formula: if the assignment is not satisfying, then $unsat$ is entailed by the program. Provided this modification only, with the same arguments as in the proof of Theorem 15, it can be seen that $(i)$ for each satisfying assignment $\sigma$, the set $\mathcal{O}^\sigma$ is an outlier with witness $\mathcal{W} = \{yes\}$ in $\mathcal{P}(\phi)$; and, $(ii)$ each outlier $\mathcal{O}$ in $\mathcal{P}(\phi)$ is of the form $\mathcal{O} = \mathcal{O}^\sigma$, for some satisfying assignment $\sigma$.

Observe now that given an assignment $\sigma$, the corresponding outlier $\mathcal{O}^\sigma$ is such that $|\mathcal{O}^\sigma| = n + |\{y_i \mid \ell_i = in \wedge Y_i \in X\}|$, and recall that $\ell_i = in$ means that the variable is false in $\sigma$. Hence, the reduction establishes a one-to-one correspondence not only between outliers and truth assignments for $X$, but also between minimum size outliers and the satisfying assignments for $\phi$ with a maximum number of $X$ variables made true. Clearly, each of such assignments is also an $X$-MAXIMAL MODEL of $\phi$. The result follows by observing that the construction can be done in polynomial time. □

It is worth pointing out that the reduction presented above is *parsimonious* [40]. In fact, by letting $Y$ be the set of all the variables in the formula $\phi$, the theorem above establishes a one-to-one correspondence between outliers and satisfying assignments. Then, as a side result, the cost of counting the number of outliers turns out to be the same as the cost of computing the number of satisfying assignments, the archetypical complete problem for the class #P [82].

**Corollary 2.** *Let $\mathcal{P} = \langle P^{\mathrm{rls}}, P^{\mathrm{obs}} \rangle$ be a rule-observation pair such that $P^{\mathrm{rls}}$ is a stratified. Then, counting the number of outliers in $\mathcal{P}$ is $\#\mathrm{P}$-complete.*

As with minimal diagnosis applications, an interesting problem is that of singling out an outlier "core", that is, those facts that belong to all minimum-size outliers. This apparently difficult problem turns out to be not more difficult than computing the value of $min(\mathcal{P})$, as shown in the following theorem.

**Theorem 19.** *Given a stratified rule-observation pair $\mathcal{P} = \langle P^{\mathrm{rls}}, P^{\mathrm{obs}} \rangle$ and a fact $f$, deciding whether, for each outlier $\mathcal{O}$ having minimum size in $\mathcal{P}$, it holds that $f \in \mathcal{O}$ is $\Delta_2^P[O(\log n)]$-complete.*

*Proof.* (Membership) The problem can be solved in polynomial time with $O(\log n)$ many NP oracle calls. The value $min(\mathcal{P})$ can be firstly computed by exploiting the algorithm presented in Theorem 15. Then, with an additional NP oracle it can be checked if there is an outlier $\mathcal{O}$ such that (i) $|\mathcal{O}| = min(\mathcal{P})$, and (ii) $f \notin \mathcal{O}$.

(Hardness) Given a formula $\phi$ in conjunctive normal form on the variables $Y = \{Y_1, ..., Y_n\}$, a subset $X \subseteq Y$, and a variable $Y_i$, deciding whether $Y_i$ is true in all the $X$-MAXIMUM models is $\Delta_2^P[O(\log n)]$-complete, where a model is maximum if it has the largest $X$-part. The result trivially follows by exploiting the construction in Theorem 18 that, in fact, relies on a one-to-one correspondence between $X$-MAXIMUM models and outliers of minimum size. □

## 5 Observing Rules

The outlier detection framework depicted so far relies on agents' observations coming as facts that encode some aspects of the current status of the world. To this aim, it suffices that agents have some "sensing" capability for monitoring the external environment. However, it would be desirable to have agents that, besides sensing, would also be able to interact with the environment in a more elaborate way.

For instance, in several multi-agent applications, agents may be involved in dialogues with other agents. Dialogues may start from the need to achieve an explicit goal, such as to persuade another party, or to find an information, or to verify an assumption (cf. [86]). Other form of dialogues may occur during a negotiation [56], i.e., when agents operate in an environment with limited resource availability and their goal is to obtain a resource (see, e.g., [87, 78]). In this contexts, agents might (maliciously) export knowledge for taking advantage of its competitor agents and, therefore, it is relevant to have agents equipped with the capability of having a set of facts and *rules* to encode their own "observations", i.e., the knowledge exchanged with other agents while communicating.

Then, outlier detection techniques can be profitably used for singling out the pieces of knowledge (defined as sets of rules and facts and acquired by communicating with other agents or generally by learning from the environment) that look anomalous w.r.t. the trustable internal agent view of the world.

*Example 5.* Consider again the agent $A^N$ that is in charge of monitoring the status of the network $\mathcal{N}$. Assume that $A^N$ has no a-priori knowledge about the connectivity, i.e., the rule component $P^N$ is empty. Then, $A^N$ might have been informed by another agent of a property about $\mathcal{N}$, which is encoded in the rule $o:\ \mathtt{up(c)} \leftarrow \mathtt{up(h)}$.

Armed with this rule, the agent monitors the status of the net and observes that `h` is up but, surprisingly, `c` is not. Clearly, there is something strange about these observations. However, in this case the agent might doubt the fact that `h` is actually up, but also about the rule $o$, which has been observed, but is not part of its trustable knowledge. ◁

In order to study this extended framework, it is necessary to introduce some changes in the basic definition. Let $P^{\rm rls}$ be a logic program encoding general knowledge about the world, and let $P^{\rm e-obs}$ be a set of facts and rules encoding some *observed* aspects of the current status of the world, called *extended observation component*. Then, the structure $\mathcal{P} = \langle P^{\rm rls}, P^{\rm e-obs}\rangle$ is an *extended rule-observation pair*.

It is worth noting that, in this novel context, the type of the logic program $P(\mathcal{P})$ is determined by the rule program *together* with the extended observation set.

It is easy to see that the complexity of the outlier detection problems studied so far remains unchanged for extended rule-observation pairs where $P(\mathcal{P})$ is a stratified or a general logic program. Indeed, all the hardness proofs refer to $P^{\rm obs}$ including facts only (and hence hold in the extended framework) and all the membership results can easily accommodate rules in the observations (belonging to the same class of logic programs to which $\mathrm{P}^{\rm rls}$ belongs to) without additional costs.

However, it is interesting to study what happens in the case of positive rule components, for which Theorem 1 showed that no outlier exists if observations are restricted to be facts. To analyze this scenario, it is first made clear that a rule $r$ is entailed by a positive program $P$ if $r$ is satisfied by the unique minimal model $M$ of $P$.

Surprisingly, Theorem 1 does not hold for the case of extended rule-observation pairs. For instance, one can consider the extended rule-observation pair $\mathcal{P} = \langle P^{\rm rls}, P^{\rm e-obs}\rangle$ with $P^{\rm rls} = \emptyset$ and $P^{\rm e-obs} = \{a \leftarrow b, b\}$. Then $b$ is an outlier with witness $a \leftarrow b$. Indeed $P(\mathcal{P})_{\{b\},\{a\leftarrow b\}} \not\models \neg(a \leftarrow b)$, while $P(\mathcal{P})_{\{a\leftarrow b\}} \models \neg(a \leftarrow b)$.

Thus, in the following, the computational complexity of the `EXISTENCE` problem is stated when extended rule-observation pairs are considered. This result helps in understanding the characteristics of detecting outliers over extended pairs — complexity figures for the other problems introduced in the paper (such as `OUTLIER−CHECKING`, `WITNESS−CHECKING` and `OW−CHECKING`) can be then obtained by simple adaptations of analogous proofs shown for the standard case, and are, therefore, omitted.

**Theorem 20.** *Let $\mathcal{P} = \langle P^{\rm rls}, P^{\rm e-obs}\rangle$ be an extended rule-observation pair. Then* `EXISTENCE` *is*

1. NP*-complete, for positive and stratified $P(\mathcal{P})$,*[6] *and*
2. *$\Sigma_2^P$-complete (under both brave and cautious semantics), for general $P(\mathcal{P})$.*

*Proof.*

1. Membership can be proven with the same line of reasoning as the proof of Theorem 2. Therefore, let us focus on the hardness which shall be proved for positive $P(\mathcal{P})$. Consider a Boolean formula in conjunctive normal form $\Phi = c_1 \wedge \ldots \wedge c_m$ over the variables $x_1, \ldots, x_n$, such that each clause contains at most three distinct (positive or negated) variables.
   An extended rule-observation pair $\mathcal{P}(\Phi) = \langle P^{\rm rls}(\Phi), P^{\rm e-obs}(\Phi)\rangle$ is defined such that: (i) $P^{\rm e-obs}(\Phi)$ contains exactly the fact $x_i^T$ and $x_i^F$ for each variable $x_i$ in $\Phi$, plus the rule $ok \leftarrow sat$; (ii) $P^{\rm rls}(\Phi)$ is

$$\left.\begin{array}{l} c_j \leftarrow \pi(t_{j,1}). \\ c_j \leftarrow \pi(t_{j,2}). \\ c_j \leftarrow \pi(t_{j,3}). \end{array}\right\} \quad \forall 1 \leq j \leq m, \text{ s.t. } c_j = t_{j,1} \vee t_{j,2} \vee t_{j,3}$$

$$\begin{array}{l} sat \leftarrow c_1,\ \ldots,\ c_m. \\ ok \leftarrow x_i^T,\ x_i^F. \qquad \forall 1 \leq i \leq n \end{array}$$

[6] We thank one of the anonymous referees for having suggested this result and its proof.

where $\pi$ is the mapping:

$$\pi(t) = \begin{cases} x_i^T \text{ , if } t = x_i,\ 1 \leq i \leq n \\ x_i^F \text{ , if } t = \neg x_i,\ 1 \leq i \leq n \end{cases}$$

Clearly, $P(\Phi)$ is positive and can be built in polynomial time. Now it is shown that $\Phi$ is satisfiable $\Leftrightarrow$ there is an outlier in $\mathcal{P}(\Phi)$.

($\Rightarrow$) Suppose that $\Phi$ is satisfiable, and take one of its satisfying truth assignments, say $\sigma$, for the variables $x_1, \ldots, x_n$. Consider the sets $\mathcal{W}^\sigma = \{ok \leftarrow sat\} \cup \{x_i^T \mid x_i \text{ is false in } \sigma\} \cup \{x_i^F \mid x_i \text{ is true in } \sigma\}$ and $\mathcal{O}^\sigma = P^{\text{e-obs}} \setminus \mathcal{W}^\sigma$. It is now shown that $\mathcal{O}^\sigma$ is an outlier with witness $\mathcal{W}^\sigma$. Indeed, the program $P(\mathcal{P}(\Phi))_{\mathcal{W}^\sigma}$ does not entail $ok$, because for each variable $x_i$ either $x_i^T$ or $x_i^F$ is evaluated true. Moreover, it entails $sat$, because of the construction of $\mathcal{W}^\sigma$ and of the fact that the encoding evaluates the truth values of the assignment $\sigma$, which is satisfying. Thus, $P(\mathcal{P}(\Phi))_{\mathcal{W}^\sigma} \models \neg(ok \leftarrow sat)$, and condition (1) in Definition 1 is satisfied. As for condition (2) in Definition 1, it is sufficient to observe that $P(\mathcal{P}(\Phi))_{\mathcal{W}^\sigma, \mathcal{O}^\sigma}$ coincides with $P(\mathcal{P}(\Phi))_{P^{\text{e-obs}}(\Phi)}$; therefore, the program contains no fact and both $sat$ and $ok$ are evaluated false. Thus, $P(\mathcal{P}(\Phi))_{\mathcal{W}^\sigma, \mathcal{O}^\sigma} \models (ok \leftarrow sat)$.

($\Leftarrow$) Assume that there is an outlier $\mathcal{O}$ with witness $\mathcal{W}$ in $\mathcal{P}(\Phi)$. It is shown that $\Phi$ is satisfiable. To this aim, notice that the rule $(ok \leftarrow sat)$ must be part of $\mathcal{W}$, because $ok$ and $sat$ are the only facts that can be entailed in the encoding. Then, because of condition (1) in Definition 1, it must be the case that $P(\mathcal{P}(\Phi))_{\mathcal{W}} \models \neg(ok \leftarrow sat)$, i.e., $P(\mathcal{P}(\Phi))_{\mathcal{W}}$ must entail $sat$ and not $ok$. By the fact that $ok$ is not entailed, it can be concluded that for each variable $x_i$, at least one element in the set $\{x_i^T, x_i^F\}$ is in $\mathcal{W}$. Therefore, $\mathcal{W}$ assigns a truth value to some of the variables of $\Phi$ and may leave undefined some other variables (those variables whose corresponding facts have been both in $\mathcal{W}$). Formally, a (partial) assignment $\sigma$ for the variables in $\Phi$ can be defined such that if $x_i^F$ (resp. $x_i^T$) is in $\mathcal{W}$, then $x_i$ is true (resp. false) in $\sigma$. Then, by the fact that $sat$ is entailed, it must be the case that $\sigma$ is satisfying.

2. The proof is similar to that of Theorems 3 and 4. □

# 6 Implementing Outliers Detection Problems

Now that the framework for outlier detection has been illustrated and its complexity has been investigated, attention can be focused on the problem of devising effective strategies for its implementation. Specifically, sound and complete algorithms are exhibited that transform any rule-observation pair $\mathcal{P}$ into a suitable logic program $\mathcal{L}(\mathcal{P})$ such that its stable models are in a one-to-one correspondence with outliers in $\mathcal{P}$.

The most interesting aspect of this transformation is that, since stable models represent the solution of the outlier problems, it is possible to implement a prototype tool for finding outliers with the support of efficient stable models engines such as GnT [49], DLV [61] and *Smodels* [67]. In fact, reformulations in terms of logic programs under stable model semantics have been already exploited in the literature for prototypically implementing other reasoning tasks such as abduction (see, e.g., [51, 35]), planning (see, e.g., [81, 80]), and diagnosis (see, e.g., [23, 29]).

## 6.1 Stratified Pairs

In this section, the case of a pair $\mathcal{P} = \langle P^{\text{rls}}, P^{\text{obs}} \rangle$ is considered such that $P^{\text{rls}}$ is a stratified logic program. The rewriting algorithm *OutlierDetectionToStableModels* is shown in Figure 5. It takes in input the pair $\mathcal{P}$, and outputs a logic program $\mathcal{L}(\mathcal{P})$, which is built according to the following ideas.

**Input:** A stratified rule-observation pair $\mathcal{P} = \langle P^{\mathrm{rls}}, P^{\mathrm{obs}} \rangle$, where $P^{\mathrm{obs}} = \{\mathtt{obs_1}, ..., \mathtt{obs_n}\}$;
**Output:** A logic program $\mathcal{L}(\mathcal{P})$;
**Method:** Perform the following steps:

1. $\mathcal{L}(\mathcal{P}) := \emptyset$;
2. /*——— Rule part rewriting ———*/
   **for each** rule $r \in P^{\mathrm{rls}}$ of the form $\mathtt{a} \leftarrow \mathtt{b_1}, \cdots, \mathtt{b_k}, \mathtt{not}\ \mathtt{c_1}, \cdots, \mathtt{not}\ \mathtt{c_n}$, **insert into** $\mathcal{L}(\mathcal{P})$ the rules
   (a) $\mathtt{a^{C1}} \leftarrow \mathtt{b_1^{C1}}, \cdots, \mathtt{b_k^{C1}}, \mathtt{not}\ \mathtt{c_1^{C1}}, \cdots, \mathtt{not}\ \mathtt{c_n^{C1}}$.
   (b) $\mathtt{a^{C2}} \leftarrow \mathtt{b_1^{C2}}, \cdots, \mathtt{b_k^{C2}}, \mathtt{not}\ \mathtt{c_1^{C2}}, \cdots, \mathtt{not}\ \mathtt{c_n^{C2}}$.
3. /*——— Outlier and witness guessing ———*/
   **for each** $\mathtt{obs_i} \in P^{\mathrm{obs}}$, **insert into** $\mathcal{L}(\mathcal{P})$ the rules
   (a) $\mathtt{o_i} \leftarrow \mathtt{not}\ \overline{\mathtt{o}}_\mathtt{i}$. $\quad \overline{\mathtt{o}}_\mathtt{i} \leftarrow \mathtt{not}\ \mathtt{o_i}$.
   (b) $\mathtt{w_i} \leftarrow \mathtt{not}\ \overline{\mathtt{w}}_\mathtt{i}$. $\quad \overline{\mathtt{w}}_\mathtt{i} \leftarrow \mathtt{not}\ \mathtt{w_i}$.
4. /*——— Observations definition ———*/
   **for each** $\mathtt{obs_i} \in P^{\mathrm{obs}}$, **insert into** $\mathcal{L}(\mathcal{P})$ the rules
   (a) $\mathtt{obs_i^{C2}} \leftarrow \mathtt{not}\ \mathtt{o_i},\ \mathtt{not}\ \mathtt{w_i}$.
   (b) $\mathtt{obs_i^{C1}} \leftarrow \mathtt{not}\ \mathtt{w_i}$.
5. /*——— Outlier and witness checking ———*/
   **for each** $\mathtt{obs_i} \in P^{\mathrm{obs}}$, **insert into** $\mathcal{L}(\mathcal{P})$ the rules
   (a) $\mathtt{badC1} \leftarrow \mathtt{w_i},\ \mathtt{obs_i^{C1}}$.
   (b) $\mathtt{satC2} \leftarrow \mathtt{w_i},\ \mathtt{obs_i^{C2}}$.
6. /*——— Constraints ———*/
   (a) **for each** $\mathtt{obs_i} \in P^{\mathrm{obs}}$, **insert into** $\mathcal{L}(\mathcal{P})$ the rule $\mathtt{s_1} \leftarrow \mathtt{o_i},\ \mathtt{w_i},\ \mathtt{not}\ \mathtt{s_1}$.
   (b) **insert into** $\mathcal{L}(\mathcal{P})$ the rule $\mathtt{s_2} \leftarrow \mathtt{not}\ \mathtt{satC2},\ \mathtt{not}\ \mathtt{s_2}$.
   (c) **if** $P^{\mathrm{rls}}$ is stratified **then insert into** $\mathcal{L}(\mathcal{P})$ the rule $\mathtt{s_3} \leftarrow \mathtt{badC1},\ \mathtt{not}\ \mathtt{s_3}$.

**Fig. 5.** Algorithm *OutlierDetectionToStableModels*.

First of all, a suitable rewriting of $P^{\mathrm{rls}}$ is inserted into $\mathcal{L}(\mathcal{P})$. Let us denote by $S[\mathtt{L}]$ the rewriting of a set $S$ of rules obtained by substituting each atom $p$ occurring in $S$ with the new atom $p^{\mathtt{L}}$. Then, the algorithm inserts in the Step 2 the programs $P^{\mathrm{rls}}[\mathtt{C1}]$ and $P^{\mathrm{rls}}[\mathtt{C2}]$. Intuitively, $P^{\mathrm{rls}}[\mathtt{C1}]$ is used for checking condition (1) in Definition 1, while $P^{\mathrm{rls}}[\mathtt{C2}]$ is used for checking condition (2) in the same definition.

Rules inserted in Step 3 serve the purpose to guess an outlier and its associated witness. Each fact $\mathtt{obs_i}$ in $P^{\mathrm{obs}}$ is associated with two new facts $\mathtt{o_i}$ and $\mathtt{w_i}$, where, intuitively, $\mathtt{o_i}$ (resp. $\mathtt{w_i}$) being true in a model means that $\mathtt{obs_i}$ belongs to an outlier (resp. witness) in $\mathcal{P}$. In other words, truth values of facts $\mathtt{o_i}$ and $\mathtt{w_i}$ in any model for $\mathcal{L}(\mathcal{P})$ univocally define an outlier and a witness set for it, respectively.

Rules inserted in Step 4 serve the purpose of introducing in the program $\mathcal{L}(\mathcal{P})$ the rewriting $P^{\mathrm{obs}}_{\mathcal{W}}[\mathtt{C1}]$ and $P^{\mathrm{obs}}_{\mathcal{O},\mathcal{W}}[\mathtt{C2}]$, in order to simulate the removal of the outlier and the witness from the set $P^{\mathrm{obs}}$ which is needed for verifying whether both conditions in Definition 1 are satisfied. For each atom $p$ in the set of observations $P^{\mathrm{obs}}$, two rules are introduced. In particular, rule 4.(a) guarantees that $\mathtt{obs_i^{C2}}$ is true in the program if it is neither an outlier nor a witness. Similarly, 4.(b) guarantees that $\mathtt{obs_i^{C1}}$ is true if $\mathtt{w_i}$ is not.

Rules inserted in Step 5 evaluate conditions (1) and (2) in Definition 1. Indeed, the atom $\mathtt{satC2}$ is evaluated true if a fact $\mathtt{obs_i^{C2}}$ is true even if assumed to belong to a witness ($\mathtt{w_i}$ true), i.e., if condition (2) in Definition 1 is satisfied in the model. Similarly $\mathtt{badC1}$ is true if $\mathtt{obs_i^{C1}}$ is true but $\mathtt{obs_i}$ belongs to a witness ($\mathtt{w_i}$ true), i.e., if condition (1) is not satisfied in the model.

Summarizing, the subprogram of $\mathcal{L}(\mathcal{P})$ built in Steps 1-5 is such that it guesses the values for each $\mathtt{o_i}$ and $\mathtt{w_i}$ and verifies that both conditions in Definition 1 are satisfied.

In order to finalize the transformation, it is necessary to add some constraints (Step 6). The rules inserted in Step 6 have the form $\mathtt{s_j} \leftarrow \mathtt{a}, \mathtt{not}\ \mathtt{s_j}$. As $\mathtt{s_j}$ does not appear in any other rule of the program $\mathcal{L}(\mathcal{P})$, then it must be false in any stable model of $\mathcal{L}(\mathcal{P})$. Thus, these rules act as constraints imposing that $\mathtt{a}$ must be false in all the models. Therefore, rules inserted in step 6.(a) impose that any $\mathtt{obs_i}$ cannot be an outlier and witness at the same time, the rule added in step 6.(b) imposes that interest is only in stable models in which $\mathtt{satC2}$ is true, and, finally, the rule added in step 6.(c) imposes that interest is only in stable models in stable models in which $\mathtt{badC1}$ is false.

The correctness of the algorithm is proved in the following theorem.

**Theorem 21.** *Let $\mathcal{P} = \langle P^{\mathrm{rls}}, P^{\mathrm{obs}} \rangle$ be a stratified rule-observation pair, and let $\mathcal{L}(\mathcal{P})$ be the rewriting obtained by the algorithm in Figure 5. Then,*

1. *for each outlier $\mathcal{O}$ with witness $\mathcal{W}$ in $\mathcal{P}$, there is a stable model $M$ of $\mathcal{L}(\mathcal{P})$ such that $\{\mathtt{obs_i} \mid \mathtt{o_i} \in M\} = \mathcal{O}$ and $\{\mathtt{obs_i} \mid \mathtt{w_i} \in M\} = \mathcal{W}$, and*
2. *for each stable model $M$ of $\mathcal{L}(\mathcal{P})$, there is an outlier $\mathcal{O}$ with witness $\mathcal{W}$ in $\mathcal{P}$, such that $\{\mathtt{obs_i} \mid \mathtt{o_i} \in M\} = \mathcal{O}$ and $\{\mathtt{obs_i} \mid \mathtt{w_i} \in M\} = \mathcal{W}$.*

*Proof.*

1. Let $\mathcal{O}$ be an outlier with witness $\mathcal{W}$ in $\mathcal{P}$. Let $M_1$ denote the stable model of the program $P(\mathcal{P})_{\mathcal{W}}$, and $M_2$ denote the stable model of the program $P(\mathcal{P})_{\mathcal{W},\mathcal{O}}$. Consider the interpretations $I^{C1} = \{\mathtt{a^{C1}} \mid \mathtt{a} \in M_1\}$, $I^{C2} = \{\mathtt{a^{C2}} \mid \mathtt{a} \in M_2\}$, $I^{guess} = \{\mathtt{o_i} \mid \mathtt{obs_i} \in \mathcal{O}\} \cup \{\overline{\mathtt{o_i}} \mid \mathtt{obs_i} \notin \mathcal{O}\} \cup \{\mathtt{w_i} \mid \mathtt{obs_i} \in \mathcal{W}\} \cup \{\overline{\mathtt{w_i}} \mid \mathtt{obs_i} \notin \mathcal{W}\}$, and $I^{obs} = \{\mathtt{obs_i^{C1}} \mid \mathtt{obs_i} \notin \mathcal{W}\} \cup \{\mathtt{obs_i^{C2}} \mid \mathtt{obs_i} \notin (\mathcal{O} \cup \mathcal{W})\}$, and let $M = I^{C1} \cup I^{C2} \cup I^{guess} \cup I^{obs} \cup \{\mathtt{satC2}\}$ be an interpretation of $\mathcal{L}(\mathcal{P})$.
   Now, it is shown that $M$ is a stable model of $\mathcal{L}(\mathcal{P})$ such that $\{\mathtt{obs_i} \mid \mathtt{o_i} \in M\} = \mathcal{O}$ and $\{\mathtt{obs_i} \mid \mathtt{w_i} \in M\} = \mathcal{W}$. To this aim, notice that by construction:
   - $\{\mathtt{obs_i} \mid \mathtt{o_i} \in M\} = \mathcal{O}$, because $\mathtt{o_i}$ is in $M$ if and only if $\mathtt{obs_i}$ is in $\mathcal{O}$; and
   - $\{\mathtt{obs_i} \mid \mathtt{w_i} \in M\} = \mathcal{W}$, because $\mathtt{w_i}$ is in $M$ if and only if $\mathtt{obs_i}$ is in $\mathcal{W}$.

   Therefore, it remains to show that $M$ is a stable model of $\mathcal{L}(\mathcal{P})$, i.e., that it is the minimal model of the positive program $\mathcal{L}(\mathcal{P})^M$. To this aim, let $P^i$ denote the program consisting of the rules added in the $i$-th step of the algorithm, and observe preliminary that $\mathcal{L}(\mathcal{P})^M$ is the program $(P^{\mathrm{rls}}[\mathtt{C1}] \cup \{\mathtt{obs_i^{C1}} \mid \mathtt{obs_i} \notin \mathcal{W}\})^M \cup (P^{\mathrm{rls}}[\mathtt{C2}] \cup \{\mathtt{obs_i^{C2}} \mid \mathtt{obs_i} \notin (\mathcal{O} \cup \mathcal{W})\})^M \cup (P^3)^M \cup (P^5)^M \cup (P^6)^M$, where (by applying the definition of reduct):

   $A:$ $(P^{\mathrm{rls}}[\mathtt{C1}] \cup \{\mathtt{obs_i^{C1}} \mid \mathtt{obs_i} \notin \mathcal{W}\})^M = (P(\mathcal{P})_{\mathcal{W}}[\mathtt{C1}])^{I^{C1}}$.
   Indeed, by definition, $P(\mathcal{P})_{\mathcal{W}}[\mathtt{C1}]$ is the program $P^{\mathrm{rls}}[\mathtt{C1}] \cup \{\mathtt{obs_i^{C1}} \mid \mathtt{obs_i} \notin \mathcal{W}\}$. Moreover, all the predicates in such a program have the form $\mathtt{p^{C1}}$, where $\mathtt{p}$ is a predicate symbol in $P(\mathcal{P})$. Therefore, $(P(\mathcal{P})_{\mathcal{W}}[\mathtt{C1}])^M = (P(\mathcal{P})_{\mathcal{W}}[\mathtt{C1}])^{I^{C1} \cup \{\mathtt{obs_i^{C1}} \mid \mathtt{obs_i} \notin \mathcal{W}\}}$, and the result follows because $\{\mathtt{obs_i^{C1}} \mid \mathtt{obs_i} \notin \mathcal{W}\}$ is a subset of $I^{C1}$. To see why the last containment holds, recall that $I^{C1}$ is a renaming of the model $M_1$ which must contain all the observations that are not in $\mathcal{W}$, by construction.

   $B:$ $(P^{\mathrm{rls}}[\mathtt{C2}] \cup \{\mathtt{obs_i^{C2}} \mid \mathtt{obs_i} \notin (\mathcal{W} \cup \mathcal{O})\})^M = (P(\mathcal{P})_{\mathcal{W},\mathcal{O}}[\mathtt{C2}])^{I^{C2}}$.
   Indeed, it can be applied the same line of reasoning of point $A$ above.

   $C:$ $(P^3)^M = \{\mathtt{p.} \mid \mathtt{p}$ is an atom in $I^{guess}\}$.

   $D:$ $(P^5)^M = \{\mathtt{badC1} \leftarrow \mathtt{w_i}, \mathtt{obs_i^{C1}}. \mid \mathtt{obs_i} \in P^{\mathrm{obs}}\} \cup \{\mathtt{satC2} \leftarrow \mathtt{w_i}, \mathtt{obs_i^{C2}}.\}$.

   $E:$ $(P^6)^M = \{\mathtt{s_1} \leftarrow \mathtt{o_i}, \mathtt{w_i}. \mid \mathtt{obs_i} \in P^{\mathrm{obs}}\} \cup \{\mathtt{s_3} \leftarrow \mathtt{badC1}.\}$.

   Then, the result follows because of the following two properties:

   **Property $P_1$:** $M$ is model of $\mathcal{L}(\mathcal{P})^M$.

   *Proof.* All the rules in $\mathcal{L}(\mathcal{P})^M$ (see points $A - E$ above) are satisfied by $M$:

– Rules in $A$ are satisfied by $M$. Indeed, program $P(\mathcal{P})_{\mathcal{W}}[\mathtt{C1}]$ coincides with $P(\mathcal{P})_{\mathcal{W}}$ modulo a renaming of the predicate symbols, and $I^{C1}$ is in fact a renaming of the unique stable model $M_1$ of $P(\mathcal{P})_{\mathcal{W}}$, by construction. Then, the result follows since $M_1$ is, by definition, the minimal model of the positive program $P(\mathcal{P})_{\mathcal{W}}^{M_1}$.

– Rules in $B$ are satisfied by $M$. Indeed, program $P(\mathcal{P})_{\mathcal{W},\mathcal{O}}[\mathtt{C2}]$ coincides with $P(\mathcal{P})_{\mathcal{W},\mathcal{O}}$ modulo a renaming of the predicate symbols, and $I^{C2}$ is in fact a renaming of the unique stable model $M_2$ of $P(\mathcal{P})_{\mathcal{W},\mathcal{O}}$, by construction. Then, the result follows since $M_2$ is, by definition, the minimal model of the positive program $P(\mathcal{P})_{\mathcal{W},\mathcal{O}}^{M_2}$.

– Rules in $C$ are satisfied by $M$ since $I^{guess}$ is a subset of $M$.

– Rules in $D$ are satisfied by $M$. Indeed, as for rules of the form $\{\mathtt{badC1} \leftarrow \mathtt{w_i}, \mathtt{obs_i^{C1}}. \mid \mathtt{obs_i} \in P^{\mathrm{obs}}\}$, notice that, by construction of $I^{guess}$, $\mathtt{w_i}$ is in $M$ if and only if $\mathtt{obs_i} \in \mathcal{W}$. Moreover, it is claimed that $\mathtt{obs_i^{C1}}$ is in $M$ if and only if $\mathtt{obs_i} \notin \mathcal{W}$, thereby having that the body of all such rules is always evaluated false by $M$. To see why the claim holds, observe that if $\mathtt{obs_i} \notin \mathcal{W}$ then $\mathtt{obs_i^{C1}}$ is in $M$ by construction. Let us now assume that there is a fact $\mathtt{obs_i}$ in $\mathcal{W}$ and, for the sake of contradiction, that $\mathtt{obs_i^{C1}}$ is in $M$ as well. It follows that $\mathtt{obs_i^{C1}}$ is in $I^{C1}$ and, hence, that $\mathtt{obs_i}$ belongs to $M_1$ by construction of the set $I^{C1}$. But this is impossible because $\mathcal{W}$ is an outlier and, therefore, is such that all the facts in the witness set do not occur in the unique stable model $M_1$ of $P(\mathcal{P})_{\mathcal{W}}$.
Moreover, as for the rules of the form $\{\mathtt{satC2} \leftarrow \mathtt{w_i}, \mathtt{obs_i^{C2}}.\}$, they are satisfied by $M$ because $\mathtt{satC2}$ is in $M$.

– Rules in $E$ are satisfied by $M$. Indeed, rules in the set $\{\mathtt{s_1} \leftarrow \mathtt{o_i}, \mathtt{w_i}. \mid \mathtt{obs_i} \in P^{\mathrm{obs}}\}$ are satisfied by construction of the set $I^{guess}$, which is in fact such that $\mathtt{w_i}$ is in $M$ if and only if $\mathtt{obs_i} \in \mathcal{W}$, and such that $\mathtt{o_i}$ is in $M$ if and only if $\mathtt{obs_i} \in \mathcal{O}$. Then, given that by definition $\mathcal{W} \cap \mathcal{O} = \emptyset$, the body of such rules is always evaluated false in $M$. To conclude, rule $\mathtt{s_3} \leftarrow \mathtt{badC1}.$ is satisfied by $M$ because $\mathtt{badC1}$ is not in $M$.

**Property $P_2$:** There is no model $M'$ of $\mathcal{L}(\mathcal{P})^M$ such that $M' \subset M$.

*Proof.* Recall that program $\mathcal{L}(\mathcal{P})^M$ has the form shown in points $A - E$ above and assume, for the sake of contradiction, that there exists a model $M' \subset M$ for $\mathcal{L}(\mathcal{P})^M$. First, observe that $I^{guess} \subseteq M'$ (see point $C$). Moreover, because of the fact that $\{\mathtt{obs_i^{C1}} \mid \mathtt{obs_i} \notin \mathcal{W}\}$ is a subset of $I^{C1}$ and that $\{\mathtt{obs_i^{C2}} \mid \mathtt{obs_i} \notin (\mathcal{W} \cup \mathcal{O})\}$ is a subset of $I^{C2}$ (points $A$ and $B$, respectively), it can be observed that $M$ has in fact the form $I^{C1} \cup I^{C2} \cup I^{guess} \cup \{\mathtt{satC2}\}$. Thus, the following scenarios can be distinguished:

(a) Assume that $I^{C1} \cap M' \subset I^{C1}$. Then, $I^{C1} \cap M'$ is a model for $(P^{\mathrm{rls}}[\mathtt{C1}] \cup \{\mathtt{obs_i^{C1}} \mid \mathtt{obs_i} \notin \mathcal{W}\})^{I^{C1}}$, i.e., for $(P(\mathcal{P})_{\mathcal{W}}[\mathtt{C1}])^{I^{C1}}$. It follows that the set $M'_1 = \{\mathtt{a} \mid \mathtt{a^{C1}} \in (I^{C1} \cap M')\}$ is a model for $P(\mathcal{P})_{\mathcal{W}}^{M_1}$ as well (notice that $(P(\mathcal{P})_{\mathcal{W}}[\mathtt{C1}])^{I^{C1}}$ coincides with $P(\mathcal{P})_{\mathcal{W}}^{M_1}$ modulo the renaming of predicate symbols). However, $M'_1$ is also a subset of $M_1$, and this is impossible because $M_1$ is the stable model of $P(\mathcal{P})_{\mathcal{W}}$ and, therefore, the minimal model of $P(\mathcal{P})_{\mathcal{W}}^{M_1}$.

(b) Assume that $I^{C2} \cap M' \subset I^{C2}$. Then, $I^{C2} \cap M'$ is a model for $(P^{\mathrm{rls}}[\mathtt{C2}] \cup \{\mathtt{obs_i^{C2}} \mid \mathtt{obs_i} \notin (\mathcal{O} \cup \mathcal{W})\})^{I^{C2}}$, i.e., for $(P(\mathcal{P})_{\mathcal{W},\mathcal{O}}[\mathtt{C2}])^{I^{C2}}$. It follows that the set $M'_2 = \{\mathtt{a} \mid \mathtt{a^{C2}} \in (I^{C2} \cap M')\}$ is also a model for $P(\mathcal{P})_{\mathcal{W},\mathcal{O}}^{M_2}$ (notice that $(P(\mathcal{P})_{\mathcal{W},\mathcal{O}}[\mathtt{C2}])^{I^{C2}}$ coincides with $P(\mathcal{P})_{\mathcal{W},\mathcal{O}}^{M_2}$ modulo the renaming of predicate symbols). However, $M'_2$ is also a subset of $M_2$, and this is impossible because $M_2$ is the stable model of $P(\mathcal{P})_{\mathcal{W},\mathcal{O}}$ and, therefore, the minimal model of $P(\mathcal{P})_{\mathcal{W},\mathcal{O}}^{M_2}$.

(c) Assume that $\mathtt{satC2}$ is not in $M'$. Then, after (a) and (b) above, it is the case that $M' = I^{C1} \cup I^{C2} \cup I^{guess}$. Then, since $\mathtt{satC2}$ is not in $M'$, because of the rules added in Step 5.(b), it holds that for each $\mathtt{w_i}$ in $M'$, $\mathtt{obs_i^{C2}}$ is not in $M'$. Hence, by construction of $I^{guess}$, it follows that for each $\mathtt{obs_i}$ in $\mathcal{W}$, it is the case that $\mathtt{obs_i^{C2}}$ is not in $I^{C2}$. However, $I^{C2}$ is a renaming of the model $M_2$ of $P(\mathcal{P})_{\mathcal{W},\mathcal{O}}$. Therefore, the model $M_2$ does not entail any fact in $\mathcal{W}$. But this is a contradiction with condition (2) in Definition 1.

2. Let $M$ be a stable model of $\mathcal{L}(\mathcal{P})$. First of all, note that by rules inserted into $\mathcal{L}(\mathcal{P})$ in Step 6 of algorithm, $M$ is such that ($i$) for each letter $\mathtt{obs_i}$ in $P^{\mathrm{obs}}$, $\mathtt{o_i}$ and $\mathtt{w_i}$ cannot belong simultaneously to $M$, ($ii$) $\mathtt{satC2} \in M$, and ($iii$) $\mathtt{badC1} \notin M$. Furthermore, by rules inserted into $\mathcal{L}(\mathcal{P})$ in Steps 3.(a) and 3.(b) it is the case that, for each letter $\mathtt{obs_i}$ in $P^{\mathrm{rls}}$, either $\mathtt{o_i}$ or $\overline{\mathtt{o}}_\mathtt{i}$ and either $\mathtt{w_i}$ or $\overline{\mathtt{w}}_\mathtt{i}$ belong to $M$.
Consider, now, the disjoint sets $\mathcal{O} = \{\mathtt{obs_i} \mid \mathtt{o_i} \in M\}$ and $\mathcal{W} = \{\mathtt{obs_i} \mid \mathtt{w_i} \in M\}$. It has to be shown that both conditions in Definition 1 are satisfied. To this aim consider the interpretation $I^{guess} = \{\mathtt{o_i} \mid \mathtt{obs_i} \in \mathcal{O}\} \cup \{\overline{\mathtt{o}}_\mathtt{i} \mid \mathtt{obs_i} \notin \mathcal{O}\} \cup \{\mathtt{w_i} \mid \mathtt{obs_i} \in \mathcal{W}\} \cup \{\overline{\mathtt{w}}_\mathtt{i} \mid \mathtt{obs_i} \notin \mathcal{W}\}$, and notice that $M$ can be written as $I^{C1} \cup I^{C2} \cup I^{guess} \cup \{\mathtt{satC2}\}$, where $I^{C1}$ and $I^{C2}$ are the subsets of $M$ containing all the predicates of the form $\mathtt{p^{C1}}$ and $\mathtt{p^{C2}}$, respectively. At this point, the reader may check that rules in $\mathcal{L}(\mathcal{P})^M$ have again the form illustrated in points $A - E$ above.
Given that $M$ is a minimal model of $\mathcal{L}(\mathcal{P})^M$, it follows that $I^{C1}$ is a minimal model of $(P^{\mathrm{rls}}[\mathtt{C1}] \cup \{\mathtt{obs_i^{C1}} \mid \mathtt{obs_i} \notin \mathcal{W}\})^M = (P(\mathcal{P})_{\mathcal{W}}[\mathtt{C1}])^{I^{C1}}$, and $I^{C2}$ is a minimal model of $(P^{\mathrm{rls}}[\mathtt{C2}] \cup \{\mathtt{obs_i^{C2}} \mid \mathtt{obs_i} \notin (\mathcal{W} \cup \mathcal{O})\})^M = (P(\mathcal{P})_{\mathcal{W},\mathcal{O}}[\mathtt{C2}])^{I^{C2}}$. Then, $M_1 = \{\mathtt{a} \mid \mathtt{a^{C1}} \in I^{C1}\}$ is a minimal model of $P_{\mathcal{W}}^{M_1}$ and $M_2 = \{\mathtt{a} \mid \mathtt{a^{C2}} \in I^{C2}\}$ is a minimal model of $P_{\mathcal{W},\mathcal{O}}^{M_2}$. Therefore, $M_1$ (resp., $M_2$) is the stable model of $P_{\mathcal{W}}$ (resp., $P_{\mathcal{W},\mathcal{O}}$).
To conclude, the following properties can be shown:
   - $P(\mathcal{P})_{\mathcal{W}} \models \neg\mathcal{W}$.

     *Proof.* Assume, for the sake of contradiction, that there is a fact $\mathtt{obs_i}$ in $\mathcal{W}$ such that $P(\mathcal{P})_{\mathcal{W}} \models \mathtt{obs_i}$. Then, $\mathtt{obs_i}$ must belong to the unique stable model $M_1$, and $\mathtt{obs_i^{C1}}$ must belong to $I^{C1}$. However, $\mathtt{w_i}$ belongs to $M$ by construction, and therefore by rule 5.(a) $\mathtt{badC1}$ is in $M$, which is impossible.

   - $P(\mathcal{P})_{\mathcal{W},\mathcal{O}} \not\models \neg\mathcal{W}$.

     *Proof.* Assume, for the sake of contradiction, that for each fact $\mathtt{obs_i}$ in $\mathcal{W}$, $P(\mathcal{P})_{\mathcal{W},\mathcal{O}} \models \neg\mathtt{obs_i}$. Then, the unique stable model $M_2$ does not contain any fact in $\mathcal{W}$, and symmetrically $I^{C2}$ does not contain any fact of the form $\mathtt{obs_i^{C2}}$, with $\mathtt{obs_i}$ in $\mathcal{W}$. Then, $\mathtt{satC2}$ is not in $M$ because of the rule 5.(b), which is impossible. □

**Minimum-size Outliers** In order to translate detection problems aiming at singling out minimum-size outliers into a suitable logic program, an approach will be exploited which was used, for instance, in the DLV system and relying on extending classic logic programming by introducing *weak constraints*.

Weak constraints, taking the form of rules such as $:\sim\ \mathtt{b_1}, \cdots, \mathtt{b_k}, \mathtt{not}\ \mathtt{b_{k+1}}, \cdots, \mathtt{not}\ \mathtt{b_{k+m}}$, express a set of desired conditions that may be however violated; their informal semantics is to minimize the number of violated instances. In fact, in [17] it is proved that the introduction of weak constraints allows the solution of optimization problems since each weak constraint can be regarded as an "objective function" of an optimization problem.

Given a program $P \cup W$ where $P$ is a set of rules and $W$ a set of weak constraints, an interpretation $M$ is a stable model for $P \cup W$ if $M$ is a stable model for $P$. The stable models of $P \cup W$ are ordered w.r.t. the number of weak constraints that are not satisfied: *best stable models* are those which minimize such a number [17].

*Example 6.* Given a graph $G = \langle V, E\rangle$, denoted by the unary predicate $node$ and the binary predicate $edge$, we can model the MAX_CLIQUE problem, asking for the clique of $G$ having maximum size, by means of the following program:

$$\begin{array}{l}
\mathtt{c(X)} \leftarrow \mathtt{not\ nc(X),\ node(X).} \\
\mathtt{nc(X)} \leftarrow \mathtt{not\ c(X),\ node(X).} \\
\mathtt{p} \leftarrow \mathtt{c(X), c(Y), X \neq Y, not\ edge(X, Y), not\ p.} \\
:\sim\ \mathtt{nc(X).}
\end{array}$$

where the first two rules are used for creating all the possible partitions of nodes into c and nc, the third one is used for ensuring that nodes in c forms a clique, i.e., each pair of nodes in c is connected by an edge, while the weak constraint minimizes the number of vertices that are not in the clique, or equivalently it maximizes the size of the clique. Then, the best stable models are in one-to-one correspondence with maximum-size cliques in $G$. ◁

Thus, the algorithm in Figure 5 can be modified by inserting the constraint $:\sim\ \mathtt{o_i}.$ into $\mathcal{L}(\mathcal{P})$, for each $\mathtt{obs_i} \in P^{\rm obs}$. Then, letting $\mathcal{L}^{\sim}(\mathcal{P})$ be the transformed program resulting from applying the modified algorithm, we have that minimum-size outliers in $\mathcal{P}$ are in one-to-one correspondence with best stable models of $\mathcal{L}^{\sim}(\mathcal{P})$.

**Theorem 22.** *Let $\mathcal{P} = \langle \mathcal{P}^{\rm rls}, \mathcal{P}^{\rm obs}\rangle$ be a stratified rule-observation pair. Then,*

1. *for each minimum-size outlier $\mathcal{O}$ with witness $\mathcal{W}$ in $\mathcal{P}$, there is a best stable model $\mathcal{M}$ of $\mathcal{L}^{\sim}(\mathcal{P})$ such that $\{\mathtt{obs_i} \mid \mathtt{o_i} \in M\} = \mathcal{O}$ and $\{\mathtt{obs_i} \mid \mathtt{w_i} \in M\} = \mathcal{W}$, and*
2. *for each best stable model $\mathcal{M}$ of $\mathcal{L}^{\sim}(\mathcal{P})$, there is an outlier $\mathcal{O}$ with witness $\mathcal{W}$ in $\mathcal{P}$, such that $\{\mathtt{obs_i} \mid \mathtt{o_i} \in M\} = \mathcal{O}$ and $\{\mathtt{obs_i} \mid \mathtt{w_i} \in M\} = \mathcal{W}$.*

*Proof.* Given a rule-observation pair $\mathcal{P}$, $\mathcal{L}^{\sim}(\mathcal{P}) = \mathcal{L}(\mathcal{P}) \cup \{:\sim\ \mathtt{o_i}. \mid \mathtt{obs_i} \in P^{\rm obs}\}$, where the stable models of the program $\mathcal{L}(\mathcal{P})$ are in one-to-one correspondence with outliers in $\mathcal{P}$ (Theorem 5).

By definition, each model $M$ in $\mathcal{SM}(\mathcal{L}(\mathcal{P}))$ is also a model of $\mathcal{L}^{\sim}(\mathcal{P})$. Moreover, $M$ is a best model if it minimizes the number of violated constraints, i.e., if there is no other model $M'$ containing a fewer number of atoms of the form $\mathtt{o_i}$. The result follows by noticing that this number is, in fact, the size of the outlier $\mathcal{O}$ constructed in the proof of Theorem 5. □

### 6.2 General Pairs

Next let us consider the case of general rule-observation pairs. In this case, it should be pointed out the constraint imposed by rule 6.(c) would not suffice for ensuring the satisfaction of condition (1) in Definition 1. Indeed, the outlier detection problems turned out to be complete for the second level of the polynomial hierarchy for general rule-observation pairs, while any polynomial time transformation into a logic program under stable model semantics may encode problems complete for the first level only.

In order to deal with this problem, a more powerful logic formalism must be exploited, while keeping the size of the encoding polynomially-bounded in the size of the rule-observation pair. Specifically, the solution relies on a rewriting into a *disjunctive* logic program accounting for outliers under the cautious semantics. Actually, a similar rewriting for brave semantics can be obtained as well.

Recall that disjunctive programs allow clauses to have both disjunction (denoted by $\vee$) in their heads and negation in their bodies. In more detail, a *disjunctive rule* $r$ is a clause of the form: $\mathtt{a_1} \vee \cdots \vee \mathtt{a_m} \leftarrow \mathtt{b_1}, \cdots, \mathtt{b_k}, \mathtt{not\ c_1}, \cdots, \mathtt{not\ c_n}$ where $n, k, m \geq 0$, $n + k + m > 0$ and $\mathtt{a_1}, \cdots, \mathtt{a_m}, \mathtt{b_1}, \cdots, \mathtt{b_k}, \mathtt{c_1}, \cdots, \mathtt{c_n}$ are atoms. The disjunction $\mathtt{a_1} \vee \cdots \vee \mathtt{a_m}$, also denoted by $\mathbf{h}(r)$, is the *head* of $r$, while the conjunction $\mathtt{b_1}, \ldots, \mathtt{b_k}, \mathtt{not}\ \mathtt{c_1}, \cdots, \mathtt{not}\ \mathtt{c_n}$, also denoted by $\mathbf{b}(r)$, is the *body* of $r$.

Given an interpretation $I$ for a disjunctive program $P$, the value of the disjunction $D = \mathtt{a_1} \vee \cdots \vee \mathtt{a_m}$ w.r.t. $I$ is $value_I(D) = max(\{value_I(\mathtt{a_i}) \mid 1 \leq i \leq n\})$. Similarly to the case of non-disjunctive programs, a ground rule $r$ is *satisfied* by $I$ if $value_I(\mathbf{h}(r)) \geq value_I(\mathbf{b}(r))$ and models of $P$ are defined to be the interpretations that satisfy all the ground rules of $P$. Then, the model-theoretic semantics of disjunctive programs is defined as follows.

For a *positive* disjunctive program $P$ (i.e., rules in $P$ do not contain negation in the body), the semantics is given in terms of the set its minimal models, denoted by $\mathcal{MM}(P)$[7]. Moreover, for a general disjunctive program $P$, the stable model semantics [42] assigns to $P$ the set $\mathcal{SM}(P)$ of its *stable models* defined as suitable extension of stable models for disjunction-free programs. In particular, let $P$ be a disjunctive logic program and let $I$ be an interpretation for $P$. Then, $P^I$ denotes the ground positive program derived from $ground(P)$ by (1) removing all rules that contain a negative literal $\mathtt{not\ a}$ in the body and $\mathtt{a} \in I$, and (2) removing all negative literals from the remaining rules. An interpretation $M$ is a stable model for $P$ if and only if $M \in \mathcal{MM}(P^M)$. Under this semantics, disjunctive programs allow to solve problems that are complete for the complexity class $\Sigma_2^P$ (see, e.g., [24]).

The algorithm *OutlierDetectionToDisjunctiveStableModels* is shown in Figure 6. To illustrate, step 2 inserts two suitable rewriting of $P^{\mathrm{rls}}$ into $\mathcal{L}^{\vee}(P)$. One of the two rewriting consists of a renaming of the original program $P^{\mathrm{rls}}$, call it $P^{\mathrm{rls}}[\mathtt{C2}]$ (see rules 2.(b)). Analogously to the rewriting shown for stratified pairs, $P^{\mathrm{rls}}[\mathtt{C2}]$ serves the purpose of checking condition (2) of Definition 1.

Instead, the task of checking condition (1) is demanded to rules inserted in Steps 2.(a), 4.(b), 7, 8, and 9 — recall that, under the cautious form of entailing, this check consists in verifying that each stable model of $P(\mathcal{P})_{\mathcal{W}}$ does not contain some atom in $\mathcal{W}$. Intuitively this is carried out as follows. First, in order to encode the stable models of the program $P(\mathcal{P})_{\mathcal{W}}$, for each atom $\mathtt{p}$ occurring in $P(\mathcal{P})$, the atoms $\mathtt{p}^{\mathtt{C1}}$ and $\overline{\mathtt{p}}^{\mathtt{C1}}$ are used: the atom $\mathtt{p}^{\mathtt{C1}}$ (resp. $\overline{\mathtt{p}}^{\mathtt{C1}}$) being true in an interpretation of $\mathcal{L}^{\vee}(P)$ means that the atom $\mathtt{p}$ is true (resp. false) in a stable model of $P(\mathcal{P})_{\mathcal{W}}$. Then, the atom $\mathtt{satC1}$ is used to check whether the truth values for the atoms of the form $\mathtt{p}^{\mathtt{C1}}$ and $\overline{\mathtt{p}}^{\mathtt{C1}}$ correctly encode a stable model for $P(\mathcal{P})_{\mathcal{W}}$ and satisfy condition (1) in Definition 1. Specifically, by rule 2.(a), $\mathtt{satC1}$ is entailed by the rewriting as soon as the atoms of the form $\mathtt{p}^{\mathtt{C1}}$ and $\overline{\mathtt{p}}^{\mathtt{C1}}$ do not satisfy some rule in $P(\mathcal{P})_{\mathcal{W}}$. Clearly, this is only a necessary condition for these atoms encoding a stable model for $P(\mathcal{P})_{\mathcal{W}}$, and details on the encoding of some further conditions are provided below.

Steps 3 and 4 of algorithm *OutlierDetectionToDisjunctiveStableModels* are similar to Steps 3 and 4 of algorithm *OutlierDetectionToStableModels* described above. Specifically, rule 4.(b) entails $\mathtt{satC1}$ whenever an atom of the form $\overline{\mathtt{obs_i}}^{\mathtt{C1}}$ comes true in the stable model of $P(\mathcal{P})_{\mathcal{W}}$ while not being part of the witness set. Therefore, this rule guarantees that the truth values for the atoms of the form $\mathtt{p}^{\mathtt{C1}}$ and $\overline{\mathtt{p}}^{\mathtt{C1}}$ consistently encode the observations that do not belong to the witness set.

Rules inserted in subsequent Step 5 evaluate the conditions of Definition 1. Indeed, the atom $\mathtt{satC2}$ is true if a fact $\mathtt{obs_i^{C2}}$ is true even if assumed to belong to a witness ($\mathtt{w_i}$ true), i.e. if condition (2) in Definition 1 is satisfied in the model, while the atom $\mathtt{satC1}$ is true if, for each fact $\mathtt{obs_i} \in P^{\mathrm{obs}}$, either (step 5.(a)) $\mathtt{obs_i}$ is not assumed to belong to a witness ($\overline{\mathtt{w}}_\mathtt{i}$ true) or (step 5.(b)) $\mathtt{obs_i}$ is assumed to belong to a witness ($\mathtt{w_i}$ true) and $\overline{\mathtt{obs_i}}^{\mathtt{C1}}$ is true, i.e. if the witness set is not entailed in the model (cf. condition (1) in Definition 1).

[7] Differently from disjunction-free positive programs, positive disjunctive programs have more than one minimal model. Hence, for simplicity and by a little abuse of notation, in the following $\mathcal{MM}(P)$ will denote a set of minimal models rather than a single minimal model.

**Input:** A rule-observation pair $\mathcal{P} = \langle P^{\rm rls}, P^{\rm obs}\rangle$, where $P^{\rm obs} = \{\mathtt{obs_1}, ..., \mathtt{obs_n}\}$;
**Output:** A disjunctive logic program $\mathcal{L}^{\vee}(\mathcal{P})$;
**Method:** Perform the following steps:

1. $\mathcal{L}(\mathcal{P}) := \emptyset$;
2. /*——— Rule part rewriting ———*/
   **for each** rule $r \in P^{\rm rls}$ of the form $\mathtt{a} \leftarrow \mathtt{b_1}, \cdots, \mathtt{b_k}, \mathtt{not}\ \mathtt{c_1}, \cdots, \mathtt{not}\ \mathtt{c_m}$, **insert into** $\mathcal{L}^{\vee}(\mathcal{P})$ the rule
   (a) $\mathtt{satC1} \leftarrow \overline{\mathtt{a}}^{\mathtt{C1}}, \mathtt{b}_1^{\mathtt{C1}}, \cdots, \mathtt{b}_k^{\mathtt{C1}}, \overline{\mathtt{c}}_1^{\mathtt{C1}}, \cdots, \overline{\mathtt{c}}_m^{\mathtt{C1}}$.
   (b) $\mathtt{a}^{\mathtt{C2}} \leftarrow \mathtt{b}_1^{\mathtt{C2}}, \cdots, \mathtt{b}_k^{\mathtt{C2}}, \mathtt{not}\ \mathtt{c}_1^{\mathtt{C2}}, \cdots, \mathtt{not}\ \mathtt{c}_m^{\mathtt{C2}}$.
3. /*——— Outlier and witness guessing ———*/
   **for each** $\mathtt{obs_i} \in P^{\rm obs}$, **insert into** $\mathcal{L}^{\vee}(\mathcal{P})$ the rules
   (a) $\mathtt{o_i} \leftarrow \mathtt{not}\ \overline{\mathtt{o}}_\mathtt{i}$. $\quad \overline{\mathtt{o}}_\mathtt{i} \leftarrow \mathtt{not}\ \mathtt{o_i}$.
   (b) $\mathtt{w_i} \leftarrow \mathtt{not}\ \overline{\mathtt{w}}_\mathtt{i}$. $\quad \overline{\mathtt{w}}_\mathtt{i} \leftarrow \mathtt{not}\ \mathtt{w_i}$.
4. /*——— Observations definition ———*/
   **for each** $\mathtt{obs_i} \in P^{\rm obs}$, **insert into** $\mathcal{L}^{\vee}(\mathcal{P})$ the rules
   (a) $\mathtt{obs}_\mathtt{i}^{\mathtt{C2}} \leftarrow \mathtt{not}\ \mathtt{o_i},\ \mathtt{not}\ \mathtt{w_i}$.
   (b) $\mathtt{satC1} \leftarrow \overline{\mathtt{obs}}_\mathtt{i}^{\mathtt{C1}}, \overline{\mathtt{w}}_\mathtt{i}$.
5. /*——— Outlier and witness checking ———*/
   **for each** $\mathtt{obs_i} \in P^{\rm obs}$, **insert into** $\mathcal{L}^{\vee}(\mathcal{P})$ the rules
   (a) $\mathtt{satC1_i} \leftarrow \overline{\mathtt{w}}_\mathtt{i}$.
   (b) $\mathtt{satC1_i} \leftarrow \mathtt{w_i}, \overline{\mathtt{obs}}_\mathtt{i}^{\mathtt{C1}}$.
   (c) $\mathtt{satC2} \leftarrow \mathtt{w_i},\ \mathtt{obs}_\mathtt{i}^{\mathtt{C2}}$.
   **insert into** $\mathcal{L}^{\vee}(\mathcal{P})$ the rule $\mathtt{satC1} \leftarrow \mathtt{satC1_1}, \ldots, \mathtt{satC1_n}$.
6. /*——— Constraints ———*/
   (a) **insert into** $\mathcal{L}^{\vee}(\mathcal{P})$ the rule $\mathtt{satC1} \leftarrow \mathtt{not}\ \mathtt{satC1}$.
   (b) **insert into** $\mathcal{L}^{\vee}(\mathcal{P})$ the rule $\mathtt{satC2} \leftarrow \mathtt{not}\ \mathtt{satC2}$.
   (c) **for each** $\mathtt{obs_i} \in P^{\rm obs}$, **insert into** $\mathcal{L}^{\vee}(\mathcal{P})$ the rule $\mathtt{s} \leftarrow \mathtt{o_i},\ \mathtt{w_i},\ \mathtt{not}\ \mathtt{s}$.
7. /*——— Checking Condition (1): guessing an interpretation, mapping $\phi$ and rule assignment ———*/
   **for each** atom $\mathtt{p} \in P^{\rm rls} \cup P^{\rm obs}$, **insert into** $\mathcal{L}^{\vee}(\mathcal{P})$ the rules
   (a) $\mathtt{p}^{\mathtt{C1}} \vee \overline{\mathtt{p}}^{\mathtt{C1}}$.
   (b) $\mathtt{p}\phi^1 \vee ... \vee \mathtt{p}\phi^\mathtt{s} \leftarrow \mathtt{p}^{\mathtt{C1}}$. (where $\mathtt{s}$ is the number of predicate symbols in $\mathrm{Lit}(P^{\rm rls}) \cup P^{\rm obs}$)
   (c) $\mathtt{p}r^1 \vee ... \vee \mathtt{p}r^\ell \leftarrow \mathtt{p}^{\mathtt{C1}}$. (where $r^1, ..., r^\ell$ are the rules in $P^{\rm rls} \cup P^{\rm obs}$ in which $\mathtt{p}$ occurs in the head)
8. /*——— Checking Condition (1): constraints ———*/
   **for each** atom $\mathtt{p}$ in $P^{\rm rls} \cup P^{\rm obs}$ **insert into** $\mathcal{L}^{\vee}(\mathcal{P})$ the rules
   (a) $\mathtt{satC1} \leftarrow \mathtt{p}^{\mathtt{C1}}, \overline{\mathtt{p}}^{\mathtt{C1}}$.
   (b) $\mathtt{satC1} \leftarrow \mathtt{p}\phi^\mathtt{i}, \mathtt{p}\phi^\mathtt{j}, \mathtt{p}^{\mathtt{C1}}$. (for each $\mathtt{i}, \mathtt{j} \in \{1, ..., \mathtt{s}\}$, with $\mathtt{i} \neq \mathtt{j}$)
   (c) $\mathtt{satC1} \leftarrow \mathtt{p}r^\mathtt{i}, \mathtt{p}r^\mathtt{j}, \mathtt{p}^{\mathtt{C1}}$. (for each $\mathtt{i}, \mathtt{j} \in \{1, ..., \ell\}$, with $\mathtt{i} \neq \mathtt{j}$)
   (d) $\mathtt{satC1} \leftarrow \mathtt{p}r^\mathtt{i}, \mathtt{w_j}, \mathtt{p}^{\mathtt{C1}}$. (if $r^\mathtt{i}$ is a fact in $P^{\rm obs}$ asserting the atom $\mathtt{obs_j} = \mathtt{p}$)
   (e) $\mathtt{satC1} \leftarrow \mathtt{p}r^\mathtt{i}, \mathtt{c}^{\mathtt{C1}}, \mathtt{p}^{\mathtt{C1}}$. (for each atom $\mathtt{c}$ occurring negatively in the body of $r^\mathtt{i}$)
   (f) $\mathtt{satC1} \leftarrow \mathtt{p}r^\mathtt{i}, \overline{\mathtt{b}}^{\mathtt{C1}}, \mathtt{p}^{\mathtt{C1}}$. (for each atom $\mathtt{b}$ occurring positively in the body of $r^\mathtt{i}$)
   (g) $\mathtt{satC1} \leftarrow \mathtt{p}r^\mathtt{i}, \mathtt{p}\phi^\mathtt{h}, \mathtt{q}\phi^\mathtt{k}, \mathtt{p}^{\mathtt{C1}}$. (for each atom $\mathtt{q}$ occurring positively in the body of $r^\mathtt{i}$, and for each $\mathtt{h} \leq \mathtt{k}$)
9. /*——— Checking Condition (1): saturation ———*/
   **for each** atom $\mathtt{p}$ in $P^{\rm rls} \cup P^{\rm obs}$ **insert into** $\mathcal{L}^{\vee}(\mathcal{P})$ the rules
   (a) $\mathtt{p}^{\mathtt{C1}} \leftarrow \mathtt{satC1}$.
   (b) $\overline{\mathtt{p}}^{\mathtt{C1}} \leftarrow \mathtt{satC1}$.
   (c) $\mathtt{p}\phi^\mathtt{i} \leftarrow \mathtt{satC1}$. (for each $\mathtt{i} \in \{1, ..., \mathtt{s}\}$)
   (d) $\mathtt{p}r^\mathtt{i} \leftarrow \mathtt{satC1}$. (for each rule $r^\mathtt{i}$ in which $\mathtt{p}$ occurs in the head)
   **insert into** $\mathcal{L}^{\vee}(\mathcal{P})$ the rule
   (e) $\mathtt{satC1_i} \leftarrow \mathtt{satC1}$. (for each atom of the form $\mathtt{satC1_i}$)

**Fig. 6.** Algorithm *OutlierDetectionToDisjunctiveStableModels*.

Step 6 add rules which are similar to the constraints of the algorithm in Figure 5. In fact, these rules impose that the interest is in stable models in which both `satC1` and `satC2` are true, and that $\mathtt{obs_i}$ cannot belong to an outlier and a witness at the same time.

The main differences w.r.t. the case of stratified rule-observation pairs are in steps 7, 8 and 9. Indeed, unlike stratified logic programs, general logic programs may have more than one stable model and, hence, under cautious semantics, rules inserted in Step 5 do not suffice to check condition (1) in Definition 1.

Specifically, rules inserted in Step 9 are such that if a stable model $M$ of $\mathcal{L}^{\vee}(P)$ contains `satC1`, then it must also contain all the atoms of the form $\mathtt{p}^{\mathtt{C1}}$ and $\overline{\mathtt{p}}^{\mathtt{C1}}$ — actually, these rules infer also other atoms, namely $\mathtt{satC1_i}$, $p\phi^i$ and $pr^i$, whose necessity will be clear in a while. Intuitively, since by rule 6.(a) `satC1` must belong to any model of the program $\mathcal{L}^{\vee}(P)$, a necessary condition for $M$ to be a minimal model of $\mathcal{L}^{\vee}(P)^M$ is that, for each subset $M'$ of $M$ being a model of $\mathcal{L}^{\vee}(P)^M$, $M'$ entails `satC1` in its turn. Due to the rules inserted in step 7.(a) and 8.(a), each model $M' \subset M$ contains a guess of a model of $P(\mathcal{P})_{\mathcal{W}}$. Hence, by looking at rule 2.(a) and rules in Step 5, it would be concluded that the minimality of $M$ guarantees that there is no model of $P(\mathcal{P})_{\mathcal{W}}$ that does not satisfy condition (1).

However, the check that the models of $P(\mathcal{P})_{\mathcal{W}}$ do not satisfy condition (1) must be restricted to its stable models only (ignoring models that are not stable and that do not satisfy condition (1)). This is precisely the purpose of the rules in Step 7 and 8. Specifically, the intended meaning of the rules 7.(b), 7.(c) and the rules inserted in Step 8 is to check for the minimality of $M'$ (w.r.t. $\mathcal{L}^{\vee}(P)^M$) so that `satC1` is entailed whenever $M'$ is not minimal. To this aim, a well-known characterization of minimal models for positive programs is exploited (indeed, the reduct of $P(\mathcal{P})_{\mathcal{W}}$ w.r.t. an interpretation is a positive program), which is formalized below for the reader's convenience.

**Lemma 1.** *(cf. [88], Theorem 2.7) Let $P$ be a (non-disjunctive) positive propositional logic program, and let $M$ be a model for it. Then, $M$ is minimal if and only if there is a function $\phi$ assigning a natural number to each atom occurring in $P$, and a function $r$ assigning a rule of $P$ to each element in $M$ such that:*

1. $\mathbf{b}(r(p)) \subseteq M$,
2. $\mathbf{h}(r(p)) = p$, *and*
3. $\phi(q) < \phi(p)$, *for each* $q \in \mathbf{b}(r(p))$.

Accordingly, to assess the minimality of the model at hand, rules 7.(b) and 7.(c) guess for each atom `p`, an assignment to a natural number (`p` is assigned to `i` iff $\mathtt{p}\phi^{\mathtt{i}}$ is true in the model), and a rule (exactly a rule $r^{\mathtt{j}}$ having `p` occurring in its head is assigned to `p` iff $\mathtt{p}r^{\mathtt{j}}$ is true in the model), while rules added in Step 8 checks whether the assignments are correct, i.e., whether they satisfy all the conditions in Lemma 1.

The following theorem accounts for the correctness of the algorithm.

**Theorem 23.** *Let $\mathcal{P} = \langle \mathcal{P}^{\mathrm{rls}}, \mathcal{P}^{\mathrm{obs}} \rangle$ be a rule-observation pair, and let $\mathcal{L}^{\vee}(\mathcal{P})$ be the rewriting obtained by the algorithm in Figure 6. Then,*

1. *for each outlier $\mathcal{O}$ with witness $\mathcal{W}$ in $\mathcal{P}$, there is a stable model $\mathcal{M}$ of $\mathcal{L}^{\vee}(\mathcal{P})$ such that $\{\mathtt{obs_i} \mid \mathtt{o_i} \in M\} = \mathcal{O}$ and $\{\mathtt{obs_i} \mid \mathtt{w_i} \in M\} = \mathcal{W}$, and*
2. *for each stable model $M$ of $\mathcal{L}^{\vee}(\mathcal{P})$, there is an outlier $\mathcal{O}$ with witness $\mathcal{W}$ in $\mathcal{P}$, such that $\{\mathtt{obs_i} \mid \mathtt{o_i} \in M\} = \mathcal{O}$ and $\{\mathtt{obs_i} \mid \mathtt{w_i} \in M\} = \mathcal{W}$.*

*Proof.*

1. Let $\mathcal{O}$ be an outlier with witness $\mathcal{W}$ in $\mathcal{P}$. Let $M_2$ denote the stable model of the program $P(\mathcal{P})_{\mathcal{W},\mathcal{O}}$ which entails an element in $\mathcal{W}$ (notice that such a model exists in order to satisfy condition (2) in Definition 1). Consider the interpretations $I^{C1} = \{\mathtt{a}^{\mathtt{C1}}, \overline{\mathtt{a}}^{\mathtt{C1}} \mid \mathtt{a}$ is an atom in $P^{\text{rls}} \cup P^{\text{obs}}\}$, $I^{C2} = \{\mathtt{a}^{\mathtt{C2}} \mid \mathtt{a} \in M_2\}$, $I^{guess} = \{\mathtt{o_i} \mid \mathtt{obs_i} \in \mathcal{O}\} \cup \{\overline{\mathtt{o}}_\mathtt{i} \mid \mathtt{obs_i} \notin \mathcal{O}\} \cup \{\mathtt{w_i} \mid \mathtt{obs_i} \in \mathcal{W}\} \cup \{\overline{\mathtt{w}}_\mathtt{i} \mid \mathtt{obs_i} \notin \mathcal{W}\}$, $I^{obs} = \{\mathtt{obs_i^{C2}} \mid \mathtt{obs_i} \notin (\mathcal{O} \cup \mathcal{W})\}$, $I^{\phi} = \{\mathtt{p}\phi^{\mathtt{i}} \mid \mathtt{p}$ is an atom in $P^{\text{rls}} \cup P^{\text{obs}}, \mathtt{1} \leq \mathtt{i} \leq \mathtt{s}\}$ (where $\mathtt{s}$ denotes the number of distinct predicates in $P^{\text{rls}} \cup P^{\text{obs}}$), and $I^{r} = \{\mathtt{p}r^{\mathtt{j}} \mid r^{\mathtt{j}}$ is a rule in $P^{\text{rls}} \cup P^{\text{obs}}$ such that $\mathbf{h}(r^{\mathtt{j}}) = \mathtt{p}\}$, and let $M = I^{C1} \cup I^{C2} \cup I^{guess} \cup I^{obs} \cup I^{\phi} \cup I^{r} \cup \{\mathtt{satC2}, \mathtt{satC1}, \mathtt{satC1_i}, ..., \mathtt{satC1_n}\}$ be an interpretation of $\mathcal{L}^{\vee}(\mathcal{P})$.
Now, it is shown that $M$ is a stable model of $\mathcal{L}^{\vee}(\mathcal{P})$ such that $\{\mathtt{obs_i} \mid \mathtt{o_i} \in M\} = \mathcal{O}$ and $\{\mathtt{obs_i} \mid \mathtt{w_i} \in M\} = \mathcal{W}$. To this aim, notice that by construction:

- $\{\mathtt{obs_i} \mid \mathtt{o_i} \in M\} = \mathcal{O}$, because $\mathtt{o_i}$ is in $M$ if and only if $\mathtt{obs_i}$ is in $\mathcal{O}$; and
- $\{\mathtt{obs_i} \mid \mathtt{w_i} \in M\} = \mathcal{W}$, because $\mathtt{w_i}$ is in $M$ if and only if $\mathtt{obs_i}$ is in $\mathcal{W}$.

Therefore, it remains to show that $M$ is a stable model of $\mathcal{L}^{\vee}(\mathcal{P})$, i.e., that it is the minimal model of the positive program $\mathcal{L}^{\vee}(\mathcal{P})^M$. To this aim, let $P^{C1}$ denote the program composed by rules added in steps 2.(a) and 4.(b) of the algorithm, and let $P^i$ denote the program consisting of the rules added in the $i$-th step of the algorithm, and observe preliminary that $\mathcal{L}^{\vee}(\mathcal{P})^M$ is the program $(P^{C1})^M \cup (P^{\text{rls}}[\mathtt{C2}] \cup \{\mathtt{obs_i^{C2}} \mid \mathtt{obs_i} \notin (\mathcal{O} \cup \mathcal{W})\})^M \cup (P^3)^M \cup (P^5)^M \cup (P^6)^M \cup (P^7)^M \cup (P^8)^M \cup (P^9)^M$, where (by applying the definition of reduct):

$A:$ $(P^{C1})^M = P^{C1}$.
Indeed, by definition $P^{C1}$ is a positive program, and therefore coincides with its reduct.

$B:$ $(P^{\text{rls}}[\mathtt{C2}] \cup \{\mathtt{obs_i^{C2}} \mid \mathtt{obs_i} \notin (\mathcal{W} \cup \mathcal{O})\})^M = (P(\mathcal{P})_{\mathcal{W},\mathcal{O}}[\mathtt{C2}])^{I^{C2}}$.
Indeed, by definition, $P(\mathcal{P})_{\mathcal{W},\mathcal{O}}[\mathtt{C2}]$ is the program $P^{\text{rls}}[\mathtt{C2}] \cup \{\mathtt{obs_i^{C2}} \mid \mathtt{obs_i} \notin (\mathcal{W} \cup \mathcal{O})\}$. Moreover, all the predicates in this program have the form $\mathtt{p}^{\mathtt{C2}}$, where $\mathtt{p}$ is a predicate symbol in $P(\mathcal{P})$. Therefore, $(P(\mathcal{P})_{\mathcal{W},\mathcal{O}}[\mathtt{C2}])^M = (P(\mathcal{P})_{\mathcal{W},\mathcal{O}}[\mathtt{C2}])^{I^{C2} \cup \{\mathtt{obs_i^{C2}} \mid \mathtt{obs_i} \notin (\mathcal{W} \cup \mathcal{O})\}}$, and the result follows because $\{\mathtt{obs_i^{C2}} \mid \mathtt{obs_i} \notin (\mathcal{W} \cup \mathcal{O})\}$ is a subset of $I^{C2}$. To see why the last containment hold, recall that $I^{C2}$ is a renaming of the model $M_2$, which must contain all the observations that are not in $(\mathcal{W} \cup \mathcal{O})$, by construction.

$C:$ $(P^3)^M = \{\mathtt{p}. \mid \mathtt{p}$ is an atom in $I^{guess}\}$.
$D:$ $(P^5)^M = P^5$.
$E:$ $(P^6)^M = \{\mathtt{s} \leftarrow \mathtt{o_i}, \mathtt{w_i}. \mid \mathtt{obs_i} \in P^{\text{obs}}\}$.
$F:$ $(P^7)^M = P^7$.
$G:$ $(P^8)^M = P^8$.
$H:$ $(P^9)^M = P^9$.

Then, the result follows because of the following two properties:

**Property $P_1$:** $M$ is a model of $\mathcal{L}^{\vee}(\mathcal{P})^M$.

*Proof.* All the rules in $\mathcal{L}^{\vee}(\mathcal{P})^M$ (see points $A - H$ above) are satisfied by $M$:

- Rules in $A$ are satisfied by $M$, as $\mathtt{satC1}$ is in $M$.
- Rules in $B$ are satisfied by $M$. Indeed, program $P(\mathcal{P})_{\mathcal{W},\mathcal{O}}[\mathtt{C2}]$ coincides with $P(\mathcal{P})_{\mathcal{W},\mathcal{O}}$ modulo a renaming of the predicate symbols, and $I^{C2}$ is in fact a renaming of the unique stable model $M_2$ of $P(\mathcal{P})_{\mathcal{W},\mathcal{O}}$, by construction. Then, the result follows since $M_2$ is, by definition, the minimal model of the positive program $P(\mathcal{P})_{\mathcal{W},\mathcal{O}}^{M_2}$.
- Rules in $C$ are satisfied by $M$ since $I^{guess}$ is a subset of $M$.
- Rules in $D$ are satisfied by $M$ since $\mathtt{satC2}, \mathtt{satC1}, \mathtt{satC1_1}, ..., \mathtt{satC1_n}$ are in $M$.
- Rules in $E$ are satisfied by $M$. Indeed, the set $I^{guess}$ is such that $\mathtt{w_i}$ is in $M$ if and only if $\mathtt{obs_i} \in \mathcal{W}$, and such that $\mathtt{o_i}$ is in $M$ if and only if $\mathtt{obs_i} \notin \mathcal{O}$. Then, given that by definition $\mathcal{W} \cap \mathcal{O} = \emptyset$, the body of such rules is always evaluated false in $M$.

– Rules in $F$ are satisfied by $M$ since the head of each of these rules belongs to $I^{C1} \cup I^{\phi} \cup I^{r}$, by construction.
– Rules in $G$ are satisfied by $M$ since satC1 is in $M$.
– Rules in $H$ are satisfied by $M$ since the head of each of these rules belongs to $I^{C1} \cup I^{\phi} \cup I^{r} \cup \{\mathtt{satC1_1}, ..., \mathtt{satC1_n}\}$, by construction.

**Property** $P_2$**:** There is no model $M'$ of $\mathcal{L}^{\vee}(\mathcal{P})^{M}$ such that $M' \subset M$.

*Proof.* Recall that program $\mathcal{L}^{\vee}(\mathcal{P})^{M}$ has the form shown in points $A - H$ above and assume, for the sake of contradiction, that there is a model $M' \subset M$ for $\mathcal{L}^{\vee}(\mathcal{P})^{M}$. First, observe that $I^{guess} \subseteq M'$ (see point $C$). Moreover, because of the fact that $\{\mathtt{obs}_\mathtt{i}^{\mathtt{C2}} \mid \mathtt{obs_i} \notin (\mathcal{W} \cup \mathcal{O})\}$ is a subset of $I^{C2}$ (point $B$, above), it can be observed that $M$ has in fact the form $M = I^{C1} \cup I^{C2} \cup I^{guess} \cup I^{\phi} \cup I^{r} \cup \{\mathtt{satC2}, \mathtt{satC1}, \mathtt{satC1_i}, ..., \mathtt{satC1_n}\}$.
Assume now that $I^{C2} \cap M' \subset I^{C2}$. Then, $I^{C2} \cap M'$ is a model for $(P^{\text{rls}}[\mathtt{C2}] \cup \{\mathtt{obs}_\mathtt{i}^{\mathtt{C2}} \mid \mathtt{obs_i} \notin (\mathcal{O} \cup \mathcal{W})\})^{I^{C2}}$, i.e., for $(P(\mathcal{P})_{\mathcal{W},\mathcal{O}}[\mathtt{C2}])^{I^{C2}}$. It follows that the set $M_2' = \{\mathtt{a} \mid \mathtt{a}^{\mathtt{C2}} \in (I^{C2} \cap M')\}$ is also a model for $P(\mathcal{P})_{\mathcal{W},\mathcal{O}}^{M_2}$ (notice that $(P(\mathcal{P})_{\mathcal{W},\mathcal{O}}[\mathtt{C2}])^{I^{C2}}$ coincides with $P(\mathcal{P})_{\mathcal{W},\mathcal{O}}^{M_2}$ modulo the renaming of predicate symbols). However, $M_2'$ is also a subset of $M_2$, and this is impossible because $M_2$ is the stable model of $P(\mathcal{P})_{\mathcal{W},\mathcal{O}}$ and, therefore, the minimal model of $P(\mathcal{P})_{\mathcal{W},\mathcal{O}}^{M_2}$.
Therefore, $I^{C2} \subseteq M'$. Consequently, $M' \subset M$ implies by construction of the program $\mathcal{L}^{\vee}(\mathcal{P})$ (rules in Step 9) that satC1 is not in $M'$. Then, by carefully looking at rules in Step 8, the following conclusion is reached:

(a) $I^{C1} \cap M'$ does not contain a pair of atoms of the form $\mathtt{p}^{\mathtt{C1}}$ and $\overline{\mathtt{p}}^{\mathtt{C1}}$ (rules 8.(a)) and each atom p occurring in $P^{\text{rls}} \cup P^{\text{obs}}$ contains at least a fact in $\{\mathtt{p}^{\mathtt{C1}}, \overline{\mathtt{p}}^{\mathtt{C1}}\}$ (rules 7.(a)). In the following, let $M_1$ denote the set $\{\mathtt{a} \mid \mathtt{a}^{\mathtt{C1}} \in (I^{C1} \cap M')\}$.
(b) For each atom p in $M_1$, exactly one atom of the form $\mathtt{p}\phi^{\mathtt{i}}$ is in $M'$ (rules 8.(b)). Thus, a mapping $\phi$ can be defined from atoms in $M_1$ to natural numbers such that $\phi(\mathtt{p}) = \mathtt{i}$ iff both $\mathtt{p}^{\mathtt{C1}}$ and $\mathtt{p}\phi^{\mathtt{i}}$ are in $M'$.
(c) For each atom p in $M_1$, exactly one atom of the form $\mathtt{p}r^{\mathtt{j}}$ among those having p in their head is in $M'$ (rules 8.(c)). Thus, an assignment $r$ can be defined from atoms in $M_1$ to the rules of $P^{\text{rls}} \cup P^{\text{obs}}$ such that $r(\mathtt{p}) = r^{\mathtt{j}}$ iff both $\mathtt{p}^{\mathtt{C1}}$ and $\mathtt{p}r^{\mathtt{j}}$ are in $M'$ and p occurs in the head of $r^{\mathtt{j}}$.
(d) For each atom p in $M_1$, the assignment $r$ defined in point (c) above is such that:
   i. $r(\mathtt{p})$ is not a fact in $P^{\text{obs}}$ belonging to $\mathcal{W}$ (rules 8.(d)), i.e. p cannot be entailed by exploiting a fact removed by $P^{\text{obs}}$ since it is part of the witness $\mathcal{W}$;
   ii. $r(\mathtt{p})$ is a rule which does not contain an atom $\overline{\mathtt{c}}$ in the body, where c is in $M_1$ (rules 8.(e)), i.e. p cannot be entailed by exploiting a rule which is not in the reduct $(P^{\text{rls}})^{M_1}$ since there is a negated atom $c$ in its body which is true in $M_1$; combined with point (i) above, these two constraints impose that $r$ is an assignment of atoms to rules in $P(\mathcal{P})_{\mathcal{W}}^{M_1}$.
   iii. $\mathbf{b}(r(\mathtt{p})) \subseteq M_1$ (rules 8.(f));
   iv. $\phi(q) < \phi(p)$, for each q occurring positively in $\mathbf{b}(r(p))$ (rules 8.(g)).

Armed with the properties above, it can be concluded that $M'$ defines a mapping $\phi$ from $M_1$ to rules in $P(\mathcal{P})_{\mathcal{W}}^{M_1}$ (with an associated mapping $r$ for the atoms to the rules) such that all the conditions in Lemma 1 are satisfied. Furthermore, by rules 2.(a) and 4.(b), $M_1$ is a model for the program $P(\mathcal{P})_{\mathcal{W}}$, otherwise satC1 would be entailed. Thus, $M_1$ is trivially a model for $P(\mathcal{P})_{\mathcal{W}}^{M1}$ and, by virtue of Lemma 1, it is in fact a stable model for $P(\mathcal{P})_{\mathcal{W}}$. However, such a stable model witnesses that condition (1) in Definition 1 is violated, because of the rules in Step 5 and the fact that satC1 is false in $M'$, which entails the existence of an observation in the witness set $\mathcal{W}$ which is in $M_1$. Contradiction.

2. Let $M$ be a stable model of $\mathcal{L}^{\vee}(\mathcal{P})$. First of all, note that by rules inserted into $\mathcal{L}^{\vee}(\mathcal{P})$ in Step 6 of the algorithm, $M$ is such that $(i)$ for each letter $\mathtt{obs_i}$ in $P^{\rm obs}$, $\mathtt{o_i}$ and $\mathtt{w_i}$ cannot belong simultaneously to $M$, $(ii)$ $\mathtt{satC2} \in M$, and $(iii)$ $\mathtt{satC1} \in M$. Furthermore, by rules inserted into $\mathcal{L}^{\vee}(\mathcal{P})$ in Steps 3.(a) and 3.(b) it is the case that, for each letter $\mathtt{obs_i}$ in $P^{\rm rls}$, either $\mathtt{o_i}$ or $\overline{\mathtt{o}}_\mathtt{i}$ and either $\mathtt{w_i}$ or $\overline{\mathtt{w}}_\mathtt{i}$ belong to $M$. Consider, now, the disjoint sets $\mathcal{O} = \{\mathtt{obs_i} \mid \mathtt{o_i} \in M\}$ and $\mathcal{W} = \{\mathtt{obs_i} \mid \mathtt{w_i} \in M\}$. It will be shown that both conditions in Definition 1 are satisfied. To this aim consider the interpretations $I^{C1}$, $I^{guess}$, and $I^r$ defined in point (1) of this proof, and notice that $M$ can be written as $I^{C1} \cup I^{C2} \cup I^{guess} \cup I^{\phi} \cup I^r \cup \{\mathtt{satC2}, \mathtt{satC1}, \mathtt{satC1_i}, ..., \mathtt{satC1_n}\}$, where $I^{C2}$ is the subset of $M$ containing all the predicates in $M$ of the form $\mathtt{p^{C2}}$ and $\overline{\mathtt{p}}^{\mathtt{C2}}$. Indeed, since $\mathtt{satC1}$ is in $M$, the model must contain also the elements in $I^{C1} \cup I^{\phi} \cup I^r \cup \{\mathtt{satC2}, \mathtt{satC1_i}, ..., \mathtt{satC1_n}\}$. At this point the reader may check that the rules in $\mathcal{L}^{\vee}(\mathcal{P})^M$ have again the form illustrated in points $A - H$ above.
To conclude, the following properties can be shown:

   - $P(\mathcal{P})_{\mathcal{W},\mathcal{O}} \not\models_c \neg\mathcal{W}$.

     *Proof.* Given that $M$ is a minimal model of $\mathcal{L}^{\vee}(\mathcal{P})^M$, it can be concluded that $I^{C2}$ is a minimal model of $(P^{\rm rls}[\mathtt{C2}] \cup \{\mathtt{obs_i^{C2}} \mid \mathtt{obs_i} \not\in (\mathcal{W} \cup \mathcal{O})\})^M = (P(\mathcal{P})_{\mathcal{W},\mathcal{O}}[\mathtt{C2}])^{I^{C2}}$. Then, $M_2 = \{\mathtt{a} \mid \mathtt{a^{C2}} \in I^{C2}\}$ is a minimal model of $(P(\mathcal{P})_{\mathcal{W},\mathcal{O}})^{M_2}$. Now, it can be shown that $P(\mathcal{P})_{\mathcal{W},\mathcal{O}} \not\models_c \neg\mathcal{W}$. Indeed, assume, for the sake of contradiction, that $P(\mathcal{P})_{\mathcal{W},\mathcal{O}} \models_c \neg\mathcal{W}$. Then, the stable model $M_2$ does not contain any fact in $\mathcal{W}$, and symmetrically $I^{C2}$ does not contain any fact of the form $\mathtt{obs_i^{C2}}$, with $\mathtt{obs_i}$ in $\mathcal{W}$. Then, $\mathtt{satC2}$ cannot be entailed by $M$, which is impossible.

   - $P(\mathcal{P})_{\mathcal{W}} \models_c \neg\mathcal{W}$.

     *Proof.* To this aim, observe that $M$ is a minimal model of $\mathcal{L}^{\vee}(\mathcal{P})^M$ and assume, for the sake of contradiction, that there is a fact $\mathtt{obs_i}$ in $\mathcal{W}$ and a stable model $M_1$ for $P(\mathcal{P})_{\mathcal{W}}$ such that $\mathtt{obs_i}$ is in $M_1$. Then, given $\bar{I}^{C1} = \{\mathtt{a^{C1}} \mid \mathtt{a} \in \mathtt{M_1}\} \cup \{\overline{\mathtt{a}}^{\mathtt{C1}} \mid \mathtt{a} \not\in M_1\}$, the set $M' = \bar{I}^{C1} \cup I^{C2} \cup I^{guess} \cup \bar{I}^{\phi} \cup \bar{I}^r$ is also a model for $\mathcal{L}^{\vee}(\mathcal{P})^M$, where $\bar{I}^{\phi}$ and $\bar{I}^r$ encode the assignments $\phi$ and $r$ in Lemma 1. Since $M' \subset M$, there is a contradiction with $M$ being a stable model of $\mathcal{L}^{\vee}(\mathcal{P})$ $\square$

This section concludes by observing that minimum-size outlier detection problems can be modelled for general rule-observation pairs by exploiting the same approach used for the stratified pairs, i.e., by introducing *weak constraints* in the program built by the algorithm in Figure 6. Specifically, given a pair $\mathcal{P}$, the algorithm can be modified by inserting into $\mathcal{L}^{\vee}(\mathcal{P})$ a constraint $:\sim\ \mathtt{o_i}.$ for each $\mathtt{obs_i} \in P^{\rm obs}$. Then, letting $\mathcal{L}^{\vee,\sim}(\mathcal{P})$ be the transformed program, the same arguments as Theorem 22 with respect to the construction of Theorem 23 can be used for showing that minimum-size outliers in $\mathcal{P}$ are in one-to-one correspondence with best stable models of $\mathcal{L}^{\vee,\sim}(\mathcal{P})$.

**Theorem 24.** *Let $\mathcal{P} = \langle \mathcal{P}^{\rm rls}, \mathcal{P}^{\rm obs} \rangle$ be a rule-observation pair. Then,*

1. *for each minimum-size outlier $\mathcal{O}$ with witness $\mathcal{W}$ in $\mathcal{P}$, there is a best stable model $\mathcal{M}$ of $\mathcal{L}^{\vee,\sim}(\mathcal{P})$ such that $\{\mathtt{obs_i} \mid \mathtt{o_i} \in \mathcal{M}\} = \mathcal{O}$ and $\{\mathtt{obs_i} \mid \mathtt{w_i} \in \mathcal{M}\} = \mathcal{W}$, and*
2. *for each best stable model $\mathcal{M}$ of $\mathcal{L}^{\vee,\sim}(\mathcal{P})$, there is an outlier $\mathcal{O}$ with witness $\mathcal{W}$ in $\mathcal{P}$, such that $\{\mathtt{obs_i} \mid \mathtt{o_i} \in \mathcal{M}\} = \mathcal{O}$ and $\{\mathtt{obs_i} \mid \mathtt{w_i} \in \mathcal{M}\} = \mathcal{W}$.*

## 7 Related Work

In this section, the most relevant related work is discussed. Research work related to that presented in this paper can be roughly divided into two groups: $(i)$ work done on outlier detection from logic theories, which is very relevant to our own, and $(ii)$ work done on outlier detection from data, which is, on the contrary, less related to concepts discussed in this paper.

### 7.1 Outlier detection from data

Outlier detection problems come in several different varieties within different settings, mainly investigated in the area of statistics, machine learning and knowledge discovery in databases. In almost all cases the presented approaches deal with data organized in a single relational table, often with all the attributes being numerical, while a metrics relating each pair of rows in the table is required.

The approaches to outlier detection can be classified as *supervised*-learning based methods, where each example must be labelled as exceptional or not [60, 77], and the *unsupervised*-learning based ones, where such labels are not required. The latter methods are obviously more general than the former ones. As the technique proposed in this work is unsupervised, the sequel of this section shall refer only to unsupervised methods. In their turn, unsupervised-learning based methods for outlier detection can be categorized in several groups.

The first group is that of *statistical-based* methods that assume that the given data set has a distribution model. Outliers are those objects that satisfy a discordancy test, i.e., which are significantly larger (or smaller) in relation to the hypothesized distribution [11].

*Deviation-based* techniques identify outliers by inspecting the typical characteristics of objects and consider an object that deviates from these features an outlier [10].

A completely different approach that finds outliers by observing *low dimensional projections* of the search space is presented in [3]. Here, a point is considered an outlier if it deviates from the other data in some subspace.

Further groups consists of *density-based* techniques [16], using a notion of *local outlier* that measures the plausibility for an object to be an outlier with respect to the density of the local neighborhood.

*Distance-based* outlier detection was introduced by Knorr and Ng [55] to overcome the limitations of statistical methods. A *distance-based* outlier is defined as follows: A point p in a data set is an outlier with respect to parameters $k$ and $\delta$ if at least $k$ points in the data set lie at a distance greater than $\delta$ from $p$. This definition generalizes the definition of outlier in statistics and is well suited when the data set does not fit any standard distribution. Ramaswamy et al. [73] modified the above definition of outlier, since that does not provide any ranking for outliers that are singled out. The new definition is based on the distance of the $k$-th nearest neighbor of a point $p$, denoted with $D^k(p)$, and it is as follows: Given $k$ and $n$, a point $p$ is an outlier if no more than $n-1$ other points $q$ in the data set have a higher value for $D^k(q)$ than $p$. This means that the points $q$ having the $n$ greatest $D^k(q)$ values are considered outliers. In [8] a new definition of outlier that considers for each point the sum of the distances from its $k$ nearest neighbors is proposed. The authors presented an algorithm using the Hilbert space-filling curve that exhibits scaling results close to linear. An analogous definition of outlier based on the $k$-nearest neighbors has been used in [32] for unsupervised anomaly detection in intrusion detection applications. In [12] a near-linear time algorithm for the detection of distance-based outliers exploiting randomization is presented.

The general differences and analogies between the approaches described above and our own should be clearly understood. In fact, those approaches deal with “knowledge”, as encoded within one single relational table that is, in a sense, flat, i.e. such that there is no explicit relationship linking the objects (tuples) of the data set under examination. Vice versa, the technique proposed in this work deals with complex knowledge bases, which may well comprise relational-like information, but generally also include semantically richer forms of knowledge, such as logical rules: in this latter case several complex relations relating objects (atoms) of the underlying theory might be explicitly available. As a consequence, even if the intuitive and general sense of computing outliers in the two contexts is analogous, the conceptual and technical developments are quite different in the two contexts, just as the formal properties of computed outliers are different.

### 7.2 Outlier detection using logic

Recently, outlier detection has emerged as an interesting knowledge representation and reasoning problem, in the context of default logic [7].

In particular, [7] originally introduced and investigated the concept of outlier detection in the context of default logics. In that context, we have a propositional default theory $\Delta = (D, W)$, where $D$ is a set of defaults and $W$ is a set of propositional formulas. An outlier is then defined as a literal in $W$ which is not justified in $\Delta$ with respect to a witness $S \subseteq W$. In that paper, the complexity of singling out outliers was studied for several fragments of Reiter's propositional default logics [75], that is, general propositional theories, disjunction-free theories, normal mixed unary theories, unary and dual unary theories, and acyclic unary and acyclic dual unary theories. The complexity results range from P to $\Sigma_3^{\mathrm{P}}$. In their analysis, the authors investigated only the cautious form of reasoning and defined an outlier as a singleton set, thus the concept of minimality has no meaning in that setting.

In this paper, the paradigm of [7] has been extended and generalized in several respects by (*i*) adapting the notion of outlier to the paradigm of logic programming in such a way that outliers are no longer constrained to denote single individuals, (*ii*) investigating the complexity of the corresponding detection problems, (*iii*) considering also the brave form of reasoning, (*iv*) defining significant minimization-based outlier detection problems and studying their complexity, (*v*) extending the framework to have observations encoded as a logic theory, and, finally, (*vi*) providing rewriting techniques for effective outlier detection implementation.

In conclusion, let us observe that the class of default theories studied in [7] most related to general logic programs is that of disjunction-free theories (DF). Indeed, it is well-known [42] that a general logic program $P$ can be translated into an equivalent DF default theory $\Delta(P)$. However, the complexity results in [7] cannot be straightforwardly translated to the case of theories expressed as logic programs, even in the restricted framework of cautious reasoning for detecting outliers as individual elements.

Indeed, the hardness results in [7] for DF default theories rely on *normal* theories, whose relationships with logic programs has been not studied in the literature so far. And, in fact, there seems to be not much sense in translating a logic program into a normal default theory. Even a simple program like $\alpha \leftarrow not\ \beta$ cannot be translated into normal defaults because some form of $\beta$ is needed in the justifications. A form that would properly translate into normal is $\alpha \leftarrow \beta, not\ \neg\alpha$, and this requires classical negation. Nevertheless, if one has a program with no negation of any form, one can translate it to normal: $\alpha \leftarrow \beta$ can be translated to $\frac{\beta:\alpha}{\alpha}$. Since there is no negation, the extension one obtains will be equivalent to the stable model.

Furthermore, since normal default theories satisfy the semi-monotonicity property, it is not clear how to design a polynomial-time translation from a logic program $P$ under stable model semantics to a default theory $\Delta(P)$ ensuring a bijection between stable models of $P$ and extensions of $\Delta(P)$.

## 8 Conclusion

In several knowledge-based applications, the scenario is significant where a rational agent equipped with his own trustable knowledge about the world has some information coming from the outside in the form of a set of observations denoting the status or facts about the external environment. Then, it is useful to let the agent be capable of discovering those observations (if any) whose truth clashes to some extent with some of the agent trustable beliefs. Such observations, called outliers, do indeed embody some abnormal status of things which is interesting, at least, to be singled out and checked.

Differently from what is most often done in the literature, outliers have been defined in the paper on the basis of some logical properties, rather than being determined by statistical characteristics. In particular, by extending the work presented in [7], the concept of outlier has been formalized in the framework of logic-programming based knowledge systems under both cautious and brave stable model semantics.

It has been shown that for a fact being an outlier depends on the given observation context: a fact may well be an outlier within a given observation set, while being normal in some other one. In words, it has been illustrated that outliers can be detected on the basis of observations to hand, by eventually singling out some properties standing out for their abnormality. In the proposed setting, it is in fact necessary to single out a supporting set, called the witness set, for the outlier to be singled out in turn.

The aim of the paper has been that of formalizing several variants of outlier detection problems, of showing their complexity and providing algorithms for outlier detection via rewriting.

The outlier detection framework introduced here resembles such important notions in AI such as diagnosis and abductive reasoning and belief revision. Still, several important differences are there that confirm the usefulness of such novel formalization: belief revision, abduction and other reasoning task related to outlier detection have been comparatively discussed in the paper, where such differences have been highlighted.

Complexity figures obtained for outlier detection problems are summarized in Figure 3 and 4 referring, respectively, to the basic and the min-cost variant of the problems. The complexity results show that outlier detection can be sometimes easy but it is intractable in most cases.

An interesting scenario, which has been analyzed as well in the paper, occurs when one focuses on the complexity of the problem where only the observation component is assumed to vary (that is, to denote the problem input), whereas the rule component is assumed to be fixed. This is the complexity measure known as *data complexity*, and is relevant whenever the size of the evidence data is large as compared to the size of the knowledge base formalizing his expected properties.

Moreover, the case where the observations are encoded as a logical theory, rather than "simple" sets of facts, has been considered. The concept of outlier has been properly generalized to arrange this idea, and the computational complexity of the problems arising in this extended setting has been also accounted for in the paper.

Finally, sound and complete translations of outlier problems have been provided into equivalent inference problems under stable model semantics. These translations can be taken advantage of in order for effectively implementing outlier detection on top of any available stable model solver (e.g., [49, 61, 67, 63]).

Several research questions are left often by the paper. As far as the basic definition of outliers (and associated witness) is concerned, it might be investigated whether there are some alternative notions that appear convenient for modelling abnormality in the observations to hand in some specific kinds of application. As an example, one may study a notion where condition (1) in Definition 1 is provided according to the brave (resp., cautious) semantics, while condition (2) is provided according to the cautious (resp., brave) semantics, and investigate how this notion compare with the one proposed in the paper.

As far as the complexity studies are concerned, it is believed to be relevant to have the picture proposed in the paper completed by investigating the *combined* complexity of detection problems for first-order programs, i.e., to study a setting where both the rule and the observation component are considered part of the input problem.

Finally, from the application side, it would be interesting to study whether logic-based outlier detection may have some fruitful applications in the database context, e.g., whether it can support the handling of sophisticated kinds of constraint or whether it can be even practicable as a data mining technique. Indeed, even though outlier detection problems appear to be intractable in most cases (unless $P = NP$), it might be still the case that suitable algorithms, approximations and heuristics can be defined to cope with them in an efficient way.

## Acknowledgments

Thanks to Rachel Ben-Eliyahu-Zohary for fruitful discussion on the subjects of this paper and for having originally set up the outlier detection scenario.